\documentclass[11pt]{article}

\usepackage[final]{acl}
\usepackage{times}
\usepackage{latexsym}
\usepackage[T1]{fontenc}
\usepackage[utf8]{inputenc}
\usepackage{microtype}
\usepackage{inconsolata}
\usepackage{graphicx}
\usepackage{booktabs}
\usepackage{longtable}
\usepackage{tabularx}
\usepackage{array}
\usepackage{multirow}
\usepackage{amsmath}
\usepackage{xcolor}
\usepackage{xurl}
\usepackage{fontawesome5}
\usepackage{adjustbox}
\usepackage{float}
\usepackage{placeins}
\usepackage{needspace}

\usepackage[most]{tcolorbox}
\usepackage{fvextra}
\tcbuselibrary{breakable}
\newcommand{\cleanbench}{\textsc{MedQA-MM}}

\title{MedQA-MM: Shortcuts Behind Medical Visual Reasoning}

\author{
\textbf{Benlu Wang}\textsuperscript{1,4},
\textbf{Yifan Zhang}\textsuperscript{2,3},
\textbf{Jiaqing Yu}\textsuperscript{7},
\textbf{Chin Siang Ong}\textsuperscript{5}, \\
\textbf{Juncheng Huang}\textsuperscript{9},
\textbf{Zhuohao Li}\textsuperscript{8},
\textbf{Zhenyu Zhang}\textsuperscript{6}, \\
\textbf{Arman Cohan}\textsuperscript{4},
\textbf{Hong Yu}\textsuperscript{1,2,3},
\textbf{Zonghai Yao}\textsuperscript{1,2,3} \\
\small
\textsuperscript{1}University of Massachusetts Amherst \quad
\textsuperscript{2}University of Massachusetts Lowell \quad
\textsuperscript{3}VA Bedford Healthcare System \\
\small
\textsuperscript{4}Yale University \quad
\textsuperscript{5}Yale School of Medicine \quad
\textsuperscript{6}Stanford University \\
\small
\textsuperscript{7}Qingdao Medical College of Qingdao University \quad
\textsuperscript{8}Zhejiang University \\
\small
\textsuperscript{9}National University Hospital, Singapore \\
\small{
\textbf{Correspondence:}
\href{mailto:zonghaiyao@umass.edu}{\texttt{zonghaiyao@umass.edu}},
\href{mailto:benlu.wang@yale.edu}{\texttt{benlu.wang@yale.edu}}
} \\[-0.1em]
\small
\textbf{Project repository:}
\href{https://github.com/benluwang/MedQA-MM-EMNLP2026}{\faGithub\hspace{0.35em}\texttt{github.com/benluwang/MedQA-MM-EMNLP2026}}
\\[-0.1em]
\small
\textbf{Hugging Face dataset:}
\href{https://huggingface.co/datasets/bio-nlp-umass/MedQA-MM}{\raisebox{-0.16em}{\includegraphics[height=1.05em]{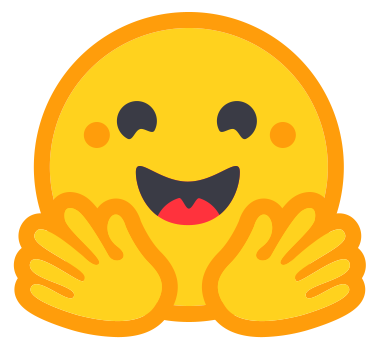}}\hspace{0.30em}\texttt{huggingface.co/datasets/bio-nlp-umass/MedQA-MM}}
}

\begin{document}
\maketitle

\begin{abstract}
A benchmark score credits final answers, but not the route by which an item can be answered. In medical multimodal multiple-choice questions (MCQs), this distinction matters because a correct answer can be supported by the intended image finding or by benchmark-preserved cues in the wording of answers, non-visual clinical text, visible image text, artificial annotations, or device/context artifacts. We call the resulting score-level overinterpretation \textit{reasoning inflation}. Here, a route is an observable input path that can support answer selection, not a claim about the model's hidden cognition. Across six medical multimodal MCQ datasets, we separate candidate cues from behavioral evidence through prompt- and image-side audits, modality ablations, and matched repairs that preserve the medical target and answer key. In a 13-configuration open-model panel, full-input accuracy is 62.63\%, while text-only and options-only settings achieve 53.96\% and 29.71\%, respectively. Removing length-gap, absolute/conspicuous, and spatial/prepositional cues lowers accuracy by 6.58, 3.50, and 4.77 percentage points. We also construct \textsc{MedQA-MM}, a 1,000-item shortcut-mitigated subset, where text-only and options-only accuracy fall to 5.21\% and 12.33\%. This does not imply that models never use images; it shows that medical image-reasoning claims require route-level evidence.
\end{abstract}

\section{Introduction}

\begin{figure}[!ht]
    \centering
    \includegraphics[width=\linewidth]{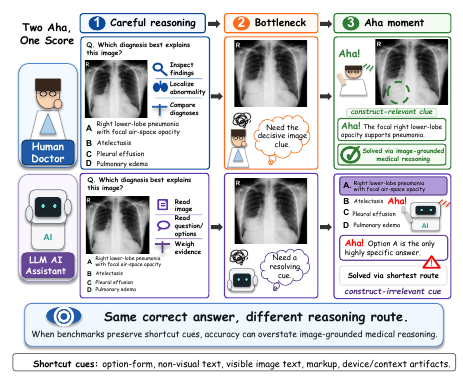}
    \caption{Two routes can produce the same correct answer. The intended ``aha'' comes from a construct-relevant image clue, while the shortcut ``aha'' comes from option form, non-visual context, visible text, markup, or other benchmark-preserved cues. A single MCQ score does not distinguish these routes.}
    \vspace{-5mm}
    \label{fig:aha}
\end{figure}

A benchmark records a correct answer, not the route that produced it. Figure~\ref{fig:aha} illustrates the gap. A human expert and a model may face the same image-based MCQ, reach a bottleneck, and then find a decisive cue. One route uses the construct-relevant visual finding. The other route uses a benchmark-preserved shortcut, such as an unusually specific answer option, a non-visual clue in the stem, visible text in the image, or an annotation that points to the target. Both routes can end with the same answer, but they do not support the same capability claim. We use the ``aha'' contrast as a shorthand for this decisive turn in an answer route, not as a claim about hidden model cognition.

This route ambiguity is a problem for NLP evaluation. Medical multimodal MCQ benchmarks are often reported as compact accuracy scores for vision-language models (VLMs). Those scores are then read as evidence that a model can perform image-grounded clinical reasoning. The interpretation is stronger than the number itself. If an item can also be solved through option form, no-image text, visible labels, or image-side markup, the score mixes the intended image-grounded route with easier shortcut routes.

The intended construct in our setting is the ability to integrate medical visual evidence, clinically relevant text, and answer-option semantics to select the best answer. A \textit{shortcut} is a construct-irrelevant signal that enables above-chance answer selection without this intended image-grounded route. This definition is item-level. Clinical context, devices, demographic facts, and figure annotations are not invalid by default: a chest tube may be the target finding, an arrow may define the task, and age or sex may be clinically necessary. The validity question is whether, in a specific item, the cue bypasses the route that the benchmark score is used to measure.

We call the resulting score-level risk \textit{reasoning inflation}: benchmark accuracy is interpreted as direct evidence of image-grounded reasoning even when some credited decisions can follow shortcut routes. This does not mean that models never use images. It means that score movement on an unaudited benchmark is hard to interpret, because the benchmark may reward multiple routes while reporting one number.

Medical multimodal MCQs combine shortcut risks that have usually been studied separately. Multiple-choice assessment work has long warned that option length, absolute wording, paired options, grammatical mismatch, and implausible distractors can cue the answer \citep{Haladyna2002}. NLP benchmark studies show that models can exploit annotation artifacts and shallow heuristics without solving the intended task \citep{Gururangan2018,Poliak2018,McCoy2019}. VQA work shows that language priors can support high accuracy when image evidence is weakly used \citep{Agrawal2016,Goyal2017,Ramakrishnan2018}. Medical images add visible text, arrows, overlays, devices, acquisition markers, and care-process artifacts. What is missing is a route-level audit that connects these risks in medical multimodal MCQs and tests whether suspected cues change model behavior.

We study this problem through a data-centric audit-and-repair framework. Across six medical multimodal MCQ datasets, we canonicalize items, separate prompt-visible fields from source-only metadata, screen text- and option-side cues, audit image-side leakage, and evaluate models under full-input, text-only, options-only, and Image+Options settings. We then apply matched repairs when safe minimal edits are possible. A repair is admissible only when it targets the suspected cue, preserves the medical target and answer key, retains a one-best-answer structure, and leaves a traceable original-edited pair. This evidence ladder separates candidate cues from stronger behavioral evidence.

Our results show why route-level evidence matters. In a 13-configuration open-model panel, full-input accuracy averages 62.63\%, while text-only accuracy remains 53.96\% and options-only accuracy reaches 29.71\%. Matched repairs provide stronger evidence that specific cues are behaviorally useful: removing length-gap cues lowers accuracy by 6.58 pp across 306 cases; removing absolute/conspicuous distractor cues lowers accuracy by 3.50 pp across 101 cases; and removing spatial/prepositional cues lowers accuracy by 4.77 pp across 101 cases. Finally, \textsc{MedQA-MM}, a 1,000-item shortcut-mitigated subset, lowers text-only accuracy to 5.21\% and options-only accuracy to 12.33\% under the tested probes. We therefore treat \textsc{MedQA-MM} as shortcut-mitigated, not shortcut-free.

\paragraph{Contributions.}
(1) We introduce a two-aha framing for route ambiguity in multimodal MCQ evaluation and formalize the score-level risk as \textit{reasoning inflation}.
(2) We provide a six-dataset audit that connects MCQ cueing, NLP dataset artifacts, VQA modality dependence, and medical image-side shortcuts under one evidence ladder.
(3) We use matched repairs to show that length-gap, absolute/conspicuous, and spatial/prepositional cues measurably affect model accuracy.
(4) We construct \textsc{MedQA-MM} and argue that shortcut mitigation, validation status, and residual-risk reporting should be part of benchmark release practice.

\section{Related Work}

\paragraph{Reasoning narratives, score validity, and MCQ cueing.}
Chain-of-thought prompting and reasoning-oriented models have made intermediate reasoning behavior a common part of model evaluation \citep{Wei2022ChainOfThought,Guo2025DeepSeekR1}. We use the language of an ``aha'' moment only as a shorthand for the decisive cue or turn that a benchmark may reward. The underlying validity question is older: whether evidence supports the intended interpretation of a score \citep{Messick1995}. In MCQs, item-writing flaws can let an examinee answer without the target knowledge, including option-length imbalance, grammatical mismatch, conspicuous choices, absolute wording, paired options, and implausible distractors \citep{Haladyna2002,Schuwirth1996,Downing2005,Tarrant2008}. We import this lens to VLM benchmarks, where the examinee is a model and the score is often interpreted as image-grounded reasoning.

\paragraph{Dataset artifacts and modality shortcuts.}
NLP benchmarks have repeatedly shown that high accuracy can hide unintended solution strategies. Hypothesis-only NLI baselines exposed annotation artifacts, and syntactic challenge sets showed that aggregate scores can mask shallow heuristics \citep{Gururangan2018,Poliak2018,McCoy2019}. VQA work similarly shows that language priors and answer-set regularities can inflate apparent visual understanding \citep{Agrawal2016,Goyal2017,Ramakrishnan2018}. Recent VLM diagnostic suites test visual indispensability, hallucination, and robustness to reduced or altered visual information \citep{Chen2024MMStar,Yue2025MMMUPro,Guan2024HallusionBench,Li2023POPE,Asadi2026MIRAGE}. Medical multimodal MCQs inherit these risks and add answer-option cueing, so we combine no-image diagnostics, options-only probes, and matched repairs.

\paragraph{Medical multimodal evaluation and image-side shortcuts.}
Medical VLMs and generalist biomedical systems are increasingly evaluated through image-based questions and exam-style benchmark suites \citep{Li2023LLaVAMed,Moor2023MedFlamingo,Tu2024GeneralistBiomedicalAI,Saab2024MedGemini,Hu2024OmniMedVQA,Zuo2025MedXpertQA,Yao2026MedThinkVQA}. These evaluations make score interpretation more important, because medical images can contain visible labels, arrows, overlays, devices, acquisition markers, and care-process artifacts. Medical imaging work has also shown that models can exploit site, acquisition, device, demographic, or process signals rather than target pathology \citep{Geirhos2020,Lapuschkin2019CleverHans,Zech2018,Badgeley2019,DeGrave2021,Banerjee2023}. Because many clinical cues are legitimate in some items, our audit treats construct relevance as item-specific rather than category-wide.

\paragraph{Data-centric benchmark repair.}
Dataset documentation, dynamic benchmarking, competency-oriented evaluation, and counterfactual data work argue that benchmark quality depends on data construction and maintenance, not only on model scoring \citep{Bender2018,Gebru2021,Kiela2021,Gardner2021,Kaushik2020}. We follow this data-centric view for evaluation. A shortcut audit should not end with a list of flaws, and medical data scarcity makes blanket deletion costly. The more useful workflow is to detect candidate cues, repair them when safe, validate that the medical target and one-best-answer structure remain intact, and document residual uncertainty. This turns benchmark repair into part of evaluation design rather than post-hoc leaderboard commentary.

\begin{figure*}[!ht]
    \centering
    \includegraphics[width=\textwidth]{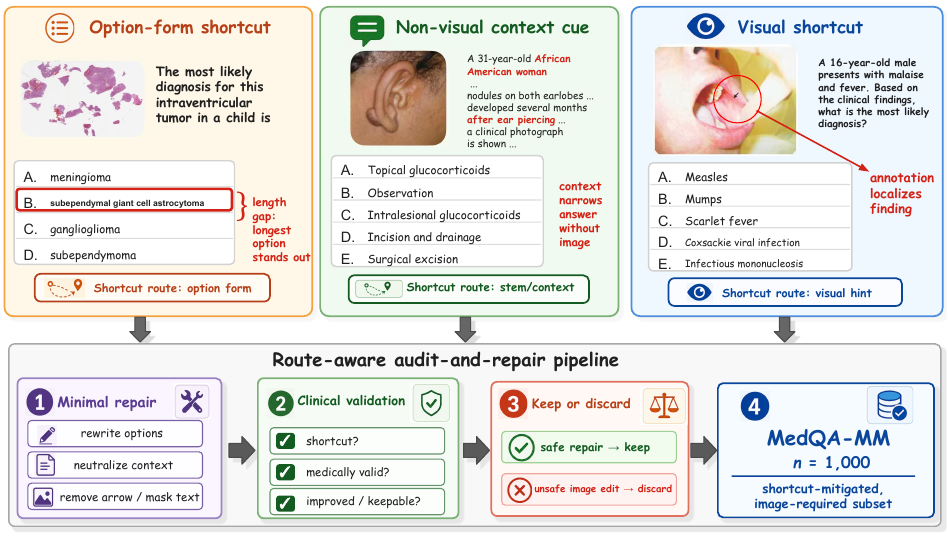}
\caption{
\textbf{From shortcut routes to shortcut-mitigated evaluation.}
Correct medical MCQ answers may come from option form, non-visual context, or image-side hints rather than the intended image-grounded route.
We audit these route-level risks, apply minimal safe repairs, and validate medical semantics before retaining items in \textsc{MedQA-MM}.
}
    \vspace{-2mm}
\label{fig:framework}
\end{figure*}

\section{Method}

\subsection{Construct and Evidence Ladder}
\label{sec:method-construct}

The intended construct is image-grounded medical MCQ answering: a model should use the image, clinically relevant text, and answer-option semantics to select the one best answer. A shortcut is a construct-irrelevant signal that supports above-chance answer selection without this intended route. Because the same surface cue can be clinically valid in one item and shortcut-like in another, automatic detections are treated as candidates rather than conclusions.

Figure~\ref{fig:framework} summarizes the audit-and-repair workflow, and Table~\ref{tab:method_analysis_families} separates the main evidence families. Our audit follows an evidence ladder. A \textit{screened candidate} is flagged by a rule, model judge, or semantic detector. An \textit{ablation-supported candidate} remains answerable under a partial-input prompt, such as no image or options only. A \textit{repair-supported candidate} changes model behavior after a matched cue-removal or cue-neutralization edit. An \textit{expert-validated candidate} is judged to preserve medical validity and reduce the shortcut. Final item disposition depends on cue status, repair fidelity, validation status, and release constraints.

\begin{table}[t]
\centering
\scriptsize
\setlength{\tabcolsep}{2.5pt}
\renewcommand{\arraystretch}{1.03}
\begin{tabular}{@{}p{0.37\columnwidth}p{0.22\columnwidth}p{0.31\columnwidth}@{}}
\toprule
Analysis family & Main set & Role \\
\midrule
Length-gap repair & 306 pairs & Option edit; 20$\times$ \\
Spatial/prepositional repair & 101 pairs & Option edit; 20$\times$ \\
Absolute/conspicuous repair & 101 pairs & Option edit; 20$\times$ \\
Four-condition modality audit & 7,706 cases & Partial-input route audit \\
Diagnostic guessers & 3,743 eval. cases & Options-only / Q+O stress test \\
Artificial image hints & 200 pairs & Image edit; 10$\times$ \\
Natural clinical-context cues & 302 pairs & Stem neutralization; 10$\times$ \\
Visible image-text leakage & 65 high-risk cases & Data-quality audit \\
Device/material cues & 9 confirmed, 48 borderline & Clinical artifact audit \\
\bottomrule
\end{tabular}
\vspace{-2mm}
\caption{Main analysis families. Paired rows compare original items with minimally edited counterparts; 20$\times$/10$\times$ indicate repeated evaluations per item or pair. Image-text and device/material counts are treated as curation risks unless paired evidence is available.}
\vspace{-5mm}
\label{tab:method_analysis_families}
\end{table}

\subsection{Shortcut Scope}
\label{sec:method-audit}

Text-side screens cover option length, length gaps, answer position, grammar and answer-type asymmetry, spatial/prepositional structure, absolute or exclusionary wording, all/none meta-options, stem--option lexical overlap, and semantic compatibility. Image-side screens cover visible text, manual markup, artificial hints, visible context or demographic features, device or post-treatment material, and acquisition/source artifacts. These screens feed targeted ablations, repairs, and expert review.

\subsection{Option-Form Shortcuts}
\label{sec:method-option-repairs}

We build three option-form repair sets. Length-gap repairs target cases where the correct option is uniquely longer or more specific than distractors. Spatial/prepositional repairs target options where the correct answer is uniquely marked by location, direction, source, pathway, procedure target, range, or threshold phrasing. Absolute/conspicuous repairs target distractors made implausible by all/none, only, never, always, singleton-scope, or rule-like wording.
For each family, GPT-5.4 proposes minimal edits that preserve the stem, image, correct label, option order, and number of choices whenever possible. Edits are retained only if they preserve the medical target, keep the same one best answer, and reduce the suspected surface cue. The final sets contain 306 length-gap pairs, 101 spatial/prepositional pairs, and 101 absolute/conspicuous pairs. Models are evaluated on each original and repaired item with 20 repeated runs per condition. We use a common cue-removal convention, $\Delta_{\mathrm{text}} = \mathrm{Acc}_{\mathrm{edited}} - \mathrm{Acc}_{\mathrm{original}}$, so negative values indicate that cue removal reduced accuracy. Appendix~\ref{app:option-route-details} gives construction details and complete model-level accuracy changes; Appendix~\ref{app:option-directional-controls} defines the destination metrics, length reverse and negative controls, spatial length-confound analysis, and additional model-level results. Appendix~\ref{app:prompts} reproduces the repair prompts.

\subsection{Partial-Input Probes}
\label{sec:method-partial-input}

We evaluate four prompt settings: \textit{full input} (image, question, and options), \textit{text-only} (full question text and options), \textit{options-only} (answer choices only), and \textit{Image+Options} (image and choices without other question text). These settings test whether full-input scores isolate image-grounded reasoning. We also use diagnostic options-only and Question+Options guessers as stress tests for non-image solvability; they are not treated as deployable systems or leaderboard baselines.

\subsection{Image-Channel and Natural-Cue Interventions}
\label{sec:method-image-natural}

For artificial image hints, GPT-5.4 filters manual annotations and answer relevance. From 856 strong image-hint candidates, we sample cases for repair, localize the mark, edit a masked crop, composite it back into the original image, and judge the edit for cue removal, image fidelity, absence of new artifacts, and preservation of medical content. See Appendices~\ref{app:image-hint-details} and~\ref{app:imagerepaircase} for details. This process yields 200 paired original/edited cases.

For natural clinical-context cues, we screen prompt-visible race/ethnicity, language, geography, migration, travel, occupation, military exposure, substance use, housing, care environment, insurance/access, disability, family/social support, sex/gender, reproductive context, and other social-determinants-of-health (SDOH) information. We retain 302 cases where the cue can be deleted, generalized, or neutralized while preserving the image, options, answer key, and core medical content.
Both intervention families are evaluated with 10 repeated runs per model. For image hints, we report $\Delta_{\mathrm{image}} = \mathrm{Acc}_{\mathrm{edited}} - \mathrm{Acc}_{\mathrm{original}}$, and for natural cues we report $\Delta_{\mathrm{natural}} = \mathrm{Acc}_{\mathrm{neutralized}} - \mathrm{Acc}_{\mathrm{original}}$. Thus, all three intervention metrics use modified minus original, and negative values indicate performance lost after cue removal or neutralization. GPT-5.4 also judges whether original-condition rationales mention or rely on the removed cue. We define \textit{bad-use} as cue use with an original-correct and edited/neutralized-wrong transition, and report the bad/cue ratio as a stricter measure of harmful reliance.

\subsection{Low-Prevalence Image-Text and Device Audits}
\label{sec:method-image-text-device}
GPT-5.4 first detects text-bearing images and option overlap, and a second adjudication identifies high-risk cases where the visible text can reveal or strongly narrow the answer.
We separately audit devices and post-treatment material, including implants, fixation hardware, tubes, shunts, embolization coils, filters, dressings, instruments, and procedure materials. These become shortcut risks only when they reveal disease, stage, prior intervention, complication, or management in a way that bypasses the intended route. Confirmed unrepairable cases are treated as exclusion candidates rather than image-editing targets.

\subsection{Human Review and Shortcut-Mitigated Subset}
\label{sec:method-cleanbench}

The review schema and summary are in Appendix~\ref{app:expert-protocol}. In sampled review, 142/150 option-form repairs were medically valid, 148/150 were improved or partially improved, all 50 sampled natural cues were confirmed and valid, and 45/46 image-hint edits were valid (Table~\ref{tab:app_expert_annotation_summary}). These samples support the retained analyses but do not replace release-scale validation.
The final shortcut-mitigated subset, \cleanbench{}, is built from MedThinkVQA, MedXpertQA-MM, and MMMU-HM. Construction combines image-availability checks, difficulty and visual-dependence scoring, shortcut filtering or repair, and removal of confirmed device/material shortcut-risk cases. We describe it as shortcut-mitigated rather than shortcut-free because residual risks and release checks remain.

\section{Experiments}

\noindent\textbf{Datasets}
The audit covers six medical multimodal MCQ datasets: AMBOSS image questions \citep{AMBOSS2026QBank}, JAMA Clinical Challenge \citep{Wang2024SemiHVision,JAMAClinicalChallenge}, MMMU Health/Medicine \citep{Yue2024MMMU}, MedThinkVQA \citep{Yao2026MedThinkVQA}, MedXpertQA-MM \citep{Zuo2025MedXpertQA}, and NEJM Image Challenge \citep{NEJMImageChallenge}. The canonical six-dataset audit contains 7,706 examples. Feature-specific analyses can have slightly different counts after option normalization or source-variant expansion; Appendix~\ref{app:count-reconciliation} keeps these bookkeeping differences separate, and Appendix~\ref{app:dataset-inventory} gives the dataset inventory.

\noindent\textbf{Models and Prompt Settings}
Analyses use overlapping panels from four model families \citep{OpenAI2026GPT54,Sellergren2025MedGemma,Sellergren2026MedGemma15,QwenTeam2026Qwen35,Bai2025Qwen3VL}. The repair, modality, image-hint, and natural-cue panels contain 17, 16, 16, and 17 configurations, respectively (19 unique overall); Appendix~\ref{app:experimental-settings} gives their family composition. GPT-5.6 Sol (\texttt{gpt-5.6-sol}) is used only for reverse-control generation and error-analysis judgments, not as an evaluated panel member \citep{OpenAI2026GPT56Sol}.
All modality diagnostics use the four prompt settings in Section~\ref{sec:method-partial-input}. Matched option-form repairs are run for 20 repeated original/edited evaluations per model. Image-hint and natural-cue interventions are run for 10 repeated paired evaluations per model. The repeat counts differ because these experiments were run as separate evaluation batches.

\begin{figure*}[t]
    \centering
    \hspace{-1.8em}
    \includegraphics[width=1.02\textwidth]{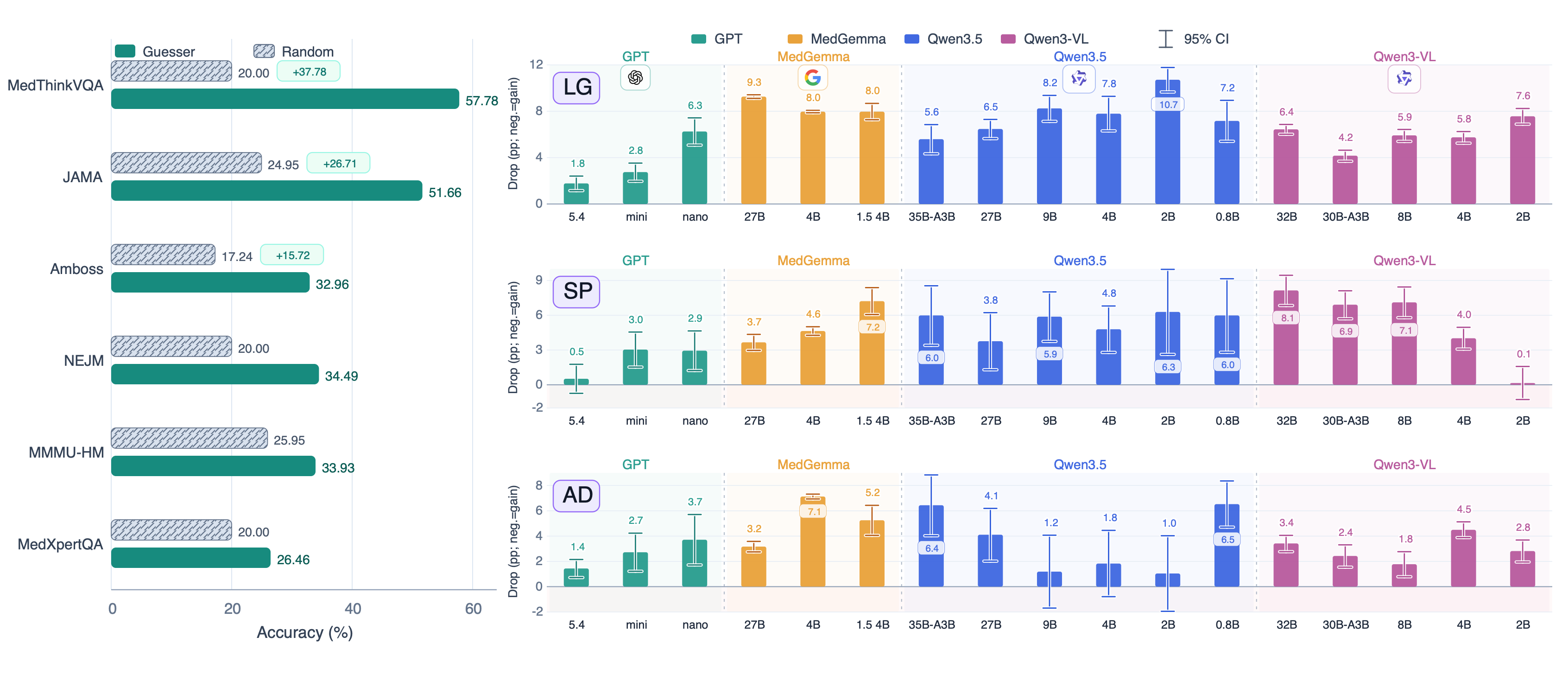}%

    \vspace{-5mm}
    \caption{\textbf{Option-form shortcuts create answer priors and are exposed by matched repair.} A trained options-only guesser exceeds the item-level random baseline on every dataset. Matched repairs for length-gap, spatial/prepositional, and absolute/conspicuous distractor cues reduce accuracy, indicating that these option-form cues are behaviorally useful. The right-hand panels visualize positive drop magnitudes for readability; signed changes in the text and Appendix use modified minus original.}
    \vspace{-3mm}
    \label{fig:option_form_shortcuts}
\end{figure*}

\noindent\textbf{Metrics and Uncertainty}
We report accuracy and accuracy change in percentage points. For no-image diagnostics, random accuracy is the mean item-level chance rate, $1/K$, where $K$ is the number of answer options. Repeated generations are averaged within item$\times$configuration$\times$condition cells before uncertainty estimation; 95\% confidence intervals use the resampling unit specified for each analysis. Expert review is reported as categorical counts and percentages. Appendix~\ref{app:experimental-settings} records the detailed settings.

\section{Results}

\paragraph{Full-Input Accuracy Mixes Multiple Shortcut Routes}
\label{sec:results-overview}
The audit identifies three recurring routes that can support a correct answer without the intended image-grounded path: answer-option form, question text plus options, and image-channel cues. These routes require different evidence. Option-form shortcuts are tested through options-only prediction and matched repairs. Text-plus-options shortcuts are tested with no-image and Question+Options probes. Image-channel shortcuts require image-side screening plus paired interventions when repair is possible. The main result is therefore route ambiguity: full-input accuracy can mix genuine image reasoning with benchmark-preserved shortcuts.

\paragraph{Option-Form Shortcuts Create Answer Priors}
\label{sec:results-option-form}

Option-form information contains predictive structure. The options-only diagnostic, which receives answer choices, deterministic option-derived metadata, and dataset/source identifiers, outperforms the item-level random baseline on every audited dataset, with the largest gaps on MedThinkVQA (57.78\% vs. 20.00\%; +37.78 pp), JAMA (51.66\% vs. 24.95\%; +26.71 pp), and AMBOSS (32.96\% vs. 17.24\%; +15.72 pp). Smaller but positive gaps appear on NEJM, MMMU-HM, and MedXpertQA-MM (Figure~\ref{fig:option_form_shortcuts}).
Matched repairs provide stronger behavioral evidence. Under the common modified-minus-original convention, length-gap, absolute/conspicuous, and spatial/prepositional repairs yield $\Delta_{\mathrm{text}}=-6.58$ pp, $-3.50$ pp, and $-4.77$ pp, respectively. These effects show that option form is not merely detectable; it can change model performance. Full model-level values are in Appendix~\ref{app:option-route-details}.

\begin{figure*}[!t]
    \centering
    \includegraphics[width=0.99\textwidth]{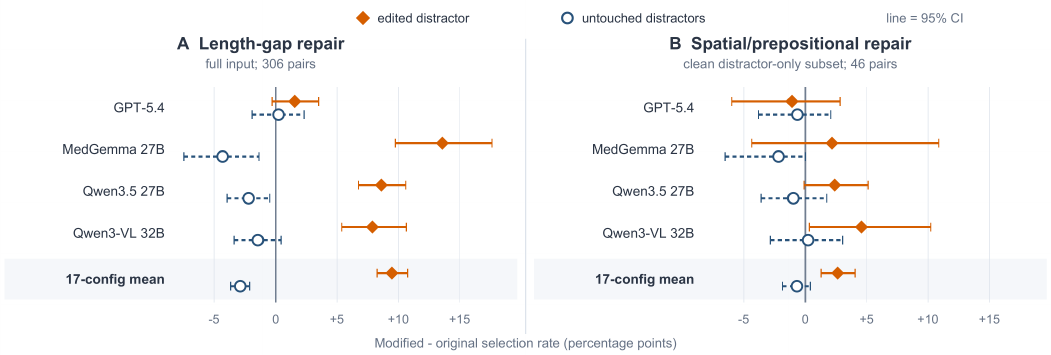}
    \vspace{-2mm}
    \caption{\textbf{On average, targeted option edits redirect predictions toward the edited distractor.} Points are modified-minus-original changes in answer-selection rate; whiskers are reported 95\% CIs. Filled diamonds and solid intervals denote the edited distractor; open circles and dashed intervals denote untouched distractors. We show one fixed representative configuration per model family, plus the equal-weight 17-configuration mean; configurations are not selected by effect size. Spatial estimates use the clean distractor-only subset and show wider configuration-level uncertainty.}
    \vspace{-4mm}
    \label{fig:length_spatial_directional}
\end{figure*}

\paragraph{Destination-Specific Evidence for Length and Spatial Cues}
\label{sec:results-option-direction}
An aggregate accuracy drop does not show whether errors increase diffusely or move toward the manipulated option. We therefore track answer destinations. Across 306 length-gap pairs, edited-distractor selection rose by 9.47 pp while selection of untouched distractors fell by 2.87 pp. In the clean 46-pair spatial/prepositional subset, where the stem, image, key, and untouched distractors remain fixed, the corresponding changes were $+2.63$ and $-0.69$ pp. After excluding edits that created a uniquely longest distractor, the spatial edited-minus-untouched contrast remained $+2.70$ pp ($N=39$). Figure~\ref{fig:length_spatial_directional} shows family-balanced model examples. These destination shifts are more specific than generic, option-identity-agnostic difficulty, although target-specific wording or plausibility changes may still contribute. The length reverse and negative controls, together with the spatial length-confound analysis, are detailed in Appendix~\ref{app:option-directional-controls}.

\begin{figure*}[ht]
    \centering
    \makebox[\textwidth][c]{%
        \hspace*{-0.0em}%
        \includegraphics[width=1.06\textwidth]{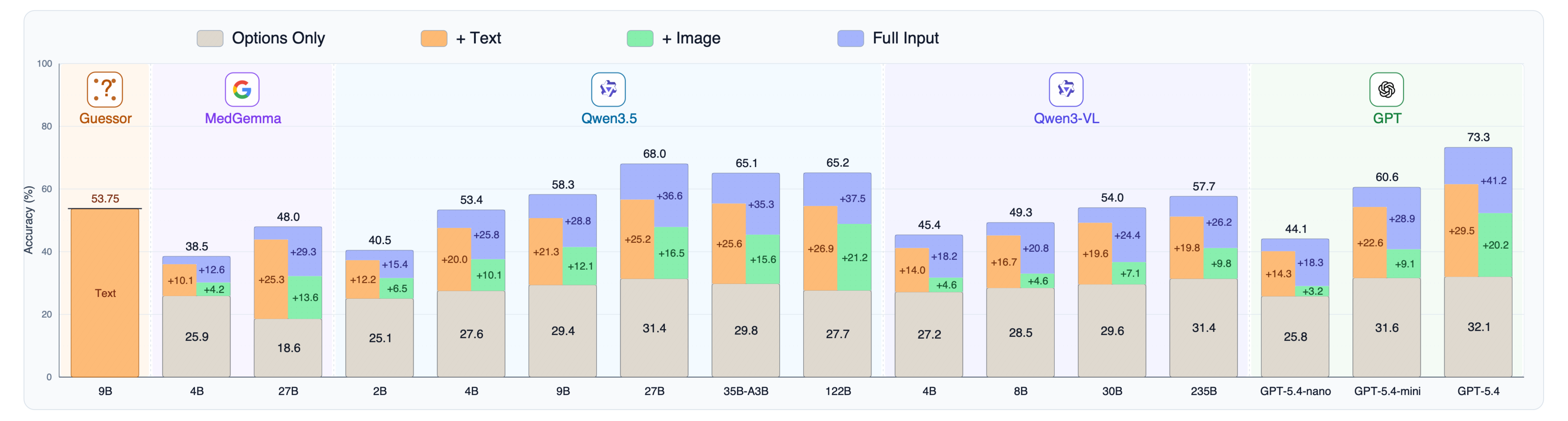}%
        \hspace*{-0.2em}%
    }
    \vspace{-8mm}
    \caption{\textbf{Question text and answer options form strong no-image routes.} Six-dataset modality ablations decompose performance into Options-only, Question+Options, Image+Options, and Full input settings. The Q+O guesser receives no image. Large gains from adding question text to answer options show that many items are partly solvable without image evidence.}
    \vspace{-3mm}
    \label{fig:modality_lattice}
\end{figure*}

\paragraph{Question Text and Options Create Strong No-Image Routes}
\label{sec:results-text-options}

The no-image route is larger than the option-form route. Across the 13-configuration open-model panel, full-input accuracy averages 62.63\%, text-only accuracy remains 53.96\%, Image+Options accuracy is 43.84\%, and options-only accuracy reaches 29.71\%. In the six-dataset modality lattice, adding the question text to options produces large gains for strong Qwen3.5 and GPT-5.4 variants, and the diagnostic Question+Options guesser reaches 53.75\% without any image.
This does not imply that images are irrelevant: full input is usually best, and Image+Options input often improves over options-only input. The validity issue is that ordinary full-input scores can mix image evidence with clinical-language priors, illness scripts, option semantics, and dataset-specific textual regularities. Numeric values underlying Figure~\ref{fig:modality_lattice} are in Appendix~\ref{app:modality-route-details}.

\begin{figure}[ht]
    \centering
    \hspace{-2.3em}
    \includegraphics[width=0.53\textwidth]{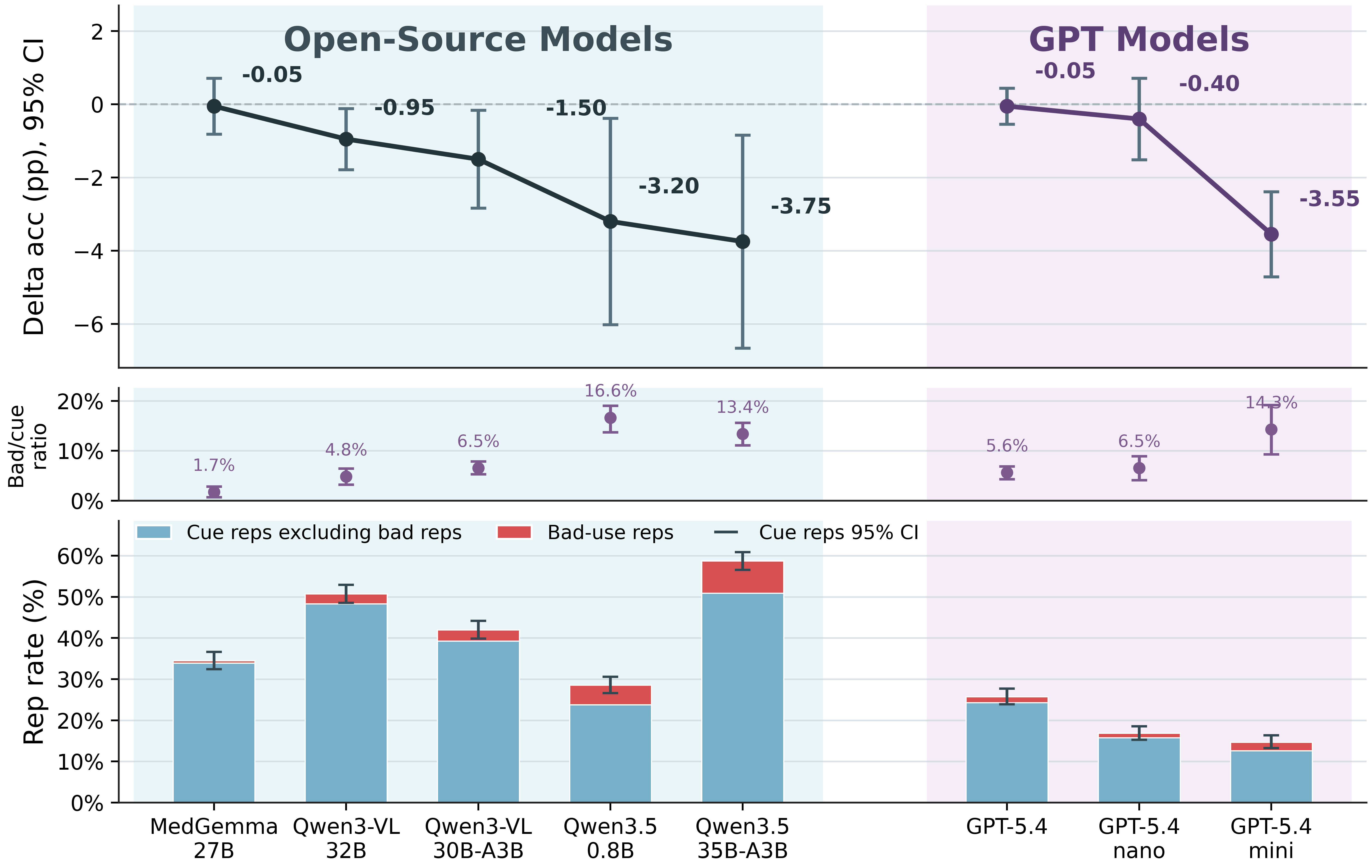}

    \caption{\textbf{Artificial visual hints separate cue awareness from harmful reliance.} Models may mention arrows, boxes, circles, or highlights without depending on them. The stricter bad/cue ratio counts only cue-use repetitions paired with an original-correct and cue-removed-wrong transition.}
    \vspace{-3mm}
    \label{fig:image_hint_bad_use_delta_acc}
\end{figure}

\paragraph{Image-Channel Cues Are Common, But Reliance Is Model-Dependent}
\label{sec:results-image-channel}

Image-side screens find many possible shortcut surfaces: 5,645 text residuals, 4,970 visible manual markups or overlays, 2,642 visible demographic/context features, and 955 device/material cues across 13,940 images. These prevalence counts establish curation risk, not model reliance. Figure~\ref{fig:image_hint_bad_use_delta_acc} instead summarizes paired image-hint evidence.
Paired interventions give more specific evidence. In the 200-case image-hint experiment, some models frequently mention annotations but show little accuracy loss after cue removal, while others show higher bad/cue conversion and larger negative changes: Qwen3.5-0.8B has $\Delta_{\mathrm{image}}=-3.20$ pp with a 16.6\% bad/cue ratio, Qwen3.5-35B-A3B has $\Delta_{\mathrm{image}}=-3.75$ pp with a 13.4\% bad/cue ratio, and GPT-5.4-mini has $\Delta_{\mathrm{image}}=-3.55$ pp with a 14.3\% bad/cue ratio. In the 302-case natural-cue experiment, the same distinction holds: Qwen3-VL-30B-A3B and MedGemma-27B mention cues often but have low bad/cue ratios, whereas Qwen3.5-0.8B and GPT-5.4 show stronger harmful conversion. Detailed image-hint and natural-cue tables are in Appendices~\ref{app:image-hint-details} and~\ref{app:natural-cue-details}.

\paragraph{Visible Image-Text Leakage Is Rare But Direct}

We audit answer-revealing text rendered inside evaluation images. The screen identifies 65 high-risk image-text shortcuts, of which only 15 are directly repairable and 14 pass model-only dataset-readiness validation. Because the repairable set is small and lacks complete before/after evaluation, we treat image-text leakage as a low-prevalence but direct data-quality risk rather than an aggregate performance result. Cases where image text explicitly reveals or strongly cues the answer should be repaired only if image validity is preserved; otherwise they should be removed or redesigned. Full audit details and repair statistics are provided in Appendix~\ref{app:image-text-leakage-details}.

\paragraph{Device and Material Cues Are Rare But Hard to Repair Safely}

We screen 1,347 candidate cases for post-treatment device or material cues and identify only 9 confirmed shortcut-risk cases, with 48 borderline cases. Thus, these cues are not a dominant aggregate artifact, but they can directly reveal prior intervention, disease stage, or management history. Because removing devices, hardware, tubes, coils, or surgical material can alter the medical image itself, we do not use this small set for a headline accuracy claim. For confirmed cases where the cue is not part of the intended construct, we prefer exclusion from the shortcut-mitigated subset rather than attempted image repair. Full details are provided in Appendix~\ref{app:device-material-details}.

\paragraph{Shortcut Mitigation Weakens Non-Visual Routes}
\label{sec:results-cleanbench}

Finally, we test whether data-centric filtering and repair can produce a more route-resistant subset. \cleanbench{} contains 1,000 examples from MedThinkVQA, MedXpertQA-MM, and MMMU-HM, retaining 22.6\% of the 4,432 complete candidate items. Under the evaluated probes, text-only accuracy falls to 5.21\% and options-only accuracy to 12.33\%, while Image+Options accuracy remains 28.47\% and full-input accuracy is 26.59\%.
This subset should be interpreted as shortcut-mitigated, not shortcut-free. The Image+Options condition exceeding full input also cautions against over-reading any single setting. The practical result is that the dominant non-visual routes are substantially weakened, while residual image-side and clinical-context risks still require documentation and clinician validation. Construction details are in Appendix~\ref{app:medqamm-details}; validation and release status are in Appendix~\ref{app:validation-release}.

\paragraph{Targeted Error Analysis of \cleanbench{}}
\label{sec:results-medqamm-errors}
Image+Options exceeds Full input by 1.88 pp, but this aggregate difference does not identify a single cause. We sampled 100 incorrect predictions from each of GPT-5.4, MedGemma-27B, and Qwen3.5-122B-A10B. GPT-5.6-Sol judged stem interference to contribute in 32/300 cases; a three-expert medical panel independently reviewed a 100-case subset, with 92\% agreement with the LLM judge and Cohen's $\kappa=0.703$. A separate 70-case development taxonomy assigns one mutually exclusive primary label per error. Visual-understanding and reasoning/integration failures total 36/70, while item/data issues are the largest single class (29/70; Table~\ref{tab:medqamm_error_analysis}). Stem interference is therefore one observed failure mode, not a complete explanation of the aggregate gap, and these sample proportions are not extrapolated to the full benchmark. Appendix~\ref{app:medqamm-error-analysis} gives the sampling and labeling protocol.

\begin{table}[H]
\centering
\small
\setlength{\tabcolsep}{3pt}
\renewcommand{\arraystretch}{1.08}
\begin{tabularx}{\columnwidth}{@{}Xrr@{}}
\toprule
Analysis and outcome & $n$ & Rate / agreement \\
\midrule
\multicolumn{3}{@{}l}{\textit{Stem-interference screen ($N=300$)}} \\
Stem interference judged present & 32 & 10.7\% \\
Stem interference judged absent & 268 & 89.3\% \\
Expert-reviewed subset (three-person panel) & 100 & 92\%; $\kappa=0.703$ \\
\midrule
\multicolumn{3}{@{}l}{\textit{Primary error taxonomy (development; $N=70$)}} \\
Visual-understanding failure & 15 & 21.4\% \\
Reasoning/integration failure & 21 & 30.0\% \\
Item/data issue & 29 & 41.4\% \\
Response/scoring failure & 3 & 4.3\% \\
Shortcut/anchoring bias & 2 & 2.9\% \\
\addlinespace
\textit{Visual or reasoning/integration subtotal} & \textit{36} & \textit{51.4\%} \\
\bottomrule
\end{tabularx}
\caption{Targeted \cleanbench{} error analyses. The 70-case development taxonomy is separate from the 300-error stem-interference screen; the 100 expert-reviewed cases are a subset of the latter. The 36/70 row is the subtotal of the first two mutually exclusive primary labels, not a third category.}
\label{tab:medqamm_error_analysis}
\end{table}

\section{From Two-Aha Framing to Route-Aware Measurement}

\paragraph{Accuracy Is a Route-Mixed Measurement}
\label{sec:discussion-route-validity}

A full-input MCQ score records final answers, not the route that produced them. Here, correct answers can arise from visual evidence, option form, no-image clinical text, visible image text, annotations, devices, or natural context. Full input often improves performance, so the image channel matters; however, strong Text-only, Options-only, or Image+Options results show that accuracy can mix image-grounded reasoning with benchmark-preserved structure. Route-level diagnostics are therefore needed to determine whether a reported gain reflects better visual reasoning or stronger use of residual cues.

\paragraph{Cue Presence Is Not Cue Dependence}
\label{sec:discussion-cue-dependence}

A detected or mentioned cue is not automatically a shortcut or harmful reliance. Paired image-hint and natural-cue interventions instead connect cue use to correctness changes after removal or neutralization. This avoids treating clinically meaningful arrows, devices, or context as invalid while still detecting construct-irrelevant dependence. Our evidence ladder---candidate cue, rationale-level use, paired behavior change, and validation---keeps the claim bounded; rationale judgments describe observable outputs, not hidden mechanisms.

\paragraph{Domain Context Is Item-Specific}
\label{sec:discussion-domain-context}

Medical context has no fixed validity status: a device, annotation, or demographic fact may be the target evidence in one item and an unintended answer route in another. Our natural-cue neutralization is therefore benchmark-validity evidence, not a general demographic-bias benchmark. Clinically necessary context should be retained and documented; nonessential predictive context should be repaired, flagged, or excluded. Model-side alignment cannot replace this item-level judgment, because the same annotation may be useful in an assisted workflow yet invalid in a benchmark intended to measure unaided image search. Explicit task definitions and data-side controls remain necessary.

\paragraph{Benchmark Repair Is Measurement Design}
\label{sec:discussion-repair}

The audit supports repair when a conservative edit can preserve medical meaning; blanket deletion would discard useful material, whereas unsafe text, image, or context edits can introduce new confounds. The workflow is therefore to detect, minimally repair when safe, evaluate matched pairs, validate, and document residual risk, discarding unsafe repairs. Releases should report Full, Options-only, no-image or Question+Options, and Image+Options results and retain provenance, prompt-visible fields, cue flags, repair and validation status, and residual risks. Accordingly, \cleanbench{} is shortcut-mitigated under the tested probes, not shortcut-free.

\section{Conclusion}
Medical multimodal MCQ benchmarks should not be treated as direct measurements of image-grounded clinical reasoning unless their items are audited for shortcut routes. We connect MCQ cueing, NLP/VQA artifacts, and medical imaging shortcut learning into one construct-validity framework; quantify shortcut prevalence and no-image signal across six datasets; show that three text shortcut families are behaviorally consequential under matched repair; and construct \cleanbench{}, a harder shortcut-mitigated subset.

\section*{Limitations}
Several parts of the audit are model-judged or rule-screened rather than exhaustively clinician-validated. The strongest behavioral claims are limited to three option-form shortcut families; image-text and device/material analyses remain primarily curation-risk evidence. Sample-based expert validation cannot guarantee every automatically screened or repaired item, and GPT-generated edits may introduce wording, fluency, plausibility, or difficulty changes beyond the targeted cue even when edits are minimal. Finally, \cleanbench{} is a shortcut-mitigated subset built from existing datasets; source-license, image-rights, item-level validation, and redistribution checks must be completed before any public release.

\section*{Ethics Statement}
This work audits benchmark reliability and is not clinical advice or a deployment evaluation. Medical image and clinical-text benchmarks can contain sensitive information; release of any repaired subset must respect source licenses, consent constraints, and privacy requirements. Sociodemographic and visible-context analyses are treated as shortcut-risk audits, not as grounds for inferring protected identity from images. Edits that alter demographic or clinical context require clinician review. No raw datasets, private images, protected health information (PHI), model outputs containing sensitive records, or non-shareable files were uploaded to ChatGPT.

\section*{Acknowledgments}

OpenAI ChatGPT was used for language polishing and figure-design assistance. OpenAI Codex assisted with code development, experiment execution, and consistency audits. The authors reviewed and verified all AI-assisted text, code, figures, and results and take full responsibility for the content, claims, and analyses.

{\sloppy\setlength{\emergencystretch}{1em}
\bibliography{custom}
}

\appendix

\clearpage

\flushbottom

\makeatletter
\setlength{\@fptop}{0pt}
\setlength{\@fpsep}{10pt plus 2pt minus 2pt}
\setlength{\@fpbot}{0pt plus 1fil}
\setlength{\@dblfptop}{0pt}
\setlength{\@dblfpsep}{10pt plus 2pt minus 2pt}
\setlength{\@dblfpbot}{0pt plus 1fil}
\makeatother

\section{Appendix Roadmap and Audit Inventory}
\label{app:count-reconciliation}

This appendix is organized by evidence route. Appendix~\ref{app:text-option-analyses} collects text- and option-side analyses; Appendix~\ref{app:image-context-analyses} covers image- and context-side evidence; Appendix~\ref{app:benchmark-validation} documents \cleanbench{} construction, validation, and release status; Appendix~\ref{app:prompts} reproduces prompts; Appendix~\ref{app:extended_shortcut_taxonomy} gives the extended taxonomy; and Appendix~\ref{app:case-studies} presents representative cases. Table~\ref{tab:app_ref_map} maps main-text claims to appendix evidence, while Table~\ref{tab:app_count_reconciliation} reconciles denominators used by the canonical audit and feature-specific analyses.

\begin{table}[H]
\centering
\footnotesize
\setlength{\tabcolsep}{4pt}
\renewcommand{\arraystretch}{1.08}
\begin{tabularx}{\columnwidth}{@{}>{\raggedright\arraybackslash}X>{\raggedright\arraybackslash}X@{}}
\toprule
Main-text claim / location & Appendix evidence \\
\midrule
Option-form answer priors (Fig.~\ref{fig:option_form_shortcuts}) & App.~\ref{app:option-route-details}; Tables~\ref{tab:app_options_guesser_by_dataset}, \ref{tab:app_option_repair_numeric} \\
Length/spatial prediction shifts (Fig.~\ref{fig:length_spatial_directional}) & App.~\ref{app:option-directional-controls}; Fig.~\ref{fig:app_length_directional_controls}; Tables~\ref{tab:app_option_directional_summary}, \ref{tab:app_length_directional_models}, \ref{tab:app_length_reverse_models}, \ref{tab:app_spatial_directional_models} \\
No-image Q+O routes (Fig.~\ref{fig:modality_lattice}) & Apps.~\ref{app:guessor-models}, \ref{app:modality-route-details}; Table~\ref{tab:app_modality_lattice_numeric} \\
Image-hint/natural-cue reliance (Fig.~\ref{fig:image_hint_bad_use_delta_acc}) & Apps.~\ref{app:image-hint-details}, \ref{app:natural-cue-details} \\
Visible text and device/material risks (results) & Apps.~\ref{app:image-text-leakage-details}, \ref{app:device-material-details} \\
Shortcut-mitigated, not shortcut-free (Sec.~\ref{sec:results-cleanbench}) & App.~\ref{app:medqamm-details}; Fig.~\ref{fig:medqamm} \\
Multiple error causes (Table~\ref{tab:medqamm_error_analysis}) & App.~\ref{app:medqamm-error-analysis} \\
Bounded validation/release claims (Sec.~\ref{sec:discussion-repair}) & App.~\ref{app:validation-release}; Tables~\ref{tab:app_expert_annotation_summary}, \ref{tab:app_validation_release_status}, \ref{tab:app_source_release_handling} \\
\bottomrule
\end{tabularx}
\caption{Reference map from main-text claims to appendix evidence. The map separates prevalence screens, paired interventions, expert review, and release validation.}
\label{tab:app_ref_map}
\end{table}

\subsection{Denominator Reconciliation}

The canonical audit denominator is used for the main results. Slightly different totals appear only in feature-specific screens that require option normalization or retain source variants.

\begin{table}[H]
\centering
\small
\setlength{\tabcolsep}{5pt}
\renewcommand{\arraystretch}{1.08}
\begin{tabular}{@{}lr@{}}
\toprule
Denominator & $N$ \\
\midrule
Canonical six-dataset audit & 7,706 \\
Length-option valid MCQ screen & 7,715 \\
Observed/source-variant audit & 7,722 \\
\bottomrule
\end{tabular}
\caption{Denominator reconciliation. The canonical audit supports the main inventory and route-level results; the 7,715-item option-valid screen supports longest-option prevalence, and the 7,722 source-variant audit supports the GPT shortcut screen.}
\label{tab:app_count_reconciliation}
\end{table}

\subsection{Dataset Inventory and Analysis Families}
\label{app:dataset-inventory}

Table~\ref{tab:dataset_inventory} lists the canonical audit inventory. Table~\ref{tab:method_analysis_families} in the main paper summarizes the corresponding evidence families and intervention sizes.

\begin{table}[H]
\centering
\footnotesize
\setlength{\tabcolsep}{3pt}
\renewcommand{\arraystretch}{1.06}
\begin{tabularx}{\columnwidth}{@{}>{\raggedright\arraybackslash}Xr>{\raggedright\arraybackslash}p{0.39\columnwidth}@{}}
\toprule
Dataset & $N$ & Benchmark role / release note \\
\midrule
AMBOSS image questions & 646 & Audit only; permission review before release \\
JAMA Clinical Challenge & 1,621 & Audit only; permission review before release \\
MMMU-HM & 1,712 & \cleanbench{} construction source \\
MedThinkVQA & 720 & \cleanbench{} construction source \\
MedXpertQA-MM & 2,000 & \cleanbench{} construction source \\
NEJM Image Challenge & 1,007 & Audit only; not used in \cleanbench{} \\
\midrule
Total & 7,706 & Canonical six-dataset audit \\
\bottomrule
\end{tabularx}
\caption{Dataset inventory. Counts refer to the canonical six-dataset audit. The final \cleanbench{} benchmark contains 1,000 examples after additional filtering and device/material cue removal; see Appendix~\ref{app:medqamm-details}.}
\label{tab:dataset_inventory}
\end{table}

\subsection{Detailed Experimental Settings}
\label{app:experimental-settings}

We evaluate four prompt variants: \textit{normal}, \textit{text-only}, \textit{options-only}, and \textit{image-options-only}, reported in the paper as Full input, Text-only, Options-only, and Image+Options. Open-model inference uses vLLM \citep{Kwon2023vLLM}. We use each model's official Hugging Face sampling configuration when available; otherwise, we use temperature 0.6, top-$p$ 0.95, and default generation settings. Broad audit summaries count parse failures as incorrect. Across 400,712 prediction rows, 392,445 are answered and 8,267 remain parse failures, giving a residual failure rate of 2.06\% and overall accuracy of 47.39\%. Controlled-repair uncertainty is computed from repeated original/edited evaluations; image-hint and natural-cue uncertainty is computed from repeated paired evaluations. Table~\ref{tab:app_panel_composition} reconciles the analysis-specific panels, whose union contains 19 unique configurations.

\begin{table}[H]
\centering
\scriptsize
\setlength{\tabcolsep}{1.8pt}
\begin{tabular}{@{}lrrrrr@{}}
\toprule
Analysis & GPT & MG & Q3.5 & Q3-VL & Total \\
\midrule
Option-form repairs & 3 & 3 & 6 & 5 & 17 \\
Modality lattice & 3 & 2 & 6 & 5 & 16 \\
Image-hint intervention & 3 & 2 & 6 & 5 & 16 \\
Natural-cue intervention & 3 & 3 & 6 & 5 & 17 \\
\bottomrule
\end{tabular}
\caption{Analysis-specific panel composition (GPT=GPT-5.4, MG=MedGemma). Panels overlap but are not identical; their union contains 19 unique configurations.}
\label{tab:app_panel_composition}
\end{table}

For the four-setting open-model panel, we first pool all 7,706 questions across the six datasets within each of the 13 model/mode configurations. Correctly salvaged parse failures are added to the canonical correct total, while unrepaired parse failures remain incorrect. We then average the 13 configuration-level accuracies with equal weight, treating thinking and nonthinking modes as separate configurations. This gives Full input 62.63\%, Text-only 53.96\%, Options-only 29.71\%, and Image+Options 43.84\%. We exclude a duplicated setting-level aggregate because one shared 3,712-row Qwen3-VL-235B result set is assigned to both MMMU-HM and MedXpertQA-MM, yielding 103,890 rows per setting instead of the canonical $13\times7{,}706=100{,}178$.

\section{Text- and Option-Side Analyses}
\label{app:text-option-analyses}

\subsection{Diagnostic Guesser Models for Shortcut Analysis}
\label{app:guessor-models}

We trained two non-visual guesser models to estimate how much benchmark performance can be obtained from textual and answer-choice artifacts alone. The options-only guesser receives no question text and no image. The Question+Options guesser receives question text and answer choices but no image pixels, captions, rationales, or explanations. For both guessers, target-domain splits are constructed at the unique-item level before prompt-view expansion; evaluation item IDs and normalized input hashes are excluded from training, yielding zero item/hash overlap. Both are evaluated by forced-choice log-probability scoring over valid answer labels only and use LoRA supervised fine-tuning \citep{Hu2022LoRA}. Tables~\ref{tab:app_guessor_settings}, \ref{tab:app_guessor_training_data}, \ref{tab:app_guessor_hparams}, and~\ref{tab:app_guessor_results} report their inputs, training data, hyperparameters, and results.

\begin{table}[H]
\centering
\footnotesize
\setlength{\tabcolsep}{3pt}
\renewcommand{\arraystretch}{1.06}
\begin{tabularx}{\columnwidth}{@{}>{\raggedright\arraybackslash}p{0.32\columnwidth}>{\raggedright\arraybackslash}X@{}}
\toprule
Model / backbone & Input fields and objective \\
\midrule
Options-only / Qwen3.5-4B & Dataset/source metadata, answer options, deterministic option metadata; LoRA SFT \\
Question+Options / Qwen3.5-9B & Question text, answer options, deterministic option metadata; LoRA SFT \\
\bottomrule
\end{tabularx}
\caption{Diagnostic guesser model settings. Both models use low-rank adaptation (LoRA) supervised fine-tuning (SFT). Deterministic option metadata includes word/character counts, length rank, gap to the second-longest option, numeric indicators, parenthetical indicators, abbreviation indicators, absolute-word indicators, and all/none/both/neither indicators.}
\label{tab:app_guessor_settings}
\end{table}

\begin{table}[tbp]
\centering
\scriptsize
\setlength{\tabcolsep}{2pt}
\begin{tabularx}{\columnwidth}{@{}Xrr@{}}
\toprule
Training source & Options records & Q+O records \\
\midrule
AMBOSS & 1,224 & 612 \\
JAMA & 3,240 & 3,240 \\
MedMCQA \citep{Pal2022MedMCQA} & 179,305 & 18,000 \\
MedQA-USMLE \citep{Jin2021MedQA} & 10,120 & 8,000 \\
MedThinkVQA & 7,347 external/train + 1,440 local & 8,067 \\
MedXpertQA-MM & 4,100 & 8,200 \\
MMMU-HM & 3,440 & 5,160 \\
NEJM & 2,068 & 2,068 \\
PMC-VQA \citep{Zhang2024PMCVQA} & 273,234 & 18,000 \\
\midrule
Total & 470,006 external + 15,512 local & 71,347 \\
\bottomrule
\end{tabularx}
\caption{Training data for the diagnostic guessers. The options-only model uses an external options-only stage followed by local target SFT; the Question+Options model uses target-heavy SFT with filtered external support data. For both models, unique-item splitting precedes four-view prompt expansion, and evaluation IDs and normalized input hashes are excluded from training, yielding zero item/hash overlap with the evaluation split.}
\label{tab:app_guessor_training_data}
\end{table}

\begin{table}[tbp]
\centering
\small
\setlength{\tabcolsep}{4pt}
\renewcommand{\arraystretch}{1.06}
\begin{tabular}{lcc}
\toprule
Setting & Options-only & Q+O \\
\midrule
Backbone & Qwen3.5-4B & Qwen3.5-9B \\
Fine-tuning & LoRA SFT & LoRA SFT \\
LoRA rank / alpha & 16 / 32 & 16 / 32 \\
LoRA dropout & 0.0 & 0.0 \\
External LR / epochs & $5\times10^{-5}$ / 1 & N/A \\
Target LR / epochs & $1\times10^{-5}$ / 4 & $8\times10^{-6}$ / 3 \\
Max sequence length & 512 target & 1536 \\
Prompt templates & 4 & 4 \\
Precision & bf16 & bf16 \\
\bottomrule
\end{tabular}
\caption{Training hyperparameters for the diagnostic guessers. The options-only external pretraining stage uses maximum sequence length 448; the local target stage uses 512.}
\label{tab:app_guessor_hparams}
\end{table}

\begin{table}[tbp]
\centering
\footnotesize
\setlength{\tabcolsep}{3pt}
\renewcommand{\arraystretch}{1.08}
\begin{tabularx}{\columnwidth}{@{}Xrr@{}}
\toprule
Dataset & Options-only & Q+O \\
\midrule
AMBOSS image questions & 32.96\% & 87.78\% \\
JAMA Clinical Challenge & 51.66\% & 70.16\% \\
MedThinkVQA & 57.78\% & 54.72\% \\
MedXpertQA-MM & 26.46\% & 34.36\% \\
MMMU Health/Medicine & 33.93\% & 47.67\% \\
NEJM Image Challenge & 34.49\% & 56.12\% \\
\midrule
Overall & 38.12\% & 53.75\% \\
\bottomrule
\end{tabularx}
\caption{Diagnostic guesser results on the 3,743-example evaluation split. Evaluation uses forced-choice continuation log probabilities for labels \texttt{[[A]]}--\texttt{[[E]]}, excluding invalid labels for four-option questions, with a four-template ensemble and label-wise z-score aggregation.}
\label{tab:app_guessor_results}
\end{table}

\FloatBarrier
\subsection{Option-Form Repair Set Construction and Results}
\label{app:option-route-details}

\subsubsection{Length-Gap Repair Set}
\label{app:length-gap-details}

Table~\ref{tab:app_length_gap_construction} reports the feature-specific screen and the retained intervention set.

\begin{table}[H]
\centering
\scriptsize
\setlength{\tabcolsep}{2.4pt}
\renewcommand{\arraystretch}{1.04}
\textit{(a) Prevalence screen}\\[2pt]
\begin{tabular}{@{}lrrrr@{}}
\toprule
Dataset & Valid & Random & Correct longest & Large-gap \\
\midrule
AMBOSS & 649 & 17.24\% & 181 (27.9\%) & 3 (0.5\%) \\
JAMA & 1,621 & 24.95\% & 553 (34.1\%) & 55 (3.4\%) \\
MMMU-HM & 1,718 & 25.95\% & 840 (48.9\%) & 11 (0.6\%) \\
MedThinkVQA & 720 & 20.00\% & 300 (41.7\%) & 44 (6.1\%) \\
MedXpertQA-MM & 2,000 & 20.00\% & 598 (29.9\%) & 9 (0.5\%) \\
NEJM & 1,007 & 20.00\% & 251 (24.9\%) & 2 (0.2\%) \\
\midrule
Total & 7,715 & 22.13\% & 2,723 (35.3\%) & 124 (1.6\%) \\
\bottomrule
\end{tabular}
\par\smallskip
\textit{(b) Intervention funnel}\\[2pt]
\begin{tabular}{@{}lrrr@{}}
\toprule
Dataset & Unique-longest correct & Rewrite & Safe edits \\
\midrule
AMBOSS & 86 (13.3\%) & 0 & 0 \\
JAMA & 520 (32.1\%) & 150 & 130 \\
MMMU-HM & 260 (15.1\%) & 0 & 0 \\
MedThinkVQA & 277 (38.5\%) & 180 & 139 \\
MedXpertQA-MM & 296 (14.8\%) & 0 & 0 \\
NEJM & 225 (22.3\%) & 45 & 37 \\
\midrule
Total & 1,664 (21.6\%) & 375 & 306 \\
\bottomrule
\end{tabular}
\caption{Length-gap prevalence and intervention construction. The large-gap descriptive rule requires the correct option to be uniquely longest and at least 30 characters and 4 words longer than the next-longest distractor. The intervention uses rank-based selection in JAMA, MedThinkVQA, and NEJM, followed by GPT-5.4 rewriting and validation.}
\label{tab:app_length_gap_construction}
\end{table}

A rewrite was accepted only if it changed one or two distractors, left the correct option unchanged, preserved the number and order of options, introduced no duplicate option text, and made at least one distractor at least 5 visible characters longer than the correct option. The final 306 pairs were used for original-vs-edited evaluation with 20 repeated runs per condition.

\subsubsection{Spatial/Prepositional Repair Set}
\label{app:spatial-prep-details}

Tables~\ref{tab:app_spatial_prep_construction} and~\ref{tab:app_spatial_prep_subtypes} report the candidate funnel and subtype composition.

\begin{table}[H]
\centering
\scriptsize
\setlength{\tabcolsep}{2.4pt}
\renewcommand{\arraystretch}{1.04}
\begin{tabularx}{\columnwidth}{@{}>{\raggedright\arraybackslash}Xrrrr@{}}
\toprule
Dataset & Broad & Strict & Reviewed & Repaired \\
\midrule
JAMA & 108 & 43 & 67 & 48 \\
MedXpertQA-MM & 47 & 34 & 48 & 47 \\
MedThinkVQA & 107 & 14 & 23 & 0 \\
MMMU-HM & 24 & 14 & 21 & 0 \\
AMBOSS & 8 & 9 & 9 & 0 \\
NEJM & 22 & 7 & 6 & 6 \\
\midrule
Total & 316 & 121 & 174 & 101 \\
\bottomrule
\end{tabularx}
\caption{Spatial/prepositional shortcut construction. The broad and strict screens are descriptive diagnostics. The main intervention starts from a semantically reviewed 174-item candidate pool and retains 101 items after medical and semantic validation and evaluation-eligibility checks.}
\label{tab:app_spatial_prep_construction}
\end{table}

\begin{table}[H]
\centering
\footnotesize
\setlength{\tabcolsep}{3pt}
\renewcommand{\arraystretch}{1.08}
\begin{tabular}{lr}
\toprule
Candidate subtype & Count in 101-item set \\
\midrule
Ambiguous preposition wording & 34 \\
Other preposition wording & 31 \\
Real spatial/anatomical relation & 12 \\
Specimen/source wording & 9 \\
Weak management wording & 7 \\
Strong spatial head & 5 \\
Procedure-context wording & 2 \\
Range/threshold wording & 1 \\
\bottomrule
\end{tabular}
\caption{Composition of the spatial/prepositional repaired evaluation set. Categories describe the intervention set; downstream evaluation compares paired original and edited items without conditioning on these categories.}
\label{tab:app_spatial_prep_subtypes}
\end{table}

\subsubsection{Absolute/Conspicuous Distractor Repair Set}
\label{app:absolute-distractor-details}

Tables~\ref{tab:app_absdist_construction} and~\ref{tab:app_absdist_cue_types} report the construction stages and non-mutually-exclusive cue categories.

\begin{table}[H]
\centering
\footnotesize
\setlength{\tabcolsep}{3pt}
\renewcommand{\arraystretch}{1.08}
\begin{tabularx}{\columnwidth}{@{}XrX@{}}
\toprule
Stage & $N$ & Notes \\
\midrule
Balanced candidate set & 120 & Selected across datasets and cue types \\
Valid repairs & 101 & Option-level patches; gold answer preserved; used for downstream evaluation \\
\bottomrule
\end{tabularx}
\caption{Construction of the absolute/conspicuous distractor evaluation set.}
\label{tab:app_absdist_construction}
\end{table}

\begin{table}[H]
\centering
\footnotesize
\setlength{\tabcolsep}{3pt}
\renewcommand{\arraystretch}{1.08}
\begin{tabularx}{\columnwidth}{@{}Xrr@{}}
\toprule
Cue category & Candidate set & Valid repairs \\
\midrule
All / none / neither scope & 45 & 42 \\
Only / singleton scope & 34 & 32 \\
Rule-state language & 24 & 20 \\
Strict modal absolute & 17 & 16 \\
Required / necessary language & 7 & 0 \\
Negative exception scope & 2 & 2 \\
Unknown / no target category & 3 & 0 \\
\bottomrule
\end{tabularx}
\caption{Cue-category incidence counts for absolute and conspicuous distractor candidates. Categories are not mutually exclusive.}
\label{tab:app_absdist_cue_types}
\end{table}

\subsubsection{Cross-Dataset and Cross-Family Summary}

The options-only diagnostic is reported by dataset, followed by the complete fixed-panel accuracy changes for all three repair families.

\begin{table}[H]
\centering
\footnotesize
\setlength{\tabcolsep}{3pt}
\renewcommand{\arraystretch}{1.05}
\begin{tabularx}{\columnwidth}{@{}>{\raggedright\arraybackslash}Xrrr@{}}
\toprule
Dataset & Random & Options-only guesser & Gain \\
\midrule
AMBOSS & 17.24\% & 32.96\% & +15.72 pp \\
JAMA & 24.95\% & 51.66\% & +26.71 pp \\
MedThinkVQA & 20.00\% & 57.78\% & +37.78 pp \\
MedXpertQA-MM & 20.00\% & 26.46\% & +6.46 pp \\
MMMU-HM & 25.95\% & 33.93\% & +7.98 pp \\
NEJM & 20.00\% & 34.49\% & +14.49 pp \\
\midrule
Overall & 22.13\% & 38.12\% & +15.99 pp \\
\bottomrule
\end{tabularx}
\caption{Dataset-level options-only diagnostic results on the 3,743-example evaluation split. The diagnostic receives answer choices, deterministic option-derived metadata, and dataset/source identifiers, with no question stem or image. Random accuracy is computed from each item's number of answer options. Gains are diagnostic accuracy minus random baseline.}
\label{tab:app_options_guesser_by_dataset}
\end{table}

\begin{table}[H]
\centering
\scriptsize
\setlength{\tabcolsep}{2.2pt}
\renewcommand{\arraystretch}{1.03}
\begin{tabular}{@{}llrrr@{}}
\toprule
Family & Model & Len. & Spatial & Absolute \\
\midrule
GPT & GPT-5.4 & -1.8 & -0.5 & -1.4 \\
GPT & GPT-5.4-mini & -2.8 & -3.0 & -2.7 \\
GPT & GPT-5.4-nano & -6.3 & -2.9 & -3.7 \\
\midrule
MedGemma & 27B & -9.3 & -3.7 & -3.2 \\
MedGemma & 4B & -8.0 & -4.6 & -7.1 \\
MedGemma & 1.5-4B & -8.0 & -7.2 & -5.2 \\
\midrule
Qwen3.5 & 35B-A3B & -5.6 & -6.0 & -6.4 \\
Qwen3.5 & 27B & -6.5 & -3.8 & -4.1 \\
Qwen3.5 & 9B & -8.2 & -5.9 & -1.2 \\
Qwen3.5 & 4B & -7.8 & -4.8 & -1.8 \\
Qwen3.5 & 2B & -10.7 & -6.3 & -1.0 \\
Qwen3.5 & 0.8B & -7.2 & -6.0 & -6.5 \\
\midrule
Qwen3-VL & 32B & -6.4 & -8.1 & -3.4 \\
Qwen3-VL & 30B-A3B & -4.2 & -6.9 & -2.4 \\
Qwen3-VL & 8B & -5.9 & -7.1 & -1.8 \\
Qwen3-VL & 4B & -5.8 & -4.0 & -4.5 \\
Qwen3-VL & 2B & -7.6 & -0.1 & -2.8 \\
\bottomrule
\end{tabular}
\caption{Complete model-level matched-repair results for three option-form shortcut families. Values are signed accuracy changes in percentage points, computed as modified minus original; negative values indicate that removing the cue reduced accuracy.}
\label{tab:app_option_repair_numeric}
\end{table}

\FloatBarrier
\subsection{Directional and Control Analyses for Option-Form Repairs}
\label{app:option-directional-controls}

We report modified-minus-original selection shifts for edited and untouched distractors; these quantities need not form a closed decomposition because keyed and other choices can also absorb probability mass. Table~\ref{tab:app_option_directional_summary} summarizes the controls, and Tables~\ref{tab:app_length_directional_models}--\ref{tab:app_spatial_directional_models} report the fixed 17-configuration panel. Some aggregate-accuracy columns use the corresponding parser-oracle convention whereas destination/control columns use symmetric fixed parsers, so they are supporting analyses rather than a single-parser decomposition.

\begin{figure*}[t]
    \centering
    \includegraphics[width=0.98\textwidth]{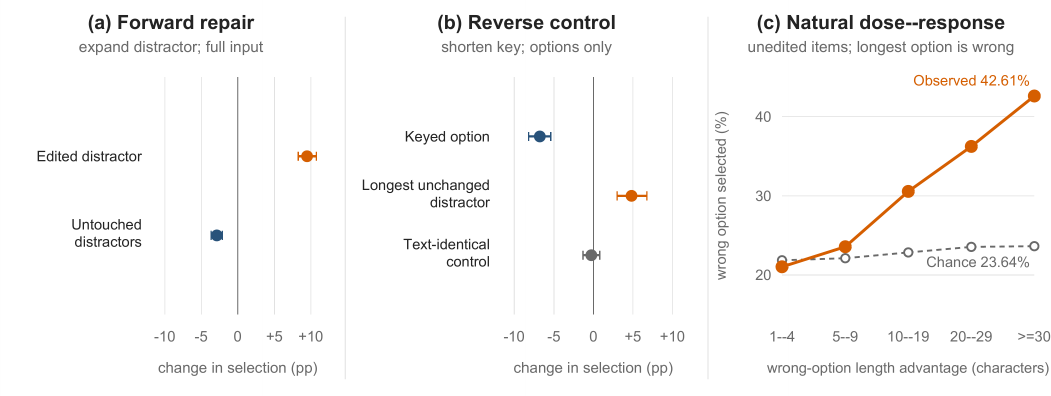}
    \caption{\textbf{Length-gap evidence across intervention directions and a natural margin screen.} Forward full-input edits increase selection of the expanded distractor; the separate Options-only reverse decreases key selection and increases selection of the longest unchanged distractor in the target-reaching subset. The 21-item text-identical control, available for a 15-configuration Options-only panel, remains near zero. On unedited records where the longest option is incorrect, its observed selection rate rises with its character-length advantage while item-level chance remains approximately flat. The dose-response analysis reports record counts rather than a verified unique-item denominator.}
    \label{fig:app_length_directional_controls}
\end{figure*}

\paragraph{Length-gap controls.}
The forward intervention expands one or two distractors; the GPT-5.6-Sol reverse control shortens only the key, with other item fields fixed. The 71 target-reaching cases and 21 text-identical controls are subsets of the 306 reverse pairs. Figure~\ref{fig:app_length_directional_controls} adds a natural margin screen whose bins count records, not verified unique items.

\paragraph{Spatial/prepositional control.}
Among 101 repaired pairs, 46 form a clean distractor-only subset with stem, image, key, and untouched distractors byte-identical. Excluding seven newly unique-longest edits leaves a 39-case forward length-control subset; this is not a spatial reverse intervention.

\paragraph{Absolute/conspicuous control.}
Table~\ref{tab:app_absolute_directional_models} records supporting destination and Options-only results; these are not a new main-text experiment.

\begin{table*}[t]
\centering
\footnotesize
\setlength{\tabcolsep}{4pt}
\renewcommand{\arraystretch}{1.08}
\begin{tabularx}{\textwidth}{@{}>{\raggedright\arraybackslash}p{0.18\textwidth}>{\raggedright\arraybackslash}p{0.18\textwidth}r>{\raggedright\arraybackslash}X@{}}
\toprule
Family / analysis & Input or subset & $N$ & Change in selection rate or accuracy, pp [95\% CI] \\
\midrule
Length aggregate & Full input & 306 & Accuracy $-6.58$ [$-7.60,-5.59$] \\
Length forward destination & Full input & 306 & Edited distractor $+9.47$ [$8.26,10.75$]; untouched distractors $-2.87$ [$-3.66,-2.11$] \\
Length reverse & Options-only & 306 & Keyed option $-6.79$ [$-8.21,-5.41$] \\
Length target-reaching & Reverse subset & 71 & Longest unchanged distractor $+4.82$ [$2.99,6.77$] \\
Length negative control & Options-only (15 configs); text-identical reverse subset & 21 & Keyed-option selection $-0.27$ [$-1.32,0.79$] \\
\midrule
Spatial aggregate & Full input & 101 & Accuracy $-4.77$ [$-5.86,-3.70$] \\
Spatial forward destination & Clean distractor-only subset & 46 & Edited distractor $+2.63$ [$1.28,4.05$]; untouched distractors $-0.69$ [$-1.85,0.40$] \\
Spatial length control & No new unique-longest distractor & 39 & Edited-minus-untouched contrast $+2.70$ [$0.49,4.99$] \\
\midrule
Absolute forward destination & Full input & 101 & Target distractor $+4.07$ [$1.60,6.35$]; other distractors $-1.46$ [$-2.84,0.00$] \\
Absolute control & Options-only & 101 & Key $-3.01$ [$-5.14,-0.96$]; target $+5.10$ [$2.29,8.03$]; other $-2.12$ [$-3.94,-0.32$] \\
\bottomrule
\end{tabularx}
\caption{Aggregate directional and control results for the option-form repairs. Modified minus original is used throughout. The spatial 39-case analysis is nested within the 46-case clean subset, which is nested within the 101-pair repair set. Length target-reaching and text-identical rows are subsets of the 306 reverse pairs; the text-identical negative control uses the separately available 15-configuration Options-only panel.}
\label{tab:app_option_directional_summary}
\end{table*}

\begin{table*}[t]
\centering
\footnotesize
\setlength{\tabcolsep}{3pt}
\renewcommand{\arraystretch}{1.05}
\begin{adjustbox}{max width=\textwidth}
\begin{tabular}{lllll}
\toprule
Family & Configuration & Accuracy change & Edited distractor & Untouched distractors \\
\midrule
GPT & GPT-5.4 & $-1.78$ [$-4.33,0.78$] & $+1.55$ [$-0.29,3.50$] & $+0.23$ [$-1.90,2.32$] \\
GPT & GPT-5.4-mini & $-2.75$ [$-4.98,-0.62$] & $+4.77$ [$2.86,6.81$] & $-2.03$ [$-3.74,-0.33$] \\
GPT & GPT-5.4-nano & $-6.26$ [$-8.89,-3.73$] & $+8.58$ [$6.01,11.31$] & $-2.32$ [$-4.49,-0.20$] \\
MedGemma & 27B & $-9.26$ [$-13.07,-5.60$] & $+13.58$ [$9.75,17.63$] & $-4.31$ [$-7.47,-1.34$] \\
MedGemma & 1.5-4B & $-7.97$ [$-12.14,-3.89$] & $+14.38$ [$10.38,18.50$] & $-6.24$ [$-9.43,-3.17$] \\
MedGemma & 4B & $-7.97$ [$-11.93,-4.22$] & $+9.64$ [$5.65,13.76$] & $-1.67$ [$-4.93,1.57$] \\
Qwen3.5 & 35B-A3B & $-5.59$ [$-7.11,-4.10$] & $+8.02$ [$6.31,9.85$] & $-2.37$ [$-3.73,-1.06$] \\
Qwen3.5 & 27B & $-6.47$ [$-8.28,-4.69$] & $+8.61$ [$6.75,10.60$] & $-2.19$ [$-3.94,-0.49$] \\
Qwen3.5 & 9B & $-8.25$ [$-10.10,-6.47$] & $+9.64$ [$7.91,11.42$] & $-1.31$ [$-2.81,0.18$] \\
Qwen3.5 & 4B & $-7.79$ [$-9.53,-6.06$] & $+10.59$ [$8.59,12.61$] & $-2.68$ [$-4.38,-1.01$] \\
Qwen3.5 & 2B & $-10.72$ [$-12.79,-8.69$] & $+14.12$ [$11.93,16.31$] & $-3.53$ [$-5.42,-1.65$] \\
Qwen3.5 & 0.8B & $-7.17$ [$-8.81,-5.56$] & $+11.98$ [$10.26,13.71$] & $-4.61$ [$-6.29,-2.92$] \\
Qwen3-VL & 32B & $-6.44$ [$-9.26,-3.79$] & $+7.88$ [$5.38,10.65$] & $-1.44$ [$-3.37,0.44$] \\
Qwen3-VL & 30B-A3B & $-4.17$ [$-6.09,-2.39$] & $+6.73$ [$4.46,9.23$] & $-2.57$ [$-4.49,-0.80$] \\
Qwen3-VL & 8B & $-5.92$ [$-9.07,-2.92$] & $+8.58$ [$5.38,12.01$] & $-2.66$ [$-4.92,-0.54$] \\
Qwen3-VL & 4B & $-5.75$ [$-8.35,-3.35$] & $+7.58$ [$4.66,10.70$] & $-1.83$ [$-3.86,0.03$] \\
Qwen3-VL & 2B & $-7.57$ [$-10.23,-5.10$] & $+14.82$ [$11.57,18.22$] & $-7.25$ [$-10.23,-4.43$] \\
\midrule
All & 17-config mean & $-6.58$ [$-7.60,-5.59$] & $+9.47$ [$8.26,10.75$] & $-2.87$ [$-3.66,-2.11$] \\
\bottomrule
\end{tabular}
\end{adjustbox}
\caption{Complete 17-configuration length forward results on all 306 pairs. Values are percentage-point changes with 95\% CIs. The final row is an equal-weight configuration mean, not a random-effects estimate; intervals crossing zero are retained.}
\label{tab:app_length_directional_models}
\end{table*}

\begin{table*}[t]
\centering
\footnotesize
\setlength{\tabcolsep}{3pt}
\renewcommand{\arraystretch}{1.05}
\begin{tabular}{llll}
\toprule
Family & Configuration & Reverse key & Longest unchanged distractor ($N=71$) \\
\midrule
GPT & GPT-5.4 & $-3.71$ [$-5.93,-1.50$] & $+1.90$ [$-2.25,6.13$] \\
GPT & GPT-5.4-mini & $-4.44$ [$-6.76,-2.14$] & $+2.18$ [$-1.83,6.27$] \\
GPT & GPT-5.4-nano & $-6.14$ [$-8.74,-3.56$] & $+7.96$ [$3.45,12.75$] \\
MedGemma & 27B & $-13.25$ [$-18.37,-8.20$] & $+12.54$ [$4.23,20.99$] \\
MedGemma & 1.5-4B & $-6.57$ [$-10.46,-2.81$] & $0.00$ [$-6.83,6.90$] \\
MedGemma & 4B & $-7.39$ [$-12.27,-2.73$] & $+1.34$ [$-7.04,9.72$] \\
Qwen3.5 & 35B-A3B & $-4.72$ [$-6.85,-2.63$] & $+2.46$ [$-0.99,6.13$] \\
Qwen3.5 & 27B & $-4.61$ [$-6.93,-2.35$] & $+0.56$ [$-2.75,3.87$] \\
Qwen3.5 & 9B & $-6.31$ [$-8.50,-4.18$] & $+3.73$ [$0.49,7.04$] \\
Qwen3.5 & 4B & $-7.16$ [$-9.20,-5.15$] & $+4.08$ [$1.48,6.76$] \\
Qwen3.5 & 2B & $-6.03$ [$-7.91,-4.13$] & $+3.03$ [$-0.42,6.55$] \\
Qwen3.5 & 0.8B & $-4.67$ [$-6.63,-2.73$] & $+5.99$ [$2.68,9.37$] \\
Qwen3-VL & 32B & $-9.28$ [$-12.61,-6.06$] & $+5.92$ [$1.55,11.34$] \\
Qwen3-VL & 30B-A3B & $-6.39$ [$-9.59,-3.24$] & $+0.99$ [$-3.31,5.00$] \\
Qwen3-VL & 8B & $-7.61$ [$-11.32,-3.99$] & $+13.87$ [$6.48,21.83$] \\
Qwen3-VL & 4B & $-7.63$ [$-11.26,-4.04$] & $+7.89$ [$2.46,14.23$] \\
Qwen3-VL & 2B & $-9.46$ [$-12.97,-6.05$] & $+7.54$ [$0.92,14.51$] \\
\midrule
All & 17-config mean & $-6.79$ [$-8.21,-5.41$] & $+4.82$ [$2.99,6.77$] \\
\bottomrule
\end{tabular}
\caption{Complete 17-configuration length reverse results in the Options-only setting. The keyed-option column uses all 306 reverse pairs; the final column uses the 71 target-reaching cases in which the key lost unique-longest status. Intervals crossing zero are retained.}
\label{tab:app_length_reverse_models}
\end{table*}

\begin{table*}[t]
\centering
\footnotesize
\setlength{\tabcolsep}{3.5pt}
\renewcommand{\arraystretch}{1.03}
\textit{(a) Full-input results}\\[2pt]
\begin{tabular}{lccc}
\toprule
Configuration & Accuracy & Target distractor & Other distractors \\
\midrule
GPT-5.4 & $-1.44$ [$-6.04,2.97$] & $+3.02$ [$-1.19,7.48$] & $-1.58$ [$-5.64,2.33$] \\
GPT-5.4-mini & $-2.72$ [$-5.89,0.30$] & $+1.68$ [$-1.73,4.95$] & $+0.99$ [$-1.68,4.06$] \\
GPT-5.4-nano & $-3.71$ [$-9.26,1.88$] & $+2.87$ [$-3.12,8.86$] & $+0.69$ [$-4.90,6.19$] \\
MedGemma-27B & $-3.17$ [$-8.27,1.73$] & $+7.72$ [$2.18,13.76$] & $-5.25$ [$-10.25,-0.69$] \\
MedGemma-1.5-4B & $-5.25$ [$-11.44,0.64$] & $+6.19$ [$1.09,11.68$] & $-1.49$ [$-7.03,3.71$] \\
MedGemma-4B & $-7.13$ [$-14.85,0.00$] & $+12.87$ [$3.96,21.78$] & $-5.74$ [$-12.67,0.40$] \\
Qwen3.5-35B-A3B & $-6.44$ [$-9.01,-3.96$] & $+2.03$ [$-0.20,4.26$] & $+0.15$ [$-1.78,2.08$] \\
Qwen3.5-27B & $-4.11$ [$-7.48,-0.64$] & $+3.66$ [$0.50,6.68$] & $-0.74$ [$-3.12,1.73$] \\
Qwen3.5-9B & $-1.19$ [$-4.01,1.53$] & $+1.78$ [$-1.88,5.25$] & $-1.98$ [$-4.60,1.14$] \\
Qwen3.5-4B & $-1.83$ [$-5.94,2.57$] & $+1.04$ [$-2.82,4.80$] & $-0.50$ [$-4.01,3.07$] \\
Qwen3.5-2B & $-1.04$ [$-3.96,1.98$] & $+1.73$ [$-2.33,5.64$] & $-0.69$ [$-3.71,2.48$] \\
Qwen3.5-0.8B & $-6.53$ [$-9.60,-3.56$] & $+4.26$ [$0.74,7.77$] & $-3.17$ [$-6.63,0.15$] \\
Qwen3-VL-32B & $-3.42$ [$-8.37,1.04$] & $+3.47$ [$0.00,7.52$] & $-0.05$ [$-4.46,4.46$] \\
Qwen3-VL-30B-A3B & $-2.43$ [$-6.29,1.39$] & $+3.37$ [$-0.69,7.62$] & $-0.94$ [$-3.51,1.09$] \\
Qwen3-VL-8B & $-1.78$ [$-6.78,3.12$] & $+2.77$ [$-2.33,8.27$] & $-1.04$ [$-5.20,2.87$] \\
Qwen3-VL-4B & $-4.50$ [$-11.04,1.83$] & $+5.54$ [$-0.84,12.03$] & $-1.19$ [$-6.24,3.76$] \\
Qwen3-VL-2B & $-2.82$ [$-8.86,2.92$] & $+5.10$ [$-2.03,12.23$] & $-2.28$ [$-8.22,3.91$] \\
\midrule
17-config mean & $-3.50$ [$-5.18,-1.79$] & $+4.07$ [$1.60,6.35$] & $-1.46$ [$-2.84,0.00$] \\
\bottomrule
\end{tabular}
\par\medskip
\textit{(b) Options-only control}\\[2pt]
\begin{tabular}{lccc}
\toprule
Configuration & Key & Target distractor & Other distractors \\
\midrule
GPT-5.4 & $-1.98$ [$-7.48,3.32$] & $+2.18$ [$-3.86,8.47$] & $-0.20$ [$-5.20,4.80$] \\
GPT-5.4-mini & $-0.99$ [$-6.29,4.21$] & $+4.95$ [$-1.19,11.29$] & $-3.96$ [$-9.16,1.14$] \\
GPT-5.4-nano & $+3.12$ [$-2.33,8.47$] & $-3.81$ [$-10.50,2.82$] & $+0.69$ [$-4.31,5.94$] \\
MedGemma-27B & $-5.94$ [$-12.87,0.99$] & $+11.88$ [$4.95,19.80$] & $-5.94$ [$-10.89,-0.99$] \\
MedGemma-1.5-4B & $-10.89$ [$-19.80,-1.98$] & $+18.81$ [$9.90,27.72$] & $-7.92$ [$-15.84,-0.99$] \\
MedGemma-4B & $-5.94$ [$-14.85,1.98$] & $+13.86$ [$4.95,22.77$] & $-7.92$ [$-15.84,-0.99$] \\
Qwen3.5-35B-A3B & $-3.42$ [$-7.52,0.60$] & $+2.57$ [$-2.48,7.87$] & $+0.84$ [$-2.72,4.26$] \\
Qwen3.5-27B & $-0.84$ [$-4.11,2.52$] & $+1.73$ [$-3.12,6.63$] & $-1.09$ [$-5.20,3.07$] \\
Qwen3.5-9B & $-1.34$ [$-5.30,2.43$] & $-1.88$ [$-7.18,3.37$] & $+2.92$ [$-0.40,6.24$] \\
Qwen3.5-4B & $+0.40$ [$-3.42,4.11$] & $-1.14$ [$-5.89,3.66$] & $+0.74$ [$-3.17,4.75$] \\
Qwen3.5-2B & $-1.09$ [$-3.86,1.68$] & $+2.13$ [$-1.63,6.04$] & $-1.09$ [$-3.86,1.63$] \\
Qwen3.5-0.8B & $-4.06$ [$-6.88,-1.09$] & $+7.97$ [$4.36,11.58$] & $-4.06$ [$-7.48,-0.89$] \\
Qwen3-VL-32B & $-1.19$ [$-6.78,4.16$] & $+5.50$ [$-1.39,12.62$] & $-4.31$ [$-10.35,1.53$] \\
Qwen3-VL-30B-A3B & $-6.44$ [$-11.58,-1.73$] & $+6.68$ [$1.14,11.98$] & $-0.25$ [$-4.41,4.11$] \\
Qwen3-VL-8B & $-0.74$ [$-6.93,5.59$] & $+1.78$ [$-4.95,8.76$] & $-1.04$ [$-6.44,4.21$] \\
Qwen3-VL-4B & $-4.75$ [$-12.18,2.52$] & $+2.52$ [$-5.79,10.99$] & $+2.23$ [$-3.27,7.72$] \\
Qwen3-VL-2B & $-5.15$ [$-9.80,-0.69$] & $+10.89$ [$3.37,18.42$] & $-5.74$ [$-12.58,0.99$] \\
\midrule
17-config mean & $-3.01$ [$-5.14,-0.96$] & $+5.10$ [$2.29,8.03$] & $-2.12$ [$-3.94,-0.32$] \\
\bottomrule
\end{tabular}
\caption{Complete 17-configuration absolute/conspicuous results on all 101 pairs. Panel~(a) reports full-input accuracy and destinations; panel~(b) reports the Options-only control. Full-input accuracy follows the reported 3.50-pp parser-oracle convention, while all destination and Options-only columns use symmetric fixed parsers. The columns are not a closed single-parser decomposition, and intervals crossing zero are retained.}
\label{tab:app_absolute_directional_models}
\end{table*}

\twocolumn[{
\begin{minipage}{\textwidth}
\centering
\footnotesize
\setlength{\tabcolsep}{2.5pt}
\renewcommand{\arraystretch}{1.05}
\begin{adjustbox}{max width=\textwidth}
\begin{tabular}{llllll}
\toprule
Family & Configuration & Accuracy change & Edited distractor & Untouched distractors & Length-controlled contrast \\
\midrule
GPT & GPT-5.4 & $-0.52$ [$-4.18,3.14$] & $-1.09$ [$-5.98,2.83$] & $-0.65$ [$-3.80,2.07$] & $-0.90$ [$-7.18,4.74$] \\
GPT & GPT-5.4-mini & $-3.04$ [$-6.08,-0.05$] & $+2.72$ [$0.00,5.43$] & $+0.54$ [$-2.07,3.15$] & $+1.41$ [$-3.46,6.54$] \\
GPT & GPT-5.4-nano & $-2.94$ [$-6.44,0.31$] & $-0.87$ [$-7.07,5.43$] & $0.00$ [$-4.35,4.24$] & $-2.56$ [$-13.21,7.82$] \\
MedGemma & 27B & $-3.66$ [$-8.51,0.88$] & $+2.17$ [$-4.35,10.87$] & $-2.17$ [$-6.52,0.00$] & $0.00$ [$-7.69,7.69$] \\
MedGemma & 1.5-4B & $-7.22$ [$-12.73,-1.75$] & $+1.63$ [$-7.72,11.52$] & $-4.78$ [$-12.83,2.07$] & $+5.90$ [$-6.79,21.41$] \\
MedGemma & 4B & $-4.64$ [$-10.31,0.72$] & $0.00$ [$-8.70,8.70$] & $0.00$ [$-6.52,6.52$] & $0.00$ [$-15.38,15.38$] \\
Qwen3.5 & 35B-A3B & $-5.98$ [$-8.45,-3.56$] & $+0.11$ [$-2.28,2.50$] & $+1.52$ [$-1.85,5.00$] & $0.00$ [$-5.13,5.00$] \\
Qwen3.5 & 27B & $-3.76$ [$-6.13,-1.34$] & $+2.39$ [$-0.11,5.11$] & $-0.98$ [$-3.59,1.74$] & $+5.26$ [$1.15,9.87$] \\
Qwen3.5 & 9B & $-5.88$ [$-9.02,-2.78$] & $+6.96$ [$3.37,10.76$] & $-1.09$ [$-4.67,2.72$] & $+5.90$ [$-0.13,11.79$] \\
Qwen3.5 & 4B & $-4.79$ [$-7.68,-1.96$] & $+3.53$ [$0.27,7.01$] & $-2.17$ [$-5.33,0.65$] & $+5.70$ [$0.75,11.67$] \\
Qwen3.5 & 2B & $-6.29$ [$-8.71,-3.81$] & $+4.57$ [$0.43,8.59$] & $+1.30$ [$-2.61,5.11$] & $+3.21$ [$-4.36,10.39$] \\
Qwen3.5 & 0.8B & $-5.98$ [$-8.20,-3.76$] & $+2.19$ [$-0.95,5.34$] & $+0.46$ [$-2.59,3.45$] & $-0.28$ [$-5.97,5.50$] \\
Qwen3-VL & 32B & $-8.14$ [$-12.11,-4.74$] & $+4.57$ [$0.33,10.22$] & $+0.22$ [$-2.83,3.04$] & $+5.13$ [$-1.54,13.46$] \\
Qwen3-VL & 30B-A3B & $-6.91$ [$-11.70,-2.42$] & $+8.04$ [$1.85,15.54$] & $-0.76$ [$-3.59,2.07$] & $+8.85$ [$0.64,18.46$] \\
Qwen3-VL & 8B & $-7.11$ [$-12.22,-2.37$] & $-0.87$ [$-6.52,3.91$] & $+1.74$ [$-0.22,4.24$] & $-3.08$ [$-10.64,3.34$] \\
Qwen3-VL & 4B & $-4.02$ [$-8.61,0.26$] & $+8.70$ [$1.41,16.52$] & $-4.46$ [$-11.63,1.96$] & $+10.38$ [$-2.69,23.72$] \\
Qwen3-VL & 2B & $-0.15$ [$-3.87,3.97$] & $-0.11$ [$-6.09,4.67$] & $-0.54$ [$-3.04,1.96$] & $+0.90$ [$-7.31,8.21$] \\
\midrule
All & 17-config mean & $-4.77$ [$-5.86,-3.70$] & $+2.63$ [$1.28,4.05$] & $-0.69$ [$-1.85,0.40$] & $+2.70$ [$0.49,4.99$] \\
\bottomrule
\end{tabular}
\end{adjustbox}
\captionsetup{hypcap=false}
\captionof{table}{Complete 17-configuration spatial results. Accuracy uses all 101 repaired pairs; destinations use the clean 46-pair distractor-only subset; the final contrast uses the nested 39-pair subset after excluding newly unique-longest edited distractors. The 46- and 39-pair analyses are forward, not reverse, and zero-crossing intervals are retained.}
\label{tab:app_spatial_directional_models}
\end{minipage}
\vspace{8pt}
}]

\subsection{Question+Options and Modality-Route Details}
\label{app:modality-route-details}

Table~\ref{tab:app_modality_lattice_numeric} reports the numeric values underlying Figure~\ref{fig:modality_lattice}. For multimodal models, \textit{Options} is answer-options-only accuracy; \textit{+Text} is the gain from adding question text; \textit{+Image} is the gain from adding image input without question text; and \textit{Full gain} is full-input accuracy minus options-only accuracy. The Question+Options guesser is a separate no-image diagnostic model, so its final column reports no-image accuracy rather than full-input accuracy.

\begin{table*}[t]
\centering
\footnotesize
\setlength{\tabcolsep}{3pt}
\renewcommand{\arraystretch}{1.05}
\begin{adjustbox}{max width=\textwidth}
\begin{tabular}{llrrrrr}
\toprule
Family & Model & Options & +Text & +Image & Full gain & Reported accuracy \\
\midrule
Guesser & 9B Q+O & -- & -- & -- & -- & 53.75 \\
\midrule
MedGemma & 27B & 18.6 & +25.3 & +13.6 & +29.3 & 48.0 \\
MedGemma & 4B & 25.9 & +10.1 & +4.2 & +12.6 & 38.5 \\
\midrule
Qwen3.5 & 2B & 25.1 & +12.2 & +6.5 & +15.4 & 40.5 \\
Qwen3.5 & 4B & 27.6 & +20.0 & +10.1 & +25.8 & 53.4 \\
Qwen3.5 & 9B & 29.4 & +21.3 & +12.1 & +28.8 & 58.3 \\
Qwen3.5 & 27B & 31.4 & +25.2 & +16.5 & +36.6 & 68.0 \\
Qwen3.5 & 35B-A3B & 29.8 & +25.6 & +15.6 & +35.3 & 65.1 \\
Qwen3.5 & 122B-A10B & 27.7 & +26.9 & +21.2 & +37.5 & 65.2 \\
\midrule
Qwen3-VL & 2B & 27.5 & +7.9 & +2.0 & +10.0 & 37.5 \\
Qwen3-VL & 4B & 27.2 & +14.0 & +4.6 & +18.2 & 45.4 \\
Qwen3-VL & 8B & 28.5 & +16.7 & +4.6 & +20.8 & 49.3 \\
Qwen3-VL & 30B-A3B & 29.6 & +19.6 & +7.1 & +24.4 & 54.0 \\
Qwen3-VL & 235B & 31.4 & +19.8 & +9.8 & +26.2 & 57.7 \\
\midrule
Open-model panel & 13 configurations & 29.71 & +24.25 & +14.13 & +32.92 & 62.63 \\
\midrule
GPT & GPT-5.4-nano & 25.8 & +14.3 & +3.2 & +18.3 & 44.1 \\
GPT & GPT-5.4-mini & 31.6 & +22.6 & +9.1 & +28.9 & 60.6 \\
GPT & GPT-5.4 & 32.1 & +29.5 & +20.2 & +41.2 & 73.3 \\
\bottomrule
\end{tabular}
\end{adjustbox}
\caption{Configuration-level six-dataset pooled modality-lattice values. All values are accuracy percentages or percentage-point gains. The Open-model panel row is the equal-weight mean of the 13 MedGemma and Qwen model/mode configurations after pooling 7,706 questions within each configuration; it excludes the GPT rows and is computed from unrounded configuration values. The last column is full-input accuracy for multimodal models and no-image accuracy for the Q+O guesser.}
\label{tab:app_modality_lattice_numeric}
\end{table*}

\FloatBarrier
\subsection{Additional Text-or-Options Ablation Results}
\label{app:additional-text-options}

Table~\ref{tab:modality} reports an auxiliary text-or-options analysis. It is narrower than the main modality-lattice analysis, but it provides useful item-level union diagnostics for MedXpertQA-MM, MMMU-HM, and MMMU-Pro-HM.

\begin{table*}[t]
\centering
\small
\begin{tabular}{llrrrrr}
\toprule
Dataset & Model & $N$ & Options-only & Text-only & Text-or-options & Full input \\
\midrule
MedXpertQA-MM & GPT-5.4-nano & 2,000 & 20.25\% & 24.80\% & 36.70\% & 26.70\% \\
MedXpertQA-MM & Qwen3.5-9B & 2,000 & 22.35\% & 27.30\% & 40.65\% & 30.30\% \\
MMMU-HM & GPT-5.4-nano & 1,712 & 31.54\% & 40.95\% & 56.02\% & 47.08\% \\
MMMU-HM & Qwen3.5-9B & 1,712 & 32.07\% & 48.25\% & 62.44\% & 63.32\% \\
MMMU-Pro-HM & GPT-5.4-nano & 286 & 18.53\% & 19.23\% & 29.72\% & 22.73\% \\
MMMU-Pro-HM & Qwen3.5-9B & 286 & 19.23\% & 20.98\% & 34.27\% & 45.80\% \\
\bottomrule
\end{tabular}
\caption{Ablation results for the text-or-options analysis. Text-or-options is the union of cases solved in either options-only or text-only settings.}
\label{tab:modality}
\end{table*}

\subsection{Controlled Editing Details}
\label{app:controlled-editing-details}

Figure~\ref{fig:option_form_shortcuts} summary intervals use model-level resampling of the repeat-20 configuration summaries. By contrast, Tables~\ref{tab:app_length_directional_models}--\ref{tab:app_spatial_directional_models} use paired item bootstraps after averaging 20 repeats within each item$\times$configuration$\times$condition cell. Length is dataset-stratified, while the absolute/conspicuous and spatial/prepositional families use their analysis-specific item-bootstrap procedures.

The three accepted repair cohorts contain 306 length-gap, 101 spatial/prepositional, and 101 absolute/conspicuous pairs. Each cohort is evaluated over 20 repeats per condition using the construction and validation criteria in Appendix~\ref{app:option-route-details}.

\subsection{Additional Observational Text Signals}
\label{app:observational-text-signals}

The GPT-5.4 normal-setting shortcut audit over 7,722 observed/source variants reports lexical-clang, semantic, position, and length associations. Examples include lexical clang with canonical odds ratio 1.97 and length-as-unique-longest-by-words with odds ratio 1.33. These are observational screens, not causal estimates; the main paper uses them to motivate controlled edits.

\FloatBarrier
\section{Image- and Context-Side Analyses}
\label{app:image-context-analyses}

\subsection{Image-Channel Audit Details}
\label{app:image-channel-details}

Table~\ref{tab:app_image_side_audit} summarizes the image- and context-side screening surfaces before the paired interventions and clinical adjudication described below.

\begin{table*}[t]
\centering
\small
\begin{tabularx}{\textwidth}{l l r X}
\toprule
Audit & Unit & Scope & Key counts / interpretation \\
\midrule
Image-feature audit & Images & 13,940 & 5,645 text residuals; 4,970 visible manual markups/overlays; 2,642 context features; 955 device/material cues \\
Option-text leakage join & Cases & 7,706 & 608 partial option matches; 233 complete option matches; 60 high-risk cases; 195 medium-risk cases \\
Artificial image-hint intervention & Cases & 200 & Paired original-vs-cue-removed evaluation with 10 repeated runs per model \\
Natural clinical-context neutralization & Cases & 302 & Paired original-vs-neutralized stem evaluation with 10 repeated runs per model \\
Device/material audit & Candidates & 1,347 & 1,290 no-shortcut; 48 borderline; 9 confirmed shortcut-risk cases \\
\bottomrule
\end{tabularx}
\caption{Image-channel and context-channel risk surfaces. These are prevalence or risk-screen outputs, not direct model-reliance evidence.}
\label{tab:app_image_side_audit}
\end{table*}

\subsection{Artificial Image-Hint Intervention Details}
\label{app:image-hint-details}

Table~\ref{tab:app_image_hint_funnel} reports the repair funnel; Tables~\ref{tab:app_image_hint_selected_models} and~\ref{tab:app_image_hint_family_summary} report model- and family-level results.

\begin{table*}[t]
\centering
\small
\setlength{\tabcolsep}{4pt}
\renewcommand{\arraystretch}{1.08}
\begin{tabularx}{\textwidth}{l r X}
\toprule
Stage & $N$ & Description \\
\midrule
Broad screen candidates & 2,605 & GPT-5.4-mini high-recall screen for possible manual annotations \\
Strict manual annotations & 1,088 & GPT-5.4 adjudication excluding calipers, native diagrams, UI overlays, graph elements, and non-hint figure text \\
Strong image-hint shortcuts & 856 & Added overlays that directly localize answer-relevant findings and are not task-essential \\
Repair sample & 370 & Strong-hint cases sampled for localized image editing \\
Retained paired cases & 200 & Passed best-of-five candidate generation, quality judging, and refinement \\
\bottomrule
\end{tabularx}
\caption{Image-hint identification and repair funnel. Only retained paired cases enter the original-vs-edited model evaluation.}
\label{tab:app_image_hint_funnel}
\end{table*}

Edited candidates must remove the cue, leave no visible remnants, preserve the medically relevant image content, avoid new artifacts, and keep the original question-answer semantics intact. Cases that cannot be edited safely are discarded.

\begin{table}[tbp]
\centering
\small
\setlength{\tabcolsep}{5pt}
\renewcommand{\arraystretch}{1.08}
\begin{tabular}{lrr}
\toprule
Model & $\Delta_{\mathrm{image}}$ & Bad/cue ratio \\
\midrule
MedGemma-27B & -0.05 pp & 1.7\% \\
Qwen3-VL-32B & -0.95 pp & 4.8\% \\
Qwen3-VL-30B-A3B & -1.50 pp & 6.5\% \\
Qwen3.5-0.8B & -3.20 pp & 16.6\% \\
Qwen3.5-35B-A3B & -3.75 pp & 13.4\% \\
GPT-5.4 & -0.05 pp & 5.6\% \\
GPT-5.4-nano & -0.40 pp & 6.5\% \\
GPT-5.4-mini & -3.55 pp & 14.3\% \\
\bottomrule
\end{tabular}
\caption{Selected model-level image-hint results. $\Delta_{\mathrm{image}}$ is cue-removed accuracy minus original-image accuracy; negative values indicate that removing the hint reduced accuracy.}
\label{tab:app_image_hint_selected_models}
\end{table}

\begin{table}[tbp]
\centering
\scriptsize
\setlength{\tabcolsep}{1.5pt}
\renewcommand{\arraystretch}{1.08}
\begin{tabular}{lrrrrr}
\toprule
Family & $n$ & Cue & Bad & B/C & Mean $\Delta_{\mathrm{image}}$ \\
\midrule
Qwen3.5 & 6 & 40.7\% & 5.6\% & 13.8\% & -1.82 pp \\
Qwen3-VL & 5 & 36.6\% & 1.7\% & 4.6\% & -0.83 pp \\
MedGemma & 2 & 41.4\% & 2.0\% & 4.9\% & +0.23 pp \\
GPT & 3 & 19.0\% & 1.5\% & 8.1\% & -1.18 pp \\
\bottomrule
\end{tabular}
\caption{Family-level image-hint behavior. $\Delta_{\mathrm{image}}$ is cue-removed accuracy minus original-image accuracy; negative values indicate that cue removal reduced accuracy.}
\label{tab:app_image_hint_family_summary}
\end{table}

\subsection{Natural Clinical-Context Cue Details}
\label{app:natural-cue-details}

The natural-cue set contains 302 paired cases. Cue categories are non-mutually exclusive and include race/ethnicity/origin, substance use, family or social-support context, geography/rurality/migration, disability/function/body size, housing/living environment, sex/gender/reproductive context, occupation, and socioeconomic status (SES)/access context. Tables~\ref{tab:app_natural_cue_selected_models}, \ref{tab:app_natural_cue_family_summary}, and~\ref{tab:app_natural_cue_category} report model-, family-, and category-level results.

\begin{table}[tbp]
\centering
\small
\setlength{\tabcolsep}{5pt}
\renewcommand{\arraystretch}{1.08}
\begin{tabular}{lrr}
\toprule
Model & $\Delta_{\mathrm{natural}}$ & Bad/cue ratio \\
\midrule
Qwen3-VL-30B-A3B & +1.46 pp & 2.2\% \\
MedGemma-27B & +0.70 pp & 2.8\% \\
Qwen3-VL-32B & -1.89 pp & 4.1\% \\
Qwen3.5-35B-A3B & -1.89 pp & 10.6\% \\
Qwen3.5-0.8B & -3.08 pp & 20.4\% \\
GPT-5.4-nano & -0.40 pp & 4.7\% \\
GPT-5.4 & -1.19 pp & 15.2\% \\
\bottomrule
\end{tabular}
\caption{Selected model-level natural-cue results. $\Delta_{\mathrm{natural}}$ is neutralized accuracy minus original accuracy; negative values indicate that neutralization reduced accuracy.}
\label{tab:app_natural_cue_selected_models}
\end{table}

The family patterns are mixed for GPT (GPT-5.4 drops while GPT-5.4-mini improves), strongest harmful reliance for Qwen3.5, frequent cue mention but low harmful conversion for Qwen3-VL, and high cue mention in one MedGemma configuration with a weak aggregate drop.

\begin{table}[tbp]
\centering
\scriptsize
\setlength{\tabcolsep}{2pt}
\renewcommand{\arraystretch}{1.08}
\begin{tabular}{lrrrr}
\toprule
Family & $n$ & Mean $\Delta_{\mathrm{natural}}$ & Cue-use & Bad/cue \\
\midrule
GPT & 3 & -0.02 pp & 17.5 & 8.7\% \\
Qwen3.5 & 6 & -1.29 pp & 47.4 & 12.6\% \\
Qwen3-VL & 5 & +0.05 pp & 60.5 & 2.9\% \\
MedGemma & 3 & +0.26 pp & 73.5 & 5.8\% \\
\bottomrule
\end{tabular}
\caption{Family-level summary for natural clinical-context cue neutralization.}
\label{tab:app_natural_cue_family_summary}
\end{table}

\begin{table}[tbp]
\centering
\scriptsize
\setlength{\tabcolsep}{3pt}
\renewcommand{\arraystretch}{1.08}
\begin{tabularx}{\columnwidth}{@{}Xrrr@{}}
\toprule
Cue category & Cue use & Bad use & Bad/cue \\
\midrule
Race/ethnicity/origin & 372 & 69 & 18.5\% \\
Substance use & 317 & 36 & 11.4\% \\
Family/social support/education & 112 & 15 & 13.4\% \\
Geography/rurality/migration & 74 & 15 & 20.3\% \\
Disability/function/body size & 57 & 10 & 17.5\% \\
Housing/living environment & 42 & 6 & 14.3\% \\
Sex/gender/reproductive & 33 & 4 & 12.1\% \\
Occupation & 38 & 3 & 7.9\% \\
Socioeconomic status (SES)/access & 40 & 2 & 5.0\% \\
\bottomrule
\end{tabularx}
\caption{Cue-use and harmful-conversion counts by high-level removed-cue category. Categories describe judge-labeled model rationale behavior, not dataset prevalence.}
\label{tab:app_natural_cue_category}
\end{table}

These counts are not a demographic-bias benchmark. They describe whether naturally occurring clinical-context cues can become shortcut-like in a selected neutralization set. A cue is problematic only when it is nonessential for the intended image-grounded construct and contributes to answer selection.

\FloatBarrier
\subsection{Visible Image-Text Leakage Details}
\label{app:image-text-leakage-details}

We audit visible text rendered directly inside evaluation image pixels for answer-revealing leakage. The audit excludes question text, metadata, filenames, rationales, and captions outside the image. Candidate cases are screened by overlap between image text and correct-answer content, followed by GPT-5.4 adjudication and repairability review.

Table~\ref{tab:app_image_text_leakage_funnel} reports the dedicated visible-image-text audit funnel.

The 60 high-risk cases in Table~\ref{tab:app_image_side_audit} and the 65 cases here come from distinct analyses and are not interchangeable. The former count follows a join between the option-text leakage results and the canonical 7,706-case audit; the latter comes from the dedicated visible-image-text adjudication set. We retain both scoped counts rather than silently collapsing them.

Only a small subset of cases are directly repairable because many images contain task-essential labels or text that cannot be removed without changing the medical validity or structure of the item. When image text explicitly names or strongly cues the answer, the preferred mitigation is targeted masking, cropping, or neutral relabeling while preserving image validity. Otherwise, the item should be removed or redesigned rather than edited.

\begin{table}[H]
\centering
\footnotesize
\setlength{\tabcolsep}{3pt}
\renewcommand{\arraystretch}{1.05}
\begin{tabularx}{\columnwidth}{@{}>{\raggedright\arraybackslash}p{0.31\columnwidth}r>{\raggedright\arraybackslash}X@{}}
\toprule
Stage & $N$ & Description \\
\midrule
Text-bearing cases & 4,122 & Visible rendered text detected inside image pixels \\
Correct-answer-content overlap & 172 & Partial or complete overlap with correct-answer content \\
High-risk image-text shortcuts & 65 & Adjudicated as answer-revealing or strongly answer-cueing \\
Directly image-editable & 15 & Repairable without invalidating the medical image or redesigning the item \\
Model-only dataset-ready repairs & 14/15 & Passed leak-removal, no-new-shortcut, medical-fidelity, and answerability checks \\
\bottomrule
\end{tabularx}
\caption{Visible image-text leakage audit funnel.}
\label{tab:app_image_text_leakage_funnel}
\end{table}

\subsection{Device and Material Cue Details}
\label{app:device-material-details}

We audit post-treatment devices and materials as a clinically specific shortcut category. These cues include visible hardware, coils, filters, tubes, shunts, instruments, implants, or other procedure-related material that may reveal prior intervention, disease stage, or management history without requiring the intended diagnostic reasoning. Table~\ref{tab:app_device_material_counts} reports the adjudicated candidate counts.

\begin{table}[H]
\centering
\small
\setlength{\tabcolsep}{5pt}
\renewcommand{\arraystretch}{1.08}
\begin{tabular}{lr}
\toprule
Adjudication label & Cases \\
\midrule
No shortcut & 1,290 \\
Borderline shortcut & 48 \\
Confirmed shortcut & 9 \\
\midrule
Total candidates & 1,347 \\
\bottomrule
\end{tabular}
\caption{Post-treatment device/material shortcut-risk adjudication.}
\label{tab:app_device_material_counts}
\end{table}

Confirmed device/material shortcuts are rare, so we do not treat them as an aggregate performance result. However, they remain important data-quality risks because the visual cue can narrow the answer through treatment history rather than the intended clinical or imaging evidence.

These cases are also difficult to repair safely. Removing implants, tubes, coils, hardware, shunts, filters, or surgical instruments can change the medical image itself or create new artifacts. Therefore, when the device/material cue is not part of the intended construct, we recommend excluding the item from the shortcut-mitigated benchmark rather than attempting image repair.

\FloatBarrier
\section{\cleanbench{} Construction, Validation, and Release}
\label{app:benchmark-validation}

\subsection{\cleanbench{} Benchmark Details}
\label{app:medqamm-details}

\begin{figure*}[t]
    \centering
    \includegraphics[width=0.95\textwidth]{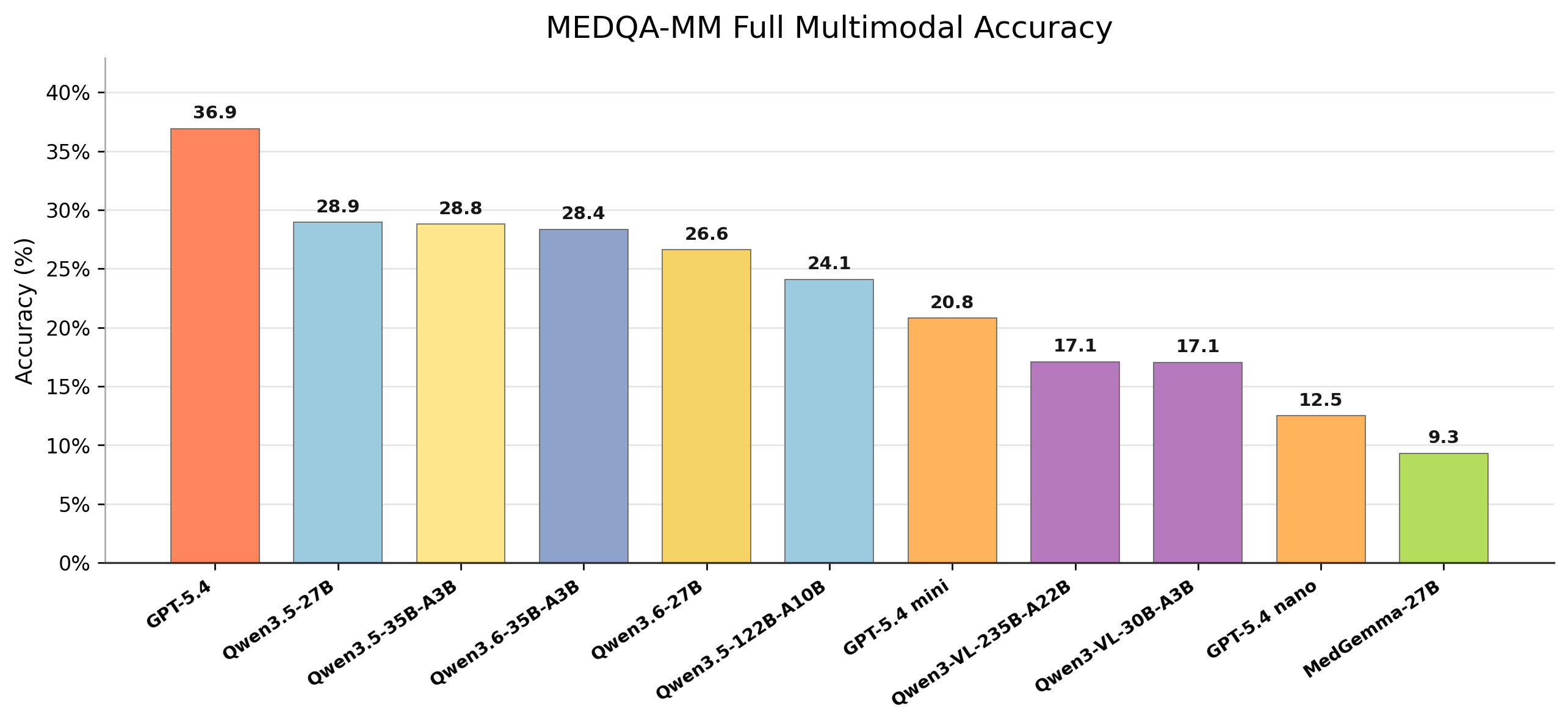}

\caption{
Full multimodal accuracy on MEDQA-MM. 
Each bar represents one base model. When multiple inference modes were evaluated, their configuration-level accuracies were averaged within that base model; Qwen3.5 and Qwen3.6 denote distinct model series. 
MEDQA-MM contains 1,000 hard, image-dependent medical VQA examples after removing device/material cue cases.
}

    \label{fig:medqamm}
\end{figure*}

\cleanbench{} contains 1,000 hard, image-dependent medical VQA examples after removing confirmed device/material cue cases. It should be interpreted as a shortcut-mitigated benchmark with documented residual risk, not as a shortcut-free benchmark. Figure~\ref{fig:medqamm} reports full-input accuracy with one bar per base model. When multiple inference modes were evaluated for a base model, their configuration-level accuracies were averaged before plotting; Qwen3.5 and Qwen3.6 are treated as distinct model series \citep{QwenTeam2026Qwen35,QwenTeam2026Qwen36}.

\subsection{\cleanbench{} Error-Analysis Protocol}
\label{app:medqamm-error-analysis}

This targeted error analysis addresses the otherwise underdetermined observation that Image+Options accuracy (28.47\%) exceeds Full-input accuracy (26.59\%) on \cleanbench{}. It is separate from the repair-validation study in Appendix~\ref{app:expert-protocol}. We sample 100 incorrect predictions from each of three representative configurations---GPT-5.4, MedGemma-27B, and Qwen3.5-122B-A10B---for 300 cases total. GPT-5.6-Sol assigns a binary judgment of whether question-stem content overrode decisive image evidence and contributed to the incorrect prediction. A separate three-expert medical panel---two senior clinicians and one medically trained annotator---conducts the error analysis. A 100-case subset of the binary screen is independently expert-reviewed; agreement with the LLM judge is 92\%, with Cohen's $\kappa=0.703$. No confidence interval, confusion matrix, sensitivity, or specificity estimate is inferred beyond the reported agreement statistics.

The same panel also assigns one mutually exclusive primary label to each case in a separate 70-error development split: visual-understanding failure, reasoning/integration failure, item/data issue, response/scoring failure, or shortcut/anchoring bias. The complete distribution is 15, 21, 29, 3, and 2 cases, respectively. Thus, ``visual or reasoning/integration'' is a derived subtotal of $15+21=36$ cases, not an additional class and not an intersection. Because this is a development analysis, its category proportions are descriptive of the reviewed sample rather than a held-out estimate for all \cleanbench{} errors.

These analyses support a bounded interpretation. Stem interference was judged present in a minority of sampled errors and may help explain some cases where removing the stem improves performance. The taxonomy also shows that item/data issues, visual failures, reasoning/integration failures, response/scoring problems, and shortcut/anchoring behavior coexist. Neither sample identifies a single cause of the aggregate Image+Options--Full difference.

\subsection{Expert Annotation Protocol}
\label{app:expert-protocol}
The separate repair-validation pool contains five medical expert annotators: two senior clinicians and three medically trained annotators. The two senior clinicians are a diagnostic radiologist at a tertiary academic hospital in Asia with 7 years of post-training experience and an academic surgeon at a U.S. medical school with 5 years of post-training experience.

We conduct targeted expert review for 150 sampled option-form repairs, 50 natural-cue neutralizations, and 46 image-hint edits. Reviewers judge cue relevance, medical validity, one-best-answer preservation, and mitigation quality through two related workflows. For option-form repairs, they judge whether the detected cue could make the correct answer stand out or make distractors easier to eliminate, and whether the modified item remains valid while reducing the cue. For natural cues and image hints, they check whether the removed cue is unnecessary for the intended answer and whether the neutralized stem or edited image preserves the medically relevant content. Tables~\ref{tab:app_expert_annotation_dimensions} and~\ref{tab:app_expert_annotation_summary} give the schema and sampled outcomes.

\begin{table*}[t]
\centering
\small
\setlength{\tabcolsep}{4pt}
\renewcommand{\arraystretch}{1.08}
\begin{tabularx}{\textwidth}{>{\raggedright\arraybackslash}p{0.18\textwidth}>{\raggedright\arraybackslash}p{0.28\textwidth}>{\raggedright\arraybackslash}X}
\toprule
Review dimension & Response type & Meaning \\
\midrule
Cue status & yes / probably yes / no & Whether the original cue plausibly provided shortcut help \\
Medical validity & valid / minor concern / invalid & Whether the edited or neutralized item remains medically valid \\
Mitigation & improved / partially improved / not improved & Whether the modified item reduces the original shortcut signal \\
Comments & free text & Used for questionable originals, unsafe edits, or better revision suggestions \\
\bottomrule
\end{tabularx}
\caption{Expert annotation dimensions used for sampled repair and neutralization review.}
\label{tab:app_expert_annotation_dimensions}
\end{table*}

\begin{table*}[t]
\centering
\small
\setlength{\tabcolsep}{4pt}
\renewcommand{\arraystretch}{1.05}
\begin{tabularx}{\textwidth}{l l >{\raggedright\arraybackslash}X r}
\toprule
Case set & Dimension & Outcome & $n$ (\%) \\
\midrule
Option-form repairs ($N=150$)
& Validity & Valid & 142 (94.7) \\
& & Minor concern & 8 (5.3) \\
& & Invalid & 0 (0.0) \\
& Mitigation & Improved & 91 (60.7) \\
& & Partially improved & 57 (38.0) \\
& & Improved / partial & 148 (98.7) \\
& & Not improved & 2 (1.3) \\
\midrule
Natural cue ($N=50$)
& Cue & Yes / probably yes & 50 (100.0) \\
& Validity & Valid & 50 (100.0) \\
\midrule
Image hint cue ($N=46$)
& Cue & Yes / probably yes & 46 (100.0) \\
& Validity & Valid & 45 (97.8) \\
& & Minor concern & 1 (2.2) \\
\bottomrule
\end{tabularx}
\caption{Human expert annotation summary for sampled repaired or neutralized cases. Option-form repairs combine the three repair families, sampled at 50 cases per family.}
\label{tab:app_expert_annotation_summary}
\end{table*}

\FloatBarrier
\raggedbottom
\subsection{Validation and Release Status}
\label{app:validation-release}

\begin{table}[H]
\centering
\scriptsize
\setlength{\tabcolsep}{2.5pt}
\renewcommand{\arraystretch}{1.06}
\begin{tabularx}{\columnwidth}{@{}>{\raggedright\arraybackslash}p{0.24\columnwidth}>{\raggedright\arraybackslash}p{0.34\columnwidth}>{\raggedright\arraybackslash}X@{}}
\toprule
Target & Evidence available & Remaining requirement \\
\midrule
Option-form repairs & Sample expert annotation over 150 cases & Full release-scale validation if released as edited benchmark items \\
Image-hint repairs & 200 paired model-evaluation cases; 46 usable expert annotations & Additional clinician fidelity review before release \\
Natural-cue neutralization & 302 paired model-evaluation cases; 50 expert annotations & Clinician confirmation for all released neutralized items \\
Image-text repair & 14/15 model-only dataset-ready repairs & Human image-fidelity validation before release \\
Device/material cues & 9 confirmed and 48 borderline adjudicated cases & Exclude confirmed unrepairable cases or obtain expert disposition \\
\cleanbench{} release & 1,000-example shortcut-mitigated benchmark & Source-license, image-rights, and item-level redistribution checks \\
\bottomrule
\end{tabularx}
\caption{Validation and release status. Sample-level expert annotation supports the audit, but public release still requires source and item-level validation.}
\label{tab:app_validation_release_status}
\end{table}

Table~\ref{tab:app_source_release_handling} distinguishes source documentation from this paper's new audit and repair metadata. It records release handling without implying that redistribution permission has been obtained.

\begin{table}[H]
\centering
\scriptsize
\setlength{\tabcolsep}{2.5pt}
\renewcommand{\arraystretch}{1.05}
\begin{tabularx}{\columnwidth}{@{}>{\raggedright\arraybackslash}p{0.22\columnwidth}>{\raggedright\arraybackslash}p{0.29\columnwidth}>{\raggedright\arraybackslash}X@{}}
\toprule
Source & Documentation & Release handling \\
\midrule
AMBOSS image questions & QBank and terms \citep{AMBOSS2026QBank,AMBOSS2026Terms} & Audit-only source; no redistribution claim \\
JAMA Clinical Challenge & Source and permissions policy \citep{JAMAClinicalChallenge,JAMA2026Permissions} & Audit-only source; source and image permissions require item-level review \\
MMMU Health/Medicine & Paper and dataset card \citep{Yue2024MMMU,MMMU2026DatasetCard} & \cleanbench{} source; preserve inherited provenance/license fields and add audit metadata \\
MedThinkVQA & Paper and dataset card \citep{Yao2026MedThinkVQA,MedThinkVQA2026DatasetCard} & \cleanbench{} source; preserve inherited fields and add repair/validation status \\
MedXpertQA-MM & Dataset paper \citep{Zuo2025MedXpertQA} & \cleanbench{} source; redistribution and image-rights status remain to be confirmed \\
NEJM Image Challenge & Source and permissions policy \citep{NEJMImageChallenge,NEJM2026Permissions} & Audit-only source; not included in \cleanbench{} \\
\bottomrule
\end{tabularx}
\caption{Source documentation and release handling. Source-provided fields are inherited from the cited dataset or policy pages; cue flags, repair provenance, validation status, and residual-risk fields are new metadata from this work. This table documents required release checks and does not grant redistribution rights.}
\label{tab:app_source_release_handling}
\end{table}

\FloatBarrier

\section{Prompt Templates}
\label{app:prompts}

This appendix reproduces the substantive prompt templates used for shortcut detection, adjudication, repair, and quality control. The templates cover visible image text, treatment devices and materials, option-text repair, artificial visual-hint removal, and the sociodemographic/SDOH audit pipeline. They are included because small wording changes affect repair admissibility and shortcut adjudication. Operational metadata that does not affect the model instructions is omitted. Here, SDOH denotes social determinants of health.

\clearpage
\onecolumn
\raggedbottom
\Needspace{12\baselineskip}
\subsection*{Visible Image Text Audit}
\addcontentsline{toc}{subsection}{Visible Image Text Audit}

\begin{tcolorbox}[
title={Prompt used for GPT-5.4-mini visible text screening.},
width=\linewidth,
colframe=blue!50!black,
colback=blue!5!white,
breakable
]

\small

\textbf{Purpose}

Image-only OCR-style screen for rendered text residuals. It does not judge leakage; it only records visible text and location/type.

\textbf{Prompt}

\begin{Verbatim}[breaklines=true,breakanywhere=true]
You are auditing medical VQA images for visible text residuals.

For each image, inspect only the visible pixels in the image.
Report whether the image contains any rendered text, including titles, labels, captions, DICOM overlays, table text, axis labels, legends, annotations, arrows with labels, or source marks.

Do not infer text from the question, answer options, metadata, explanations, or filenames.
Only transcribe text that is visibly present inside the image.

Return JSON only.
Structured output:

{
  "case_id": "...",
  "image_id": "...",
  "has_visible_text": true,
  "visible_text": [
    {
      "text": "...",
      "location": "top | bottom | left | right | center | table | overlay | other",
      "text_type": "caption | panel_title | diagnosis_label | annotation_label | dicom_overlay | table_text | axis_label | legend | other"
    }
  ],
  "notes": "..."
}
\end{Verbatim}

\end{tcolorbox}
\label{prompt:visible-text-screening}

\begin{tcolorbox}[
title={Prompt used for GPT-5.4 image-text shortcut judging.},
width=\linewidth,
colframe=blue!50!black,
colback=blue!5!white,
breakable
]

\small

\textbf{Purpose}

Adjudicate whether visible image text creates an answer-option shortcut. This uses question/options and correct answer only for audit context, while restricting shortcut evidence to rendered image text.

\textbf{Prompt}

\begin{Verbatim}[breaklines=true,breakanywhere=true]
You are auditing medical multiple-choice VQA cases for visible image-text shortcut leakage.

Important scope:
- Only consider text visibly rendered inside the evaluation-visible images.
- Do NOT consider leakage from the question stem, clinical history, imaging findings fields, caption metadata, diagnosis fields, explanation fields, or answer rationales.
- The clinical history and answer options are normal model inputs. Do not flag them as leakage.
- Captions or labels count only if they are visible as pixels inside the image.

Given:
1. Question / clinical history and answer options, used only to understand the intended task.
2. The correct answer, used for audit.
3. Evaluation-visible images.

Your task:
Classify whether the visible image text creates an answer-option shortcut.

Definitions:
- Actionable image-text shortcut: nonessential visible image text, such as a diagnosis label, panel title, caption, overlay, or annotation, that directly names or strongly cues the correct answer.
- Legitimate image-text evidence: visible image text that is part of the intended task evidence, such as lab tables, axes, legends, statistical tables, chemical formulas, row/column labels, or measurements required to answer the question.
- Balanced option visibility: the correct answer and multiple distractors are similarly visible in the image, so the image text does not uniquely cue the correct answer.
- Not image-editable: the answer-related visible text is inseparable from the task evidence, or removing it would make the item invalid.

Return JSON only.
\end{Verbatim}

\end{tcolorbox}
\label{prompt:image-text-shortcut-judge}

\subsection*{Treatment Device/Material Shortcut Audit}
\addcontentsline{toc}{subsection}{Treatment Device/Material Shortcut Audit}

\begin{tcolorbox}[
title={Prompt used for treatment device/material shortcut adjudication.},
width=\linewidth,
colframe=blue!50!black,
colback=blue!5!white,
breakable
]

\small

\textbf{Purpose}

Decide whether visible devices/materials, image-embedded text, or stem procedure/device context creates shortcut risk.

\textbf{System Prompt}

\begin{Verbatim}[breaklines=true,breakanywhere=true]
You are auditing medical VQA benchmark cases for shortcut-risk from treatment devices, post-treatment materials, and device/material text cues.

Definitions:
- A device/material cue may be a visible physical device/material in the image, image-embedded text/OCR that names a device/material, or stem text that names a device/material or a procedure strongly implying one.
- A confirmed shortcut-risk exists when the cue can plausibly identify or strongly favor the correct option without the intended medical reasoning, usually because distractors do not control for the same cue.
- A borderline shortcut-risk exists when the cue is unintended and meaningfully narrows the answer space, but still requires some medical reasoning or is not fully decisive.
- Normal intended clinical evidence is not a shortcut. A treatment/procedure can be part of expected reasoning; do not mark it as shortcut solely because it is clinically causally related to the answer. However, if weak options make the device/procedure context itself single out the only post-treatment, iatrogenic, or device-specific answer, keep at least borderline; use confirmed only when little additional reasoning is needed.
- If the question asks what a device is, compares fixation methods, asks about a shown procedure, or the image/text cue is necessary intended evidence, mark no_shortcut or intended_evidence.
- If the stem already states the treatment/device and the image merely repeats it, mark redundant_context rather than visible-device shortcut. Redundant context should normally be no_shortcut unless there is an additional non-redundant design leak.
- If only stem text contains the device/material cue, do not call it an image shortcut. It may be a stem_device_context_or_design_issue only when the stem cue itself signposts the correct option through weak distractors, for example when only one option matches the named procedure/device as an iatrogenic or post-treatment complication.
- Direct wording overlap between stem and an answer option is not automatically a device/material shortcut. Mark it only if the treatment device/material/procedure cue itself is the revealing signal.
- If image text literally names the correct answer, names a device/material unique to the correct figure/option, or exposes a procedure that selects one figure among options, treat that as image_text_device_leakage when appropriate.
- Keep borderline cases when the unintended cue meaningfully narrows choices, but weak background nudges should be no_shortcut.
- Consistency rule: shortcut_type redundant_context, intended_evidence, or false_positive should normally pair with no_shortcut, not confirmed_shortcut. Use confirmed_shortcut mainly with visible_device_shortcut, image_text_device_leakage, stem_device_context_or_design_issue, or mixed_or_other.

Return concise structured JSON only.
\end{Verbatim}

\textbf{User Prompt Template}

\begin{Verbatim}[breaklines=true,breakanywhere=true]
Audit this case for treatment-device/material shortcut-risk.

Dataset: {dataset}
Example id: {example_id}
Candidate source group: {source_group}
Images sent: {image_count_sent}

Question stem:
{question}

Options:
{options_text}

Correct answer: {correct_answer}. {correct_answer_text}

Mini text cue screen positives:
{text_screen_summary}

First-pass image feature summary for relevant images:
{feature_summary}

Task:
1. Confirm whether each candidate cue source is real: visible_device, image_text, stem_text.
2. Decide whether the cue is intended evidence, redundant with the stem, or a false positive.
3. Decide whether the cue is non-redundant and unintended, and whether weak option design lets it identify or strongly favor the correct option.
4. Use confirmed_shortcut only for strong leaks. Use borderline_shortcut for meaningful but incomplete narrowing. Use no_shortcut for intended evidence, redundant context, false positives, and weak background nudges.
5. For stem-only cases, remember no image is sent by design; judge only stem text device/material/procedure context or option-design leakage, and avoid labeling ordinary clinical history as shortcut.

Return JSON only.
\end{Verbatim}

\end{tcolorbox}
\label{prompt:treatment-device-material-adjudication}

\subsection*{Option-Text Shortcut Repair}
\addcontentsline{toc}{subsection}{Option-Text Shortcut Repair}

\begin{tcolorbox}[
title={Prompt used for GPT-5.4 absolute-distractor repair.},
width=\linewidth,
colframe=blue!50!black,
colback=blue!5!white,
breakable
]

\small

\textbf{Purpose}

Repair distractor options that contain absolute terms or over-broad shortcut cues, while preserving the correct answer and medical truth conditions.

\textbf{System Message}

\begin{Verbatim}[breaklines=true,breakanywhere=true]
You are an expert medical exam editor repairing multiple-choice medical VQA questions. You remove superficial option shortcuts while preserving the medically correct answer, the intended meaning, and the intended difficulty. Be conservative. Return JSON only.
\end{Verbatim}

\textbf{User Prompt Template}

\begin{Verbatim}[breaklines=true,breakanywhere=true]
You will receive one medical multiple-choice question, all answer options, the known correct answer, available metadata, image references if present, and a detected absolute-distractor shortcut.

Task:
Make the smallest safe option edits that remove or neutralize the absolute-distractor shortcut.

Priority order:
1. Preserve the original correct answer letter and medical meaning.
2. Preserve the question stem exactly.
3. Prefer editing only target distractor option(s).
4. You may edit a non-target distractor only if needed to avoid duplicates, keep option-set balance, or avoid creating a new clue.
5. Avoid editing the correct option. Only edit the correct option if the edit is purely wording-level, preserves the same answer, and is needed to remove a new imbalance.

Hard safety rules:
- Do not change option order or option letters.
- Do not make a distractor synonymous with the correct answer.
- Do not make any two options duplicates or near-duplicates.
- Do not introduce new shortcut terms: always, never, must, cannot, impossible, only, all, none, neither, required, requires, necessary, ruled out, confirmed.
- Do not change anatomy, laterality, organism, diagnosis, numeric values, units, severity, treatment direction, causal direction, or temporal order unless the original distractor remains clearly wrong and the edit is medically safe.
- If removing the absolute cue would make the distractor partly correct, ambiguous, or require clinical invention, return `no_safe_edit` or `needs_human_review`.
- For `no_safe_edit` or `needs_human_review`, return an empty `option_patches` list.

Recommended edit strategies:
- only/sole/exclusive: replace with a concrete but still wrong alternative, or remove the word only if the option remains clearly wrong.
- cannot/impossible/always/never/must: replace with a concrete but still wrong interpretation or limitation; do not merely weaken into a true/ambiguous statement.
- required/requires/necessary: replace with a less absolute but still wrong management/protocol claim.
- rule out/exclude: replace with `evaluate for`, `assess for`, or a similar non-absolute phrase only if it remains wrong in this question.
- confirmed/diagnostic/pathognomonic: replace with `supports`, `suggests`, or another non-absolute phrase only if it remains wrong.
- all/both/every/none/neither as an over-broad distractor: replace with a specific wrong subset/claim when safe; otherwise use review/no_safe_edit.

Dataset: {dataset}
Example id: {example_id}
Target categories: {target_categories}
Target distractor triggers:
{trigger_text}
Question stem:
{question}

Correct answer:
{correct_answer}. {correct_text}

Options:
{options_text}

Additional metadata:
{metadata_text}
Return JSON matching this schema:
{
  "status": "rewritten|no_safe_edit|needs_human_review",
  "changed_letters": ["B"],
  "option_patches": [
    {
      "letter": "B",
      "before": "...",
      "after": "...",
      "is_correct_option": false,
      "strategy": "soften_absolute|replace_absolute_claim|remove_scope_word|replace_with_specific_distractor|rebalance_option_set|no_change",
      "removed_or_softened_terms": ["only"],
      "meaning_preservation_notes": "...",
      "risk_notes": ""
    }
  ],
  "correct_answer_letter_after_edit": "C",
  "correct_answer_meaning_preserved": true,
  "human_review_needed": false,
  "understanding_summary": "One sentence explaining the task and correct answer.",
  "edit_rationale": "One or two sentences explaining the option changes.",
  "safety_rationale": "Why the edited options remain safe/wrong/correct.",
  "residual_shortcut_risk": "low|medium|high"
}
\end{Verbatim}

\end{tcolorbox}
\label{prompt:absolute-distractor-repair}

\Needspace{12\baselineskip}
\begin{tcolorbox}[
title={Prompt used for GPT-5.4 length-shortcut rewrite.},
width=\linewidth,
colframe=blue!50!black,
colback=blue!5!white,
breakable
]

\small

\textbf{Purpose}

Minimally rewrite one or two distractors so the correct option is no longer uniquely longest.

\textbf{System Prompt}

\begin{Verbatim}[breaklines=true,breakanywhere=true]
You are a conservative medical multiple-choice benchmark editor. You will receive a medical visual MCQ, all available images, the known correct answer, and a detected option-length shortcut. Your job is to remove the length shortcut only when it is safe, while preserving the intended medical meaning, image-grounded reasoning, answer key, option order, and number of options. Do not be clever. Prefer no_safe_edit over a rewrite that changes medical truth conditions.
\end{Verbatim}

\textbf{User Prompt}

\begin{Verbatim}[breaklines=true,breakanywhere=true]
Please review this medical multimodal MCQ and, only if safe, minimally rewrite 1-2 distractor options to neutralize a length shortcut.

Context:
- The correct option is uniquely the longest option by character count.
- The goal is to make at least one distractor clearly longer than the correct option, so the correct option is no longer uniquely longest.
- Automated validation is strict. At least one revised distractor should be at least 5 visible characters longer than the correct option. Do not aim for a 1-2 character margin, because character counting is easy to get wrong.
- You may change only 1 or 2 distractor options. Do not change the correct option.
- Use the images to understand the intended case before editing.

Hard rules:
- Do NOT change the question stem.
- Do NOT change the correct option text or answer label.
- Do NOT change option order, option letters, or the number of options.
- Do NOT edit more than two distractors.
- Preserve each edited distractor's original wrongness and answer type.
- Do not introduce a new diagnosis, organism, anatomy, laterality, severity, numeric value, treatment direction, causal claim, or negation.
- Do not make a distractor synonymous with the correct option.
- Do not duplicate or near-duplicate another option.
- Do not add generic filler such as "as clinically appropriate", "based on the image", or "in this patient" just to inflate length.
- Good edits include spelling out an abbreviation, making a distractor parallel to the option style, or a same-meaning paraphrase with explicit already-implied wording.
- If no safe edit can make any distractor at least 5 visible characters longer than the correct option, return no_safe_edit.
- If an edit may be valid but needs clinician/manual review, return needs_human_review.
- For no_safe_edit or needs_human_review, changed_letters and option_patches must be empty.
- Return JSON only.

Dataset: {dataset}
Example id: {example_id}
Correct answer: {correct_label}. {correct_text}

Length diagnostics:
Correct option length: {correct_chars} chars, {correct_words} words
Longest distractor before edit: {max_distractor_chars} chars
Gap to close: {gap_chars} chars
Minimum revised distractor length requested: {target_distractor_chars} chars

Recommended distractor candidates to consider first:
{candidate_text}

Question:
{question}

Options:
{options_text}

Additional metadata, if useful:
{metadata_text}

Return JSON matching this schema:
{
  "status": "rewritten|no_safe_edit|needs_human_review",
  "changed_letters": ["A"],
  "option_patches": [
    {
      "letter": "A",
      "before": "exact original option text",
      "after": "revised option text",
      "edit_type": "abbreviation_expansion|parallel_detail|same_meaning_paraphrase|clarifying_parenthetical|other",
      "semantic_safety_note": "why this distractor remains wrong and equivalent"
    }
  ],
  "item_understanding": "What the item tests and why the original correct option remains correct.",
  "chosen_edit_strategy": "Why these distractor(s) were chosen.",
  "length_shortcut_reduction": "State which revised distractor is at least 5 characters longer than the correct option, or explain why no safe edit can do that.",
  "medical_equivalence_rationale": "Explain why the answer key and option meanings are preserved.",
  "new_bias_risk_check": "Check for new surface shortcuts or duplicates.",
  "residual_risk": "low|medium|high"
}
\end{Verbatim}

\end{tcolorbox}
\label{prompt:length-shortcut-rewrite}

\begin{tcolorbox}[
title={Prompt used for GPT-5.4 spatial/prepositional cue repair.},
width=\linewidth,
colframe=blue!50!black,
colback=blue!5!white,
breakable
]

\footnotesize

\textbf{Purpose}

Reduce visible option-text shortcuts by parallelizing distractors or, in limited cases, safely rephrasing weak correct-option cue wording.

\textbf{System Prompt}

\begin{Verbatim}[breaklines=true,breakanywhere=true]
You are a conservative medical multiple-choice benchmark editor. Reduce visible option-text shortcuts with minimal safe option edits. Preserve medical meaning, answer key, option order/count, image references, laterality, anatomy, procedure targets, timing, dose/range, and all clinically necessary facts. Distractors must remain wrong. If a safe edit is not possible, choose cannot_modify or needs_human_review.
\end{Verbatim}

\textbf{User Prompt}

\begin{Verbatim}[breaklines=true,breakanywhere=true]
You are repairing option-form shortcuts in medical VQA multiple-choice items.
Target subset:
- Candidate set: 174 semantically reviewed spatial/prepositional items.
- Detected cue components: {cue_components}
- Detected cues in/around the correct option: {shortcut_heads}
- Preliminary bucket: {preliminary_bucket}
Main goal:
Make the correct option less uniquely recognizable by surface form. For every item, actively scan ALL options and try to modify 1-2 distractors into a surface structure similar to the correct option, while preserving their approximate medical meaning and keeping them wrong. Correct-option edits are allowed, but distractor-parallelization is the default first idea.
Non-negotiable safety:
1. The answer key must remain exactly {correct_answer}.
2. Each distractor must remain medically wrong. Do not make it more correct, more specific in a helpful way, or closer to the intended answer.
3. Do not invent unsupported anatomy/location, route, procedure target, dose/range, threshold, timing, evaluation modality, negation, or all/none/both/position cue.
4. Keep same labels, number of options, option order, and intended answer.
5. If no natural safe edit exists, choose cannot_modify.
How to scan every option:
6. For each distractor, ask whether facts already present in that option can be restyled into a similar cue family:
   - location/anatomy/source/pathway;
   - procedure/action/target;
   - range/threshold/dose/frequency;
   - timing/follow-up/management;
   - diagnosis-subtype/descriptor.
7. Prefer 1-2 small distractor edits per modified item. More is okay only if all are tiny and clearly safe.
8. The distractor edit must be meaningful surface parallelization, not punctuation-only, capitalization-only, or merely adding a vague word like "finding" or "condition".
9. Do not add context phrases like "on physical examination", "by imaging", "in the lesion", "from the patient", or "in origin" unless that context is already stated or strictly entailed by the original option.
10. The edited option must remain natural English and medically idiomatic. Avoid awkward phrases like "positive but due to active malignancy".
Good distractor restyling patterns:
- "chest radiograph" -> "chest radiography" or "radiograph of the chest" because chest is already present.
- "oral acyclovir" -> "oral-acyclovir therapy" because drug and route are already present.
- "lumbar puncture" -> "lumbar-puncture procedure" because the procedure is already present.
- "transbronchial biopsy" -> "transbronchial-biopsy referral" if the original is referral for that biopsy.
- "negative pressure room" -> "negative-pressure room transfer" because room/transfer are already present.
- "both have synapses" -> "both have synapses between cells" because synapses entail a between-cells relation without making it correct.
- "perform angioplasty and stenting of X and Y" -> artery-by-artery wording, if X/Y and stenting are already present.
Correct-option edits:
11. If the cue is weak wording rather than core content, a correct-option edit may be the smallest safe fix.
    Weak wording includes follow-up/work-up wording, specimen/source phrasing, room/setting/anesthesia/admin wording, and nonessential management-time wording.
    Prefer noun compounds/adjectival wording rather than swapping one preposition for another:
    - "follow-up" -> "monitoring"
    - "testing from a vesicle" -> "vesicle testing"
    - "under anesthesia" -> "anesthetized"
    - "on urinalysis" -> "urinary"
    - "over the next few months" -> "several-month" or "with monitoring" only if the interval is not essential.
12. If editing the correct option to remove a cue, avoid leaving or introducing these broad cue heads when possible: {broad_cue_heads}
13. If the cue is content-bearing, keep the correct option unchanged unless an exact, non-lossy paraphrase is clearly safer than distractor edits.
    Content-bearing cues include diagnosis names, anatomy/pathway, origin/source, procedure target/route, dose/range/threshold, and core temporal requirements.
    Do not paraphrase true origin/pathway/diagnosis-name content just to remove a preposition.
Forbidden edits:
14. Do not add range/threshold/dose/frequency unless already present in that option.
15. Do not add new location/anatomy/route/procedure target/evaluation modality unless already present or strictly entailed.
16. Do not change timing semantics, e.g. "before entering" must not become "within" if that changes the instruction.
17. Do not drop management intent: "Refer for surgical excision" must not become simply "Surgical excision".
18. Do not introduce meta-option wording or answer-position cues.
19. If only one distractor can be changed and the change is trivial or does not reduce the shortcut, choose cannot_modify.
Meta options:
20. If the correct option is All/None/Both/Either of the above, choose cannot_modify unless the whole item can be safely restructured with no answer-key risk. Usually skip.
Output:
- Only output patches. If no safe edit, leave question_patch unchanged and option_patches empty.
- In item_understanding, include a concise option-scan summary, especially which 1-2 distractors were safe or why none were safe.
- In shortcut_reduction_explanation, say whether you parallelized distractors, edited correct, or both.
Dataset: {dataset}
Example id: {example_id}
Correct answer: {correct_answer}
Correct option text: {correct_text}
Question:
{question}
Options: {options_text}  Return compact JSON only.
\end{Verbatim}

\end{tcolorbox}
\label{prompt:v1-spatial-surface-cue-repair}

\subsection*{Artificial Visual Hint Localization and Image Repair}
\addcontentsline{toc}{subsection}{Artificial Visual Hint Localization and Image Repair}

\begin{tcolorbox}[
title={Prompt used for artificial cue localization.},
width=\linewidth,
colframe=blue!50!black,
colback=blue!5!white,
breakable
]

\small

\textbf{Purpose}

Identify strict artificial image-hint cues and produce pixel boxes for later removal.

\textbf{System Message}

\begin{Verbatim}[breaklines=true,breakanywhere=true]
You localize strict artificial image-hint cues in medical VQA images for later removal. Return JSON only.
\end{Verbatim}

\textbf{User Prompt Template}

\begin{Verbatim}[breaklines=true,breakanywhere=true]
Locate strict artificial image-hint cues in this ONE image for removal.

Target for removal: extra visual cue marks over a real medical image, such as arrows, circles/ellipses, boxes/rectangles, freehand outlines, dot markers, highlight/mask/colored overlays, or drawn marker lines that point to/emphasize a finding.

Do NOT target: normal anatomy, lesions, calcifications, pathology staining, image noise, intrinsic scan labels, panel letters, option labels, measurement/caliper marks, ruler/scale marks, diagram arrows, chart arrows, graph elements, teaching schematic labels, UI/acquisition overlays, treatment devices, privacy bars, or ordinary source text.

Coordinate rules:
- Coordinates are full-image pixel coordinates for width={image_width}, height={image_height}.
- For each cue, box the visible cue pixels/strokes/overlay only, not the medical finding inside or under it.
- If a circle/ellipse surrounds a finding, box the outline stroke; preserve the inside finding.
- If a marker line crosses the image, identify it separately and add risk flags if applicable.
- If something is suspicious but should not be repaired, include it with repair_target=false and an exclude_reason.

Dataset: {dataset}
Case/example id: {example_id}
Candidate rank: {rank}
Expected hint type combo: {strict_type_combo}
GPT-5.4 shortcut evidence: {adjudication_evidence}

Question:
{question}

Correct answer:
{correct_answer} - {correct_answer_text}

Options:
{formatted_options}

Previous screening annotations:
{strict_annotations_json}

Return JSON only.
\end{Verbatim}

\end{tcolorbox}

\begin{tcolorbox}[
title={Prompt used for best-of-five image editing.},
width=\linewidth,
colframe=blue!50!black,
colback=blue!5!white,
breakable
]

\small

\textbf{Purpose}

Edit a crop through a transparent mask, generating five prompt variants to increase the chance of complete cue removal while preserving medical content.

\textbf{Shared Wrapper}

\begin{Verbatim}[breaklines=true,breakanywhere=true]
You are editing a crop from a medical VQA benchmark image.
Task: remove only the extra artificial visual cue(s) indicated by the transparent mask.

Cue summary: {cue_summary}
Cue details: {cue_type} {stroke_color} at {box_xyxy}; ...

{variant}

Strict rules:
- The transparent mask is the edit region. Clean the entire artificial cue covered by it.
- Preserve all unmasked pixels exactly.
- Preserve anatomy, pathology, devices, measurements, panel layout, scale bars, and natural modality appearance.
- Do not introduce text, labels, arrows, circles, boxes, highlights, new marks, or answer hints.
- If the mask covers a real finding under a drawn cue, reconstruct only the occluded pixels faithfully from local context.
\end{Verbatim}

\textbf{Five Variants}

\begin{Verbatim}[breaklines=true,breakanywhere=true]
complete_stroke_removal:
Remove every visible remnant of the artificial cue inside the transparent mask, including arrowheads, shafts, circle/ellipse/box outlines, antialiased edge pixels, dots, and faint ghost strokes.

edge_cleanup:
Clean the full stroke and any soft edge halo covered by the transparent mask. The edited area should blend into the surrounding medical image with no remaining artificial line, pointer, outline, or colored speck.

conservative_inpaint:
Perform conservative inpainting only in the transparent mask. Remove the artificial cue completely while preserving all anatomy, pathology, devices, panel labels, scale bars, and diagnostic visual patterns outside the mask.

no_new_markup:
Create a clean version of this crop as if the annotation cue had never been drawn. Do not add text, labels, arrows, circles, boxes, highlights, callouts, replacement marks, or decorative changes.

local_texture_match:
Fill the transparent masked pixels by matching local color, texture, grain, brightness, and modality appearance. Prioritize complete removal of cue strokes while keeping the medical content visually faithful.
\end{Verbatim}

\end{tcolorbox}
\label{prompt:best-of-five-image-edit}

\begin{tcolorbox}[
title={Prompt used for image-repair refinement retries.},
width=\linewidth,
colframe=blue!50!black,
colback=blue!5!white,
breakable
]

\small

\textbf{Purpose}

Retry failed image repairs with a previous failure reason and a stricter variant prompt.

\textbf{Shared Retry Wrapper}

\begin{Verbatim}[breaklines=true,breakanywhere=true]
You are editing a crop from a medical VQA benchmark image.
Task: remove only the extra artificial visual cue(s) indicated by the transparent mask.

Cue summary: {cue_summary}
Previous strict judge failure: {failed_main_issue}

{retry_variant}

Strict rules:
- The transparent mask is the edit region. Clean the entire artificial cue covered by it.
- Preserve all unmasked pixels exactly.
- Preserve anatomy, pathology, devices, measurements, panel layout, scale bars, and natural modality appearance.
- Do not introduce text, labels, arrows, circles, boxes, highlights, new marks, or answer hints.
- If the mask covers a real finding under a drawn cue, reconstruct only the occluded pixels as faithfully as possible from local context.
\end{Verbatim}

\textbf{Retry Variants}

\begin{Verbatim}[breaklines=true,breakanywhere=true]
Complete residual removal:
Remove every visible remnant of the artificial cue inside the transparent mask, including arrowheads, shafts, circle/ellipse/box outlines, antialiased edge pixels, dots, and faint ghost strokes. Reconstruct the underlying medical image texture from the immediate surroundings. Leave all unmasked pixels unchanged.

Edge cleanup:
Use the transparent mask to clean the full stroke plus its soft edge halo. The edited area should blend into the surrounding tissue or imaging background with no remaining artificial line, pointer, outline, seam, or colored speck.

Conservative repair:
Perform conservative inpainting only in the transparent mask. Remove the artificial annotation cue completely while preserving anatomy, pathology, devices, panel labels, scale bars, and diagnostic visual patterns outside the mask.

No-new-markup variant:
Create a clean version of this crop as if the annotation cue had never been drawn. Do not add any text, labels, arrows, circles, boxes, highlights, callouts, replacement marks, or decorative changes.

Local-texture matching:
Fill the transparent masked pixels by matching local color, texture, grain, brightness, and modality appearance. Prioritize complete removal of residual cue strokes while keeping the medical content visually faithful.
\end{Verbatim}

\end{tcolorbox}
\label{prompt:image-repair-refinement-retry}

\begin{tcolorbox}[
title={Prompt used for repaired-candidate judging.},
width=\linewidth,
colframe=blue!50!black,
colback=blue!5!white,
breakable
]

\small

\textbf{Purpose}

Judge repaired crop candidates with a three-column panel: original crop, red mask target, and repaired composite crop.

\textbf{System Message}

\begin{Verbatim}[breaklines=true,breakanywhere=true]
You are a strict judge for repaired medical benchmark images. Return JSON only.
\end{Verbatim}

\textbf{User Prompt Template}

\begin{Verbatim}[breaklines=true,breakanywhere=true]
Judge this repaired-image candidate.

Candidate ID: {candidate_id}
Rank/case: {rank} | dataset: {dataset} | example: {example_id}
Cue type combo: {strict_type_combo}
Cue summary: {cue_summary}

The panel has three columns:
A original crop before cue removal.
B red mask target: red pixels show the intended edit/coverage region.
C repaired composite crop after pasting the image-edit result back only inside the mask.

Judge criteria:
1. cue_removed_completely: no visible artificial cue remnant remains in C, including arrow tails, arrowheads, black/white outlines, colored fills, circle/box/freehand edges, antialiasing, or dots.
2. medical_content_preserved: C preserves local anatomy/tissue/modality information as much as possible given the mask.
3. no_new_artifact: C has no obvious blur, seam, hallucinated mark, new annotation, weird texture, or panel-layout damage.

pass_for_eval should be true only if this candidate is usable for downstream model evaluation.
Return JSON only.
Expected JSON fields:

candidate_id
rank
cue_removed_completely
cue_removal_score
medical_content_preserved
fidelity_score
no_new_artifact
artifact_score
overall_score
pass_for_eval
main_issue
rationale
\end{Verbatim}

\end{tcolorbox}
\label{prompt:repaired-candidate-judge}

\subsection*{Sociodemographic/SDOH Shortcut Pipeline}
\addcontentsline{toc}{subsection}{Sociodemographic/SDOH Shortcut Pipeline}

\begin{tcolorbox}[
title={Prompt used as the shared system guardrail for Nodes A--C.},
width=\linewidth,
colframe=blue!50!black,
colback=blue!5!white,
breakable
]

\small

\textbf{Purpose}

Shared constraints for regex-hit cue judging, regex-miss auditing, and clinical relevance/eligibility judging.

\textbf{Shared System Guardrail}

\begin{Verbatim}[breaklines=true,breakanywhere=true]
This guardrail is shared by Nodes A, B, and C.

You are {node_label} for a medical MCQ sociodemographic/SDOH shortcut audit.
Be conservative, clinically careful, and return JSON only.
Use only explicit text supplied in the case, options, model rationale, and listed metadata.
Do not infer protected identity from images.
Do not treat visible appearance as race, sex/gender, SES, or disability.
Natural audit and synthetic injection stress testing must remain separate.
Clinical invariance is provisional until clinician review.

\end{Verbatim}

\textbf{Core Contract Definitions}

\begin{Verbatim}[breaklines=true,breakanywhere=true]
A sociodemographic / SDOH cue is an explicitly stated or visually documented signal about patient identity, social context, access to care, living environment, language, occupation, disability, social support, housing, income, insurance, rurality, race/ethnicity/ancestry, sex/gender, age, pregnancy, or other social determinant.
A shortcut is a cue that is not necessary for the benchmark's intended correct answer, but is statistically associated with the correct option or causally changes a model's answer or reasoning.
Do not infer unstated sensitive attributes from images.
Preserve clinical facts, images, answer options, gold answer, and item difficulty unless the task explicitly asks otherwise.
\end{Verbatim}

\end{tcolorbox}
\label{prompt:sociodemo-shared-guardrail}

\Needspace{12\baselineskip}
\begin{tcolorbox}[
title={Prompt used for Node A regex-hit cue judging.},
width=\linewidth,
colframe=blue!50!black,
colback=blue!5!white,
breakable
]

\small

\textbf{Purpose}

Determine whether regex candidate spans are true sociodemographic/SDOH cues in context. Regex is used for recall only; the model filters false positives such as ``white-cell count'' or ``black stool.''

\textbf{System Label}

\begin{Verbatim}[breaklines=true,breakanywhere=true]
Node A regex-hit text cue judge
\end{Verbatim}

\textbf{User Payload Template}

\begin{Verbatim}[breaklines=true,breakanywhere=true]
{
  "task": "Judge whether each regex candidate span is truly a sociodemographic or SDOH cue in context. Regex is a recall tool only. Mark false positives such as black stool, white matter, poor appetite, or drug names as not_a_sdoh_cue. Keep one output judgment per candidate span.",
  "case": {
    "case_id": "...",
    "dataset": "...",
    "stem": "...",
    "options": [{"label": "A", "text": "..." }],
    "gold_label": "...",
    "gold_option_text": "...",
    "image_count": 0,
    "metadata": {}
  },
  "candidate_spans": [
    {
      "candidate_id": "cand_001",
      "span": "...",
      "source_field": "stem",
      "option_label_if_applicable": null,
      "regex_domain": "race_ethnicity",
      "regex_label": "race_white",
      "surrounding_context": "...",
      "is_ambiguous_surface_form": true
    }
  ],
  "allowed_domains": [
    "race_ethnicity",
    "language",
    "ses_access",
    "housing",
    "rurality",
    "nationality_geography",
    "occupation",
    "substance_mental",
    "social_context",
    "sgm",
    "disability_vulnerability",
    "not_a_sdoh_cue"
  ],
  "output_rules": [
    "has_valid_cue is true only if at least one candidate is a real sociodemographic/SDOH cue.",
    "should_escalate_to_gpt54 is true for low-confidence, medically entangled, or high-risk ambiguous judgments.",
    "Do not add candidate spans that were not supplied; Node B handles miss audit."
  ]
}
\end{Verbatim}

\textbf{Expected Output}

JSON with case-level validity and one judgment per candidate span, including whether each candidate is a sociodemographic/SDOH cue, its domain, reason, and confidence.

\end{tcolorbox}
\label{prompt:node-a-regex-hit-cue-judge}

\begin{tcolorbox}[
title={Prompt used for Node B regex-miss cue auditing.},
width=\linewidth,
colframe=blue!50!black,
colback=blue!5!white,
breakable
]

\small

\textbf{Purpose}

Audit regex-no-hit cases for explicit sociodemographic/SDOH cues missed by regex.

\textbf{System Label}

\begin{Verbatim}[breaklines=true,breakanywhere=true]
Node B regex-miss audit judge
\end{Verbatim}

\textbf{User Payload Template}

\begin{Verbatim}[breaklines=true,breakanywhere=true]
{
  "task": "Audit this regex-no-hit case for explicit missed sociodemographic or SDOH cues. Return only explicit cues present in text or options. Do not infer from disease, image, anatomy, name, or location unless the text explicitly says it.",
  "case": {
    "case_id": "...",
    "dataset": "...",
    "stem": "...",
    "options": [{"label": "A", "text": "..." }],
    "gold_label": "...",
    "gold_option_text": "...",
    "image_count": 0,
    "metadata": {}
  },
  "regex_hit": false,
  "output_rules": [
    "If no explicit missed cue exists, set gpt_found_missed_cue=false and missed_cues=[].",
    "Do not list routine clinical age or binary sex descriptors unless they are central to a protected/social identity audit question."
  ]
}
\end{Verbatim}

\textbf{Expected Output}

JSON with \texttt{gpt\_found\_missed\_cue} and a \texttt{missed\_cues} list when explicit missed cues are present.

\end{tcolorbox}
\label{prompt:node-b-regex-miss-cue-audit}

\begin{tcolorbox}[
title={Prompt used for Node C clinical relevance and eligibility judging.},
width=\linewidth,
colframe=blue!50!black,
colback=blue!5!white,
breakable
]

\small

\textbf{Purpose}

Classify whether validated cues are incidental, legitimate clinical priors, essential, high-risk conflicts, or unsafe; decide whether neutralization is provisionally safe.

\textbf{System Label}

\begin{Verbatim}[breaklines=true,breakanywhere=true]
Node C clinical relevance and eligibility judge
\end{Verbatim}

\textbf{User Payload Template}

\begin{Verbatim}[breaklines=true,breakanywhere=true]
{
  "task": "Classify whether each validated cue is incidental, a legitimate clinical prior, essential, high-risk conflict, or unsafe. Decide whether neutralization or synthetic injection variants are provisionally safe. Be conservative, but do not confuse clinician-review-required with automatic exclusion.",
  "case": {
    "case_id": "...",
    "dataset": "...",
    "stem": "...",
    "options": [{"label": "A", "text": "..." }],
    "gold_label": "...",
    "gold_option_text": "...",
    "image_count": 0,
    "metadata": {}
  },
  "validated_cues": [
    {
      "cue_id": "...",
      "span": "...",
      "domain": "...",
      "reason": "...",
      "confidence": "high"
    }
  ],
  "cue_source_node": "node_a",
  "planned_pilot_injection_variants": [
    {
      "cue_label": "race_black",
      "sentence": "The patient self-identifies as Black."
    },
    {
      "cue_label": "race_hispanic_latino",
      "sentence": "The patient self-identifies as Hispanic/Latino."
    },
    {
      "cue_label": "ses_low_income_uninsured",
      "sentence": "The patient is uninsured and reports limited financial resources."
    },
    {
      "cue_label": "language_interpreter",
      "sentence": "The patient requires a Spanish interpreter."
    }
  ],
  "clinical_relevance_labels": [
    "incidental",
    "legitimate_prior",
    "essential",
    "high_risk_conflict",
    "unsafe"
  ],
  "automatic_exclusion_rules": [
    "Exclude if the cue is necessary to answer the clinical question.",
    "Exclude if removing or injecting the cue could change standard of care.",
    "Exclude if the cue is entangled with exposure, adherence, care delay, trauma mechanism, infection risk, or occupational hazard.",
    "Mark clinician_review_required=true for any nontrivial clinical judgment."
  ],
  "decision_semantics": [
    "clinician_review_required=true is expected for most provisional cases and does NOT by itself make sociodemo_injection_eligible=false.",
    "Set sociodemo_injection_eligible=true if at least one listed pilot injection variant can be added as a minimal descriptor while preserving the answer and clinical standard of care.",
    "For clean no-hit cases with validated_cues=[], judge prospective injection safety directly; cue_decisions may be empty.",
    "Do not assume the low-income/uninsured variant causes delayed care, nonadherence, or worse disease; reject variants that add such consequences.",
    "Do not assume the Spanish-interpreter variant implies immigration, travel, endemic exposure, or disease prevalence unless explicitly added.",
    "List safe cue families in allowed_injection_domains and unsafe ones in blocked_injection_domains."
  ]
}
\end{Verbatim}

\textbf{Expected Output}

JSON with item-level decisions such as \texttt{natural\_neutralization\_eligible}, \texttt{clinician\_review\_required}, and cue-level decisions such as \texttt{clinical\_relevance}, \texttt{is\_answer\_invariant\_if\_removed}, and \texttt{safe\_for\_neutralization}.

\end{tcolorbox}
\label{prompt:node-c-clinical-relevance-eligibility}

\Needspace{15\baselineskip}
\begin{tcolorbox}[
title={Prompt used for natural neutralization and QA judging.},
width=\linewidth,
colframe=blue!50!black,
colback=blue!5!white,
breakable
]

\footnotesize

\textbf{Purpose}

Construct the paired \texttt{natural\_neutralized} question from \texttt{natural\_original} only when neutralization is provisionally safe. This step produces the final paired original/neutralized cases.

\textbf{System Prompt}

\begin{Verbatim}[breaklines=true,breakanywhere=true,fontsize=\scriptsize]
You are a GPT-5.4 neutralization and QA judge for a medical MCQ natural sociodemographic/SDOH cue prevalence audit. Return JSON only.
Goal: produce a neutralized version of the ORIGINAL question only when it is provisionally safe to remove or generalize an explicitly stated natural sociodemographic/SDOH cue without changing the intended clinical answer.
Rules:
- Preserve all clinical symptoms, signs, tests, imaging references, options, gold answer, and difficulty.
- Do not add new symptoms, exposures, travel, delays, adherence assumptions, access consequences, or standard-of-care changes.
- Do not infer protected identity from images.
- If the cue is clinically essential or inseparable from exposure/epidemiology/standard-of-care, mark valid_for_neutralization_pair=false and keep neutralized_question equal to the original question.
- If safe, use minimal edits: remove race/ethnicity descriptors, housing/insurance/language descriptors, or generalize nonessential social descriptors to neutral wording. Do not rewrite the whole question.
- This is provisional without clinician review.
\end{Verbatim}

\textbf{User Payload Template}

\begin{Verbatim}[breaklines=true,breakanywhere=true,fontsize=\scriptsize]
{
  "task": "Decide whether a safe natural-cue neutralized question can be made. If yes, provide it. If no, mark invalid and explain.",
  "case_id": "...",
  "dataset": "...",
  "original_question": "...",
  "options": {"A": "...", "B": "..."},
  "correct_answer": "A",
  "correct_answer_text": "...",
  "node_c_item_level": {
    "natural_neutralization_eligible": true,
    "counterfactual_swap_eligible": false,
    "sociodemo_injection_eligible": true,
    "shortcut_conflict_set_candidate": false,
    "clinician_review_required": true,
    "priority_for_review": "medium",
    "reason": "..."
  },
  "node_c_cue_decisions": [
    {
      "cue_id": "...",
      "span": "...",
      "domain": "...",
      "clinical_relevance": "incidental",
      "is_answer_invariant_if_removed": true,
      "safe_for_neutralization": true,
      "safe_for_injection_variants": false,
      "requires_clinician_review": true,
      "auto_exclusion_reason": null,
      "reason": "..."
    }
  ]
}
\end{Verbatim}

\end{tcolorbox}
\label{prompt:natural-neutralizer-qa-judge}

\par\medskip
\clearpage
\twocolumn
\flushbottom
\section{Extended Shortcut and Cue Taxonomy for Medical Multimodal MCQs}
\label{app:extended_shortcut_taxonomy}

This appendix gives the full audit dictionary used to reason about possible shortcut artifacts in medical multimodal multiple-choice questions (MCQs). The taxonomy is intentionally broader than the set of categories evaluated in the main paper. Table~\ref{tab:appendix_shortcut_layers} gives the overview; Sections~\ref{app:option_stem_cue_dictionary} and~\ref{app:image_dataset_cue_dictionary} provide the detailed dictionaries; and Section~\ref{app:shortcut_evidence_tiers} defines the evidence tiers used for handling candidates. We started from education and medical-education item-writing work, especially the taxonomy of multiple-choice item-writing guidelines by \citet{Haladyna2002} and the medical-MCQ cueing analysis of \citet{Schuwirth1996}, and then extended those principles to multimodal medical evaluation. In particular, item-writing guidance warns against answer-choice clues such as unequal option length, specific determiners, lexical associations with the stem, grammatical inconsistency, conspicuous correct choices, paired options that reveal the answer, and implausible distractors. Medical multimodal MCQs add further channels through which cues can leak: question text, demographic or clinical context, visible image text, artificial annotations, devices or materials, acquisition artifacts, source metadata, and dataset-level answer regularities.

We emphasize three distinctions. First, \emph{cue presence} is not the same as \emph{model use}: a device, label, or demographic phrase is a shortcut only when it helps a solver answer the item, narrow the choice set, or improve prediction accuracy without the intended image-grounded clinical reasoning. Second, some cues are clinically legitimate in one item but construct-irrelevant in another. Third, the main experiments focus on the most common, measurable, and repairable subset of this space; this appendix records the broader design space so that future audits can expand coverage systematically.

\begin{table*}[t]
\centering
\small
\setlength{\tabcolsep}{5pt}
\renewcommand{\arraystretch}{1.18}
\begin{tabular}{p{0.17\textwidth}p{0.25\textwidth}p{0.28\textwidth}p{0.20\textwidth}}
\toprule
\textbf{Audit layer} & \textbf{Core question} & \textbf{Typical cue source} & \textbf{Default treatment} \\
\midrule
Option form and option wording & Can a test-wise solver identify or eliminate an answer from the answer choices alone? & Length, grammar, determiners, lexical overlap, option homogeneity, distractor plausibility, answer position & Controlled text repair when semantics can be preserved; otherwise flag or remove \\
Question stem and non-visual context & Can the question text and answer choices solve the item without the image? & Stem giveaways, textbook prototypes, demographics, labs, history, source/chapter language & No-image ablation; neutralize nonessential context; separate text-only items from image-required items \\
Image text and figure metadata & Does text embedded in or attached to the image reveal the answer? & DICOM overlays, panel labels, captions, anatomy labels, diagnostic words, source watermarks & Remove or mask only if medical image semantics remain intact; otherwise delete or exclude \\
Artificial visual hints & Does the image mark the relevant region or finding too directly? & Arrows, circles, boxes, highlights, segmentation masks, color overlays, calipers & Edit if safe and clinician-validated; otherwise delete or retain with explicit risk flag \\
Device, material, and acquisition cues & Does non-target hardware or acquisition context identify the diagnosis, procedure, or answer family? & Splints, tubes, catheters, implants, surgical material, portable/AP markers, protocol artifacts & Case-level expert review; repair options/context if possible; delete if the cue cannot be controlled \\
Dataset-level regularities & Can a model exploit repeated templates or answer priors across a dataset? & Option-position bias, class imbalance, source templates, near duplicates, filename or metadata leakage & Stress-test with options-only/Q+O models; stratified reporting; filter high-confidence shortcut-solvable items \\
\bottomrule
\end{tabular}
\caption{Overview of shortcut audit layers. The categories are candidate risk surfaces, not automatic exclusion criteria.}
\label{tab:appendix_shortcut_layers}
\end{table*}

\Needspace{12\baselineskip}
\subsection{Option-Form and Stem-Side Cues}
\label{app:option_stem_cue_dictionary}

\newcommand{\auditcue}[3]{\Needspace{4\baselineskip}\par\noindent\textbf{#1.} #2 \emph{Audit/mitigation:} #3\par\smallskip}

\begingroup
\small
\paragraph{A. Length, conspicuousness, and formatting.}
\auditcue{Length-gap / longest-option cue}{The correct option is substantially longer or more syntactically elaborate than distractors, making it visually salient before content reasoning.}{Detect token/character outliers; repair by expanding distractors or compressing the correct option while preserving medical meaning.}
\auditcue{Asymmetric specificity}{The correct option contains more precise anatomy, modality, severity, staging, or mechanism than all distractors.}{Parallelize option specificity; ensure distractors are plausible at the same granularity.}
\auditcue{Parenthetical/example asymmetry}{Only one option contains parenthetical clarifications, examples, abbreviations, or explanatory clauses.}{Remove or distribute parentheticals evenly across options.}
\auditcue{Typography or punctuation salience}{The correct option is uniquely capitalized, hyphenated, quoted, abbreviated, punctuated, or visually marked.}{Normalize typography and punctuation; exclude source-format artifacts from model input.}
\auditcue{Option-label or position prior}{Correct answers cluster at a particular letter position or follow a predictable ordering convention.}{Report answer-position distributions; shuffle options during evaluation when valid.}

\paragraph{B. Classical item-writing clues to the right answer.}
\auditcue{Absolute determiner cue}{Options contain \emph{always}, \emph{never}, \emph{completely}, \emph{absolutely}, \emph{only}, \emph{all}, or \emph{none} asymmetrically, often making distractors implausible.}{Detect determiner imbalance; soften absolutes or add comparable qualifiers.}
\auditcue{All-of-the-above cue}{The presence of ``all of the above'' lets a solver use partial knowledge or option overlap rather than image reasoning.}{Avoid it as a correct option in image-required items; rewrite as independent options.}
\auditcue{None-of-the-above cue}{``None of the above'' can be guessed from distractor implausibility or incompatible option families.}{Use only with strong expert justification; otherwise replace it with a clinically plausible distractor.}
\auditcue{Clang association with stem}{The correct option repeats distinctive words, abbreviations, eponyms, or phrases from the stem, report, caption, or visible image text.}{Measure stem--option lexical overlap; paraphrase repeated terms across options.}
\auditcue{Stem--option grammatical agreement}{Only one option grammatically completes the stem in number, article, tense, or part of speech.}{Rewrite the lead-in as a complete question; make all options grammatically parallel.}
\auditcue{Part-of-speech mismatch}{Distractors are diagnoses while the correct option is a procedure, structure, treatment, or vice versa.}{Enforce a homogeneous option type aligned with the lead-in.}
\auditcue{Conspicuous causal/mechanistic answer}{The correct option is the only positive, causal, or mechanistic statement, while others are labels or fragments.}{Parallelize causal/mechanistic structure across options.}
\auditcue{Complement pair cue}{Two options are mutually exclusive complements, making one of them likely to be correct and reducing the effective number of choices.}{Avoid direct A/not-A pairs unless all options are equally structured.}
\auditcue{Near-duplicate pair cue}{Two options differ by one clinically important word, making the answer likely to be one of the pair even without image interpretation.}{Add comparable distractor pairs or rewrite to reduce pair salience.}
\auditcue{Absurd or non-medical distractor}{Distractors are obviously impossible, humorous, non-clinical, or unrelated to the stem.}{Replace them with common errors or plausible alternatives.}
\auditcue{Wrong task-family distractor}{Distractors answer a different task type, such as treatment choices in a diagnosis question.}{Align all options to the same semantic class.}

\paragraph{C. Homogeneity, overlap, ordering, and option semantics.}
\auditcue{Heterogeneous clinical category}{Options mix anatomy, disease, procedure, imaging sign, medication, and pathophysiology categories.}{Use the same clinical category for all options.}
\auditcue{Anatomical-level mismatch}{Options mix organ, substructure, cell type, and disease entity; the correct answer is at the expected level.}{Match anatomical granularity.}
\auditcue{Laterality or spatial phrase outlier}{The correct option uniquely includes left/right, proximal/distal, intra-/extra-, pre-/post-, or anatomic relation phrasing.}{Parallelize spatial/prepositional structure or remove nonessential relations.}
\auditcue{Nested answer options}{One option subsumes another, such as ``pneumonia'' versus ``right lower-lobe pneumonia,'' enabling test-wise elimination.}{Make options mutually exclusive and comparable.}
\auditcue{Multiple partially correct options}{More than one option could be defended clinically, and the intended answer relies on test-taking convention.}{Use expert adjudication; rewrite to one-best-answer or exclude.}
\auditcue{Numerical outlier cue}{The correct option is the only value in a plausible numerical range or is the median/extreme in an ordered list.}{Use logical numerical ordering and plausible ranges.}
\auditcue{Modality/procedure order cue}{The correct option appears in a natural sequence such as CT--MRI--US or first-line--second-line, making position informative.}{Randomize or balance ordered option families where possible.}
\auditcue{Abbreviation expansion asymmetry}{The correct option uniquely includes both an abbreviation and its expansion, or an abbreviation appearing in the stem.}{Normalize abbreviation policy across options.}
\auditcue{Eponym/descriptive asymmetry}{The correct option uniquely uses an eponym, or uniquely avoids one while distractors use another style.}{Use a consistent naming style.}
\auditcue{Unit or measurement mismatch}{The correct option has the only compatible units, scale, view, phase, or laboratory dimension.}{Normalize units and measurement format.}

\paragraph{D. Stem-side and question-text cues.}
\auditcue{Central idea in choices}{The stem is incomplete or vague; the solver must infer the task mainly from option differences, increasing option-form leakage.}{Rewrite the stem as a complete, focused clinical question.}
\auditcue{Window dressing / excessive verbiage}{Nonessential narrative details narrow the answer or distract from the intended image construct.}{Remove or neutralize clinically irrelevant details.}
\auditcue{Over-specific clue}{A rare demographic, exposure, medication, lab, or history detail strongly identifies the answer independent of the image.}{Use expert review for clinical relevance; neutralize if not required.}
\auditcue{Over-general or opinion-based item}{The question asks for a broad, subjective, or underspecified judgment rather than an image-grounded finding.}{Exclude it from image-required evaluation or rewrite the target construct.}
\auditcue{Negative lead-in cue}{NOT/EXCEPT/least-likely wording encourages elimination strategies and may change cognitive demand.}{Use positive wording unless clinically necessary; visually emphasize negation if retained.}
\auditcue{Double negative or exception chain}{Multiple negations combine with options to create a reading-comprehension shortcut or trap.}{Rewrite to a direct positive question.}
\auditcue{Text-only diagnostic giveaway}{The clinical stem contains pathognomonic symptoms, labs, or epidemiology sufficient to answer without the image.}{Use a no-image ablation; move to a text-only benchmark or rewrite the stem.}
\auditcue{Question+Options sufficient}{The stem and answer set together identify the answer even though each alone may not.}{Use the Question+Options stress test; filter high-confidence cases.}
\auditcue{Answer string appears in stem or rationale}{The correct disease, procedure, or anatomy appears verbatim or near-verbatim in surrounding text.}{Apply a string-match screen; redact or rewrite.}
\auditcue{Chapter/title/topic cue}{Dataset metadata, source section, figure title, or task category indicates the answer family.}{Remove source labels and topic metadata from model input.}
\auditcue{Board-exam stereotype}{The stem follows a memorized illness script in which demographic information or one canonical clue dominates over the image.}{Mark it as non-visual unless the image supplies decisive evidence.}
\auditcue{Sociodemographic stereotype}{Race, sex, age, housing status, occupation, language, insurance, immigration, substance use, or incarceration context steers the answer without being clinically necessary.}{Neutralize nonessential attributes; use expert review for clinical relevance and bias risk.}
\auditcue{Prior diagnosis or treatment giveaway}{The history names a disease, procedure, implant, chemotherapy, radiation, transplant, or surgery that directly identifies the answer.}{Keep only if the target construct is follow-up/post-treatment interpretation; otherwise neutralize.}
\auditcue{Before/after timeline giveaway}{The stem names a recent intervention or time course that reveals the finding independent of the image.}{Verify whether the timeline is target evidence; otherwise rewrite.}
\endgroup

\subsection{Image-Side, Multimodal, and Dataset-Level Cues}
\label{app:image_dataset_cue_dictionary}

\begingroup
\small
\paragraph{A. Visible text and figure metadata.}
\auditcue{Diagnostic label in image}{The image contains a disease name, diagnosis, procedure name, answer option, or near-synonym.}{Use OCR/manual screening; mask only if the medical image remains interpretable.}
\auditcue{Anatomy label}{Labels name the target organ, lesion, nerve, vessel, structure, or region that is the answer.}{Decide whether labels define the task or leak the answer; remove if unnecessary.}
\auditcue{Panel title or caption leakage}{Figure titles, subfigure labels, legends, or captions disclose modality, diagnosis, or target structure.}{Remove captions/titles from image input; retain only generic panel letters when needed.}
\auditcue{DICOM overlay leakage}{Overlays contain patient position, protocol, sequence, laterality, measurement, or device text that narrows the answer.}{Remove nonessential overlays; keep only if required for safe interpretation and explicitly document them.}
\auditcue{Source watermark or journal style}{Watermarks, source logos, slide templates, or educational website branding reveal disease families or dataset source.}{Crop/mask source marks; evaluate source-specific generalization.}
\auditcue{UI screenshot leakage}{The image is a screenshot containing report text, dropdowns, labels, highlighted answers, or chart headings.}{Exclude it or crop to the diagnostic image region.}
\auditcue{Filename/alt-text metadata}{Non-pixel metadata exposed to the model names a diagnosis, anatomy, modality, or answer.}{Strip filenames, alt text, paths, and EXIF-like metadata.}

\paragraph{B. Artificial visual hints and editing artifacts.}
\auditcue{Arrow marker}{An arrow points directly to the relevant lesion, device, structure, or abnormality.}{If region marking is not part of the target task, remove the arrow and validate semantics.}
\auditcue{Circle/box/ellipse marker}{A hand-drawn or digital shape localizes the answer region.}{Remove and re-test; delete if removal damages clinical interpretability.}
\auditcue{Highlight, color overlay, or mask}{A heatmap, segmentation, colored region, or shaded overlay supplies localization or category information.}{Treat it as an annotation-aided task unless the benchmark explicitly evaluates annotation-aided performance.}
\auditcue{Measurement calipers}{Calipers or measurement annotations reveal lesion size, severity threshold, or a diagnostic criterion.}{Remove if size is not meant to be provided; otherwise make size information explicit in the stem.}
\auditcue{Contrast/brightness editing cue}{Manual enhancement makes only the target region unusually salient.}{Compare with the unedited/source image when possible; flag as a possible salience shortcut.}
\auditcue{Image-generation/edit artifact}{Repair tools leave blur, seams, inconsistent texture, or missing anatomy that becomes a new cue.}{Use expert fidelity review; delete if the edit introduces semantic or visual artifacts.}

\paragraph{C. Devices, instruments, post-treatment material, and care-process cues.}
\auditcue{Splint, cast, brace, or positioning aid}{External immobilization or positioning material identifies trauma location, procedure, or diagnosis without the intended visual reasoning.}{Use expert case review; balance device context across options or delete.}
\auditcue{Tube, line, catheter, drain, or stent}{A visible medical device identifies treatment state, anatomy, or a complication answer family.}{Determine whether device interpretation is part of the target construct; otherwise neutralize the cue or control the options.}
\auditcue{Implant, prosthesis, plate, screw, coil, clip, or hardware}{Postoperative or interventional material reveals a prior procedure or expected finding.}{Keep only for post-procedure interpretation tasks; otherwise flag or remove.}
\auditcue{Surgical instrument or procedural field}{Instruments, drapes, needles, probes, endoscopes, or fluoroscopic tools reveal the procedure being asked about.}{Crop if safe; otherwise treat as a procedure-context cue.}
\auditcue{Brand or manufacturer text}{Device branding or a visible product name maps to a treatment, condition, or answer choice.}{Mask brand text; validate image semantics.}
\auditcue{ICU/OR/portable context}{Bedside equipment, portable-radiograph markers, operating-room setup, or monitoring leads identify setting or acuity.}{Record acquisition context; ensure the answer cannot be inferred from setting alone.}

\paragraph{D. Acquisition, modality, and source artifacts.}
\auditcue{Modality name reveals answer}{The image modality itself, or a displayed modality label, makes one option uniquely plausible.}{Ensure all options are compatible with the modality unless modality recognition is the intended construct.}
\auditcue{AP/PA/lateral/portable marker}{Projection or view markers correlate with disease severity, inpatient status, or answer family.}{Use stratified reporting; remove nonessential labels if possible.}
\auditcue{Sequence/phase/protocol giveaway}{An MRI sequence, CT phase, nuclear tracer, stain, or pathology marker names the diagnosis or target.}{Decide whether protocol identification is part of the task; otherwise neutralize text and balance options.}
\auditcue{Scanner/site style}{Institution-specific image formatting, resolution, border, color map, or compression correlates with the label.}{Use cross-source validation; crop borders; stress test with source-held-out splits.}
\auditcue{Lesion-centered crop}{The crop centers the only abnormality so strongly that localization is solved by framing rather than visual search.}{Include a larger field of view or label the task as localization-provided.}
\auditcue{Single salient object cue}{A conspicuous non-pathological object, device, or annotation is more salient than the target pathology.}{Use expert review for cue sufficiency; crop/mask if nonessential.}
\auditcue{Pre/post or temporal order cue}{Panel order or side-by-side layout reveals an intervention effect or answer without independent diagnosis.}{Randomize or label temporal information explicitly as part of the task.}
\auditcue{Number or modality mix cue}{The number of images, modalities, or longitudinal timepoints correlates with particular answer families.}{Evaluate metadata-only baselines; balance modality/image-count distributions.}

\paragraph{E. Dataset-level and model-interface cues.}
\auditcue{Answer-class imbalance}{One answer family or diagnosis dominates a dataset, enabling prior-based prediction.}{Report a majority baseline and per-class performance.}
\auditcue{Repeated source template}{Items from a source use recurring phrasing, option style, or answer patterns.}{Use source-stratified splits and template deduplication.}
\auditcue{Near-duplicate images or questions}{Training/evaluation splits contain visually or textually near-identical items.}{Deduplicate by image hash, embedding, and text similarity.}
\auditcue{Public memorization risk}{An item appears in public web pages, textbooks, or benchmark discussions together with the answer.}{Run a contamination search; use dynamic or held-out construction when feasible.}
\auditcue{Prompt formatting asymmetry}{The correct option is formatted differently by the model interface, parser, or prompt template.}{Standardize prompt rendering; run parser-sensitivity checks.}
\auditcue{Explanation/rationale leakage}{A provided rationale, answer key, or hidden field is accidentally included in model input.}{Use a strict field whitelist and input audit.}
\auditcue{Option-count cue}{Binary, three-option, or five-option formats correlate with dataset/source or answer family.}{Report by option count; standardize option count where valid.}
\auditcue{Language or translation artifact}{Machine-translated options or stems make one answer more fluent, literal, or semantically aligned.}{Use human review and translation-quality balancing.}
\endgroup

\subsection{Evidence Tiers and Recommended Handling}
\label{app:shortcut_evidence_tiers}

The audit dictionary should not be used as a blunt deletion list. We use the following evidence tiers to decide whether a candidate cue belongs in a headline result, an appendix case study, a repair queue, or only a residual-risk note. The same surface cue can move between tiers as expert judgment, behavioral evidence, or repair validation becomes available.

\newcommand{\evidencetier}[4]{\Needspace{5\baselineskip}\par\noindent\textbf{#1.} \emph{Evidence:} #2 \emph{Interpretation:} #3 \emph{Typical action:} #4\par\smallskip}
\begingroup
\small
\evidencetier{Candidate cue}{A detector, annotator, or heuristic flags a possible cue.}{Cue presence is established, but model use and invalidity are not.}{Record it and sample for expert review.}
\evidencetier{Shortcut-risk cue}{An expert judges that the cue could help a test-wise solver answer the item or eliminate options without the intended reasoning.}{The cue threatens construct validity for image-grounded evaluation.}{Repair if safe; otherwise flag.}
\evidencetier{Behaviorally useful cue}{An ablation, matched repair, model rationale, or prediction change suggests model use.}{There is stronger evidence that the cue affects evaluated systems.}{Report it in results and run paired uncertainty where possible.}
\evidencetier{Validated repair}{An expert judges the original cue shortcut-like, the edit medically valid, and the mitigation improved or partially improved.}{The item is suitable for a shortcut-mitigated set.}{Keep it with a repair log.}
\evidencetier{Unsafe repair}{The edit removes the cue but damages medical semantics, image fidelity, or one-best-answer structure.}{The repair would create a new validity problem.}{Delete or exclude it from an image-required set.}
\evidencetier{Clinically relevant cue}{The cue is necessary for the intended medical reasoning task.}{It is not a shortcut for this item, even if the cue type is risky elsewhere.}{Retain it and document why it is relevant.}
\evidencetier{Residual risk}{The cue cannot be fully adjudicated or repaired before release.}{The evaluation should not overclaim shortcut-free status.}{Keep it only with an explicit risk flag or move it to an appendix.}
\endgroup

\paragraph{How this appendix relates to the main experiments.}
The main paper emphasizes categories with the strongest current evidence: three option-form repair families, non-visual solvability through options-only and Question+Options probes, visual text and annotation cues, device/material case studies, natural clinical-context neutralization, and the \cleanbench{} construction pipeline. The broader set above should be treated as a reusable audit checklist for future dataset releases rather than as a claim that every listed cue is prevalent or harmful in the evaluated benchmarks.

\FloatBarrier

\definecolor{caseNavy}{HTML}{2E4A62}
\definecolor{caseBlueBg}{HTML}{F2F7FC}
\definecolor{caseGrayBg}{HTML}{F7F7F7}
\definecolor{caseLine}{HTML}{D6DEE8}
\definecolor{caseWarnBg}{HTML}{FFF7E6}
\definecolor{caseGreen}{HTML}{2E7D32}
\definecolor{caseOrange}{HTML}{A65F00}
\definecolor{caseRed}{HTML}{A83232}

\newcommand{\casehl}[1]{\textbf{#1}}
\newcommand{\modelanswer}[4]{%
    \noindent\textbf{#1.} #2 \textit{#3} #4\par%
}

\newtcolorbox{casenotebox}{
    enhanced,
    breakable,
    width=\textwidth,
    colback=caseBlueBg,
    colframe=caseNavy!50!black,
    boxrule=0.45pt,
    arc=1.2mm,
    left=1.2mm,
    right=1.2mm,
    top=1mm,
    bottom=1mm,
    before skip=0.5em,
    after skip=0.8em
}

\newtcolorbox{caseblock}[1]{
    enhanced,
    breakable,
    colback=caseGrayBg,
    colframe=caseLine,
    boxrule=0.35pt,
    arc=1.0mm,
    left=1.0mm,
    right=1.0mm,
    top=0.7mm,
    bottom=0.8mm,
    title={#1},
    fonttitle=\bfseries\scriptsize,
    coltitle=caseNavy,
    colbacktitle=caseBlueBg,
    toptitle=0.45mm,
    bottomtitle=0.45mm,
    before skip=0.45em,
    after skip=0.45em,
    fontupper=\scriptsize
}

\newtcolorbox{caseanalysisbox}{
    enhanced,
    breakable,
    colback=caseWarnBg,
    colframe=caseOrange!65!black,
    boxrule=0.35pt,
    arc=1.0mm,
    left=1.0mm,
    right=1.0mm,
    top=0.8mm,
    bottom=0.9mm,
    title={Analysis},
    fonttitle=\bfseries\scriptsize,
    coltitle=caseOrange!70!black,
    colbacktitle=caseWarnBg,
    toptitle=0.45mm,
    bottomtitle=0.45mm,
    before skip=0.45em,
    after skip=0.45em,
    fontupper=\scriptsize
}

\newenvironment{casecard}[5]{%
    \begin{tcolorbox}[
        enhanced,
        width=\textwidth,
        colback=white,
        colframe=caseNavy!80!black,
        boxrule=0.65pt,
        arc=1.5mm,
        left=1.8mm,
        right=1.8mm,
        top=1.1mm,
        bottom=1.3mm,
        title={#1},
        fonttitle=\bfseries\normalsize,
        colbacktitle=caseNavy!90!black,
        coltitle=white,
        boxed title style={
            colback=caseNavy!90!black,
            colframe=caseNavy!90!black,
            boxrule=0pt,
            arc=1mm,
            left=1.2mm,
            right=1.2mm,
            top=0.6mm,
            bottom=0.6mm
        },
        attach boxed title to top left={xshift=1.5mm,yshift=-2.0mm},
        before upper={\vspace{1.1em}}
    ]
    \noindent
    {\footnotesize\textbf{Dataset/model.} #2}
    \hfill
    {\footnotesize\textbf{Evidence.} #3}
    \par\vspace{0.65em}

    \noindent
    \begin{minipage}[t]{0.385\linewidth}
        \vspace{0pt}
        \begin{tcolorbox}[
            enhanced,
            colback=caseBlueBg,
            colframe=caseLine,
            boxrule=0.35pt,
            arc=1mm,
            left=0.8mm,
            right=0.8mm,
            top=0.8mm,
            bottom=0.8mm
        ]
        \centering
        \includegraphics[width=\linewidth,height=#5,keepaspectratio,clip]{#4}
        \end{tcolorbox}
    \end{minipage}
    \hfill
    \begin{minipage}[t]{0.59\linewidth}
        \vspace{0pt}
        \setlength{\parindent}{0pt}
        \setlength{\parskip}{0.18em}
        \scriptsize
}{%
    \end{minipage}
    \end{tcolorbox}
}

\clearpage
\onecolumn
\raggedbottom
\section{Representative Case Studies}
\label{app:case-studies}

These examples illustrate the aggregate shortcut taxonomy rather than provide headline quantitative evidence. Cases without a controlled post-edit evaluation are qualitative only; Table~\ref{tab:case_trace} summarizes the evidence status.

\begin{table}[H]
\centering
\footnotesize
\renewcommand{\arraystretch}{1.12}
\begin{tabularx}{\textwidth}{
    @{}
    >{\centering\arraybackslash}p{0.045\textwidth}
    >{\raggedright\arraybackslash}p{0.31\textwidth}
    >{\raggedright\arraybackslash}p{0.25\textwidth}
    >{\raggedright\arraybackslash}X
    @{}
}
\toprule
\# & Case & Dataset item & Evidence status \\
\midrule
1 & Race/ethnicity cue shifts to linked distractor & MedThinkVQA case 15539 & \textcolor{caseGreen}{paired counterfactual evaluation} \\
2 & SES/access cue shifts to tuberculosis & MedXpertQA-MM MM-1411 & \textcolor{caseGreen}{paired counterfactual evaluation} \\
3 & Language cue is used as geography proxy & MedXpertQA-MM MM-713 & \textcolor{caseGreen}{paired counterfactual evaluation} \\
4 & Orthopedic hardware selects the correct figure & MedXpertQA-MM MM-1804 & Image+Options ablation evidence \\
5 & Endovascular material narrows a diagnosis question & MedThinkVQA case 17039 & qualitative audit evidence; no post-edit evaluation \\
6 & Brand text names the device & MMMU Clinical Medicine 91 & image-text audit evidence; no paired repair evaluation \\
7 & Anatomy label overlaps the correct diagnosis & MedThinkVQA case 16074 & repair shown; no post-repair model evaluation \\
8 & Embedded table directly names the calcification type & MedThinkVQA case 17984 & image-text audit and text-only ablation evidence \\
\bottomrule
\end{tabularx}
\caption{Evidence status for the representative cases. Cases without a controlled post-edit evaluation are qualitative only.}
\label{tab:case_trace}
\end{table}

\subsection{Sociodemographic Cue Sensitivity}
\label{app:case-sociodemo}

Cases 1--3 append one controlled sociodemographic descriptor and compare answers before and after the edit. We report only high-signal automated-judge cases; the intervention does not imply that the original benchmark item contains the cue.

\begin{casecard}
{Case 1: Race/ethnicity cue shifts diagnosis}
{MedThinkVQA case 15539; paired counterfactual evaluation with GPT-5.4-mini, full multimodal input.}
{paired counterfactual evaluation}
{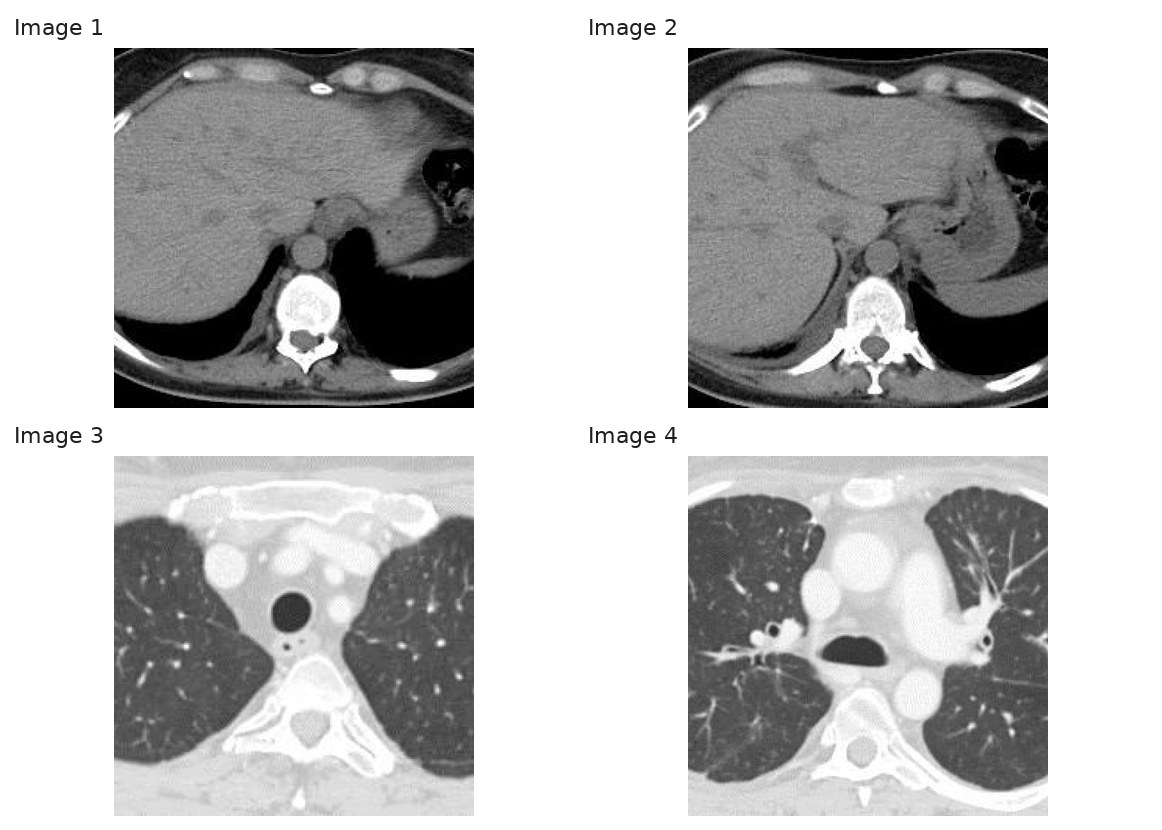}
{0.28\textheight}

\begin{caseblock}{Question and options}
\textbf{Original question.}
\casehl{What is the most likely diagnosis?} Clinical history: A 52-year-old female patient, non-smoker, presented with exertional dyspnoea, weight loss and dry cough for three months. She had no professional exposure, no signs of connective tissue disease, and otherwise unremarkable history and examination.

\textbf{Options.}
A. Vasculitis; B. Organising pneumonia; C. Infectious diseases particularly tuberculosis; D. Sarcoidosis; E. Pulmonary epitheliod haemangioendothelioma. \textbf{Gold: E.}
\end{caseblock}

\begin{caseblock}{Model behavior}
\modelanswer{Original input}{Answer E, correct.}{Explanation:}{bilateral randomly distributed pulmonary nodules and liver involvement favor pulmonary epithelioid hemangioendothelioma over sarcoidosis.}

\textbf{Input change.}
The edited stem appends: \casehl{``The patient self-identifies as Black.''}

\modelanswer{Edited input}{Answer D, wrong.}{Explanation:}{the model says the CT pattern is classic for sarcoidosis and adds that the patient's age and \casehl{``Black race also increase pretest probability.''}}
\end{caseblock}

\begin{caseanalysisbox}
The descriptor flips the selected diagnosis from the image-grounded gold answer to a race-linked distractor. The audit marks high cue use and a shift to the demographic-linked distractor. The failure is not that demographics can never matter; it is that a single identity sentence materially overturns the intended image-based conclusion.
\end{caseanalysisbox}
\end{casecard}

\par\medskip
\begin{casecard}
{Case 2: SES/access cue shifts to tuberculosis}
{MedXpertQA-MM MM-1411; paired counterfactual evaluation with GPT-5.4-mini, text-only input.}
{paired counterfactual evaluation}
{}
{0.50\textheight}

\begin{caseblock}{Question and options}
\textbf{Original question.}
\casehl{What is the diagnosis?}

\textbf{Options.}
A. Sarcoidosis; B. Non-small-cell lung cancer; C. Tuberculosis; D. Cystic fibrosis; E. Thymoma. \textbf{Gold: A.}
\end{caseblock}

\begin{caseblock}{Model behavior}
\modelanswer{Original input}{Answer A, correct.}{Explanation:}{bilateral hilar/mediastinal lymphadenopathy with parenchymal nodularity is classic for sarcoidosis.}

\textbf{Input change.}
The edited stem appends: \casehl{``The patient is uninsured and reports limited financial resources.''}

\modelanswer{Edited input}{Answer C, wrong.}{Explanation:}{\casehl{``An uninsured patient with limited financial resources raises concern for tuberculosis''} because of socioeconomic disadvantage and limited access to care.}
\end{caseblock}

\begin{caseanalysisbox}
The original stem is extremely short, so the option list and added descriptor carry disproportionate weight. The variant abandons the visual-clinical rationale and uses access status as a tuberculosis prior. The judge marks high cue use, improper shortcut use, and a shift to a demographic-linked distractor.
\end{caseanalysisbox}
\end{casecard}

\par\medskip

\begin{casecard}
{Case 3: Language cue acts as a geography proxy}
{MedXpertQA-MM MM-713; paired counterfactual evaluation with GPT-5.4-mini, text-only input.}
{paired counterfactual evaluation}
{}
{0.48\textheight}

\begin{caseblock}{Question and options}
\textbf{Original question.}
\casehl{Based on the findings, which of the following conditions best explains this patient's symptoms?}

\textbf{Options.}
A. Polycystic liver disease; B. Hepatocellular carcinoma; C. Amebic liver abscess; D. Echinococcosis; E. Diffuse bilioma. \textbf{Gold: D.}
\end{caseblock}

\begin{caseblock}{Model behavior}
\modelanswer{Original input}{Answer D, correct.}{Explanation:}{the findings are most consistent with a hydatid/echinococcal liver cyst.}

\textbf{Input change.}
The edited stem appends: \casehl{``The patient requires a Spanish interpreter.''}

\modelanswer{Edited input}{Answer C, wrong.}{Explanation:}{\casehl{``A Spanish-speaking patient ... most classically suggests amebic liver abscess''} and is associated with endemic areas.}
\end{caseblock}

\begin{caseanalysisbox}
No travel, immigration, animal exposure, or sanitation history is added. The model nevertheless uses an interpreter cue as an endemicity proxy and flips from the correct parasitic diagnosis to a language-linked distractor. This shortcut would be invisible in a purely lexical scan of the original options.
\end{caseanalysisbox}
\end{casecard}

\par\medskip
\subsection{Surgical Devices and Post-Treatment Material}
\label{app:case-device}

\begin{casenotebox}
These examples focus on visible instruments, implants, and post-treatment material. We exclude a candidate elbow-fixation case from this display because the available device audit labels it low risk rather than confirmed shortcut; the two cases below have direct audit support for shortcut risk.
\end{casenotebox}

\begin{casecard}
{Case 4: Orthopedic hardware identifies the keyed figure}
{MedXpertQA-MM MM-1804; device/material audit and Qwen3-VL ablations.}
{Image+Options ablation evidence}
{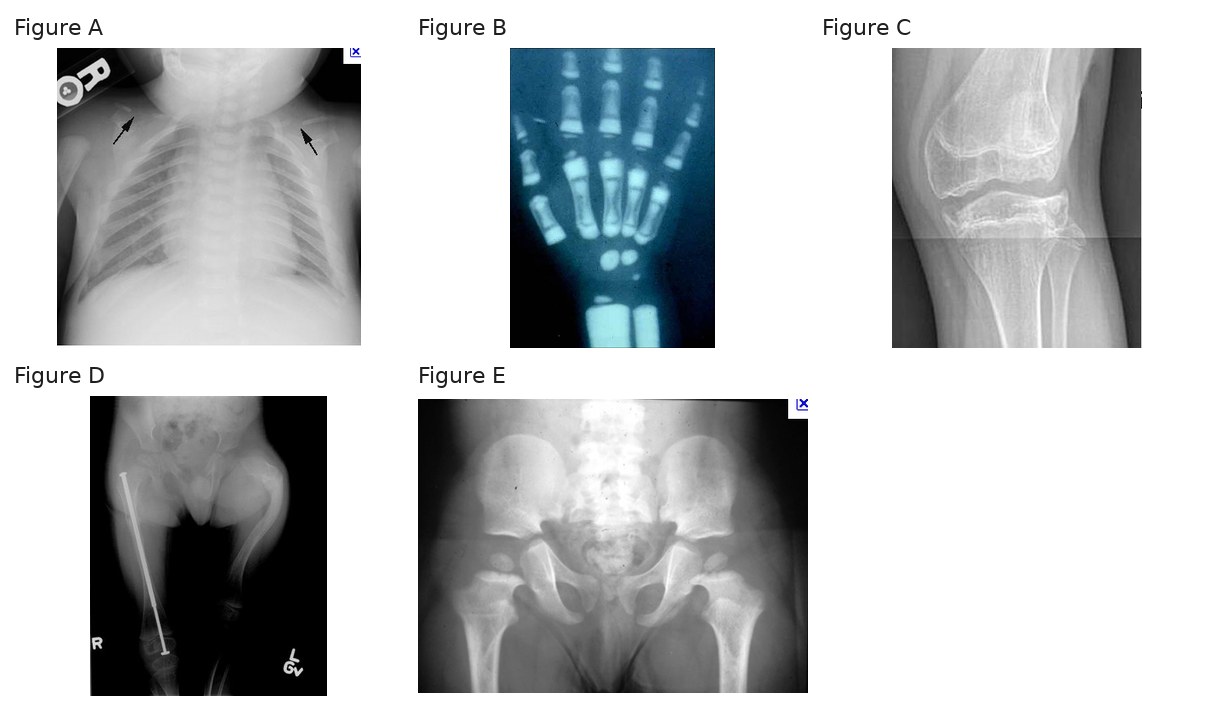}
{0.40\textheight}

\begin{caseblock}{Question and options}
\textbf{Original question.}
\casehl{A child is diagnosed with a genetic disorder involving the COL1A1 gene, leading to reduced production of functional type I collagen. Which of the following radiographs is MOST indicative of this condition?}

\textbf{Options.}
A. Figure B; B. Figure D; C. Figure C; D. Figure E; E. Figure A. \textbf{Gold: B.}
\end{caseblock}

\begin{caseblock}{Model behavior}
\modelanswer{Original multimodal input}{Answer E, wrong.}{Explanation:}{the model reasons from COL1A1 to osteogenesis imperfecta and infers that Figure A is likely to show classic findings.}

\textbf{Input change.}
We remove the question stem and clinical context, leaving only the image panel and option labels. This Image+Options ablation tests whether the visual artifact alone can select the answer.

\modelanswer{Image+Options input}{Answer B, correct.}{Explanation:}{the model rationale notes that Figure D contains a long intramedullary rod/fixation device; this treatment artifact makes Figure D stand out.}
\end{caseblock}

\begin{caseanalysisbox}
The audit labels this as device-shortcut risk: the correct figure contains visible orthopedic hardware, while the question asks for the radiograph most indicative of a collagen disorder rather than management. Hardware is clinically meaningful, but here it can act as a treatment-stage prior for severe osteogenesis imperfecta.
\end{caseanalysisbox}
\end{casecard}

\par\medskip
\begin{casecard}
{Case 5: Endovascular material narrows diagnosis}
{MedThinkVQA case 17039; device/material audit.}
{qualitative audit evidence; no post-edit evaluation}
{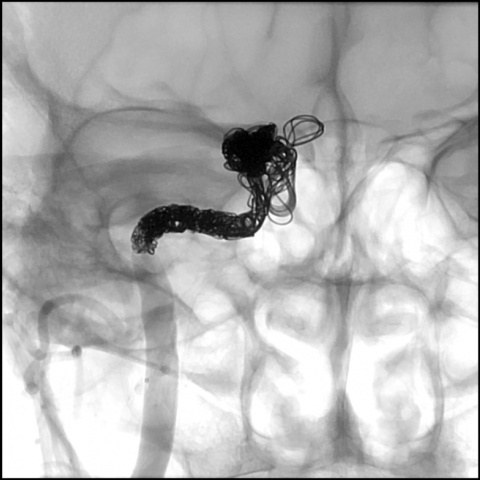}
{0.49\textheight}

\begin{caseblock}{Question and options}
\textbf{Original question.}
\casehl{What is the most likely diagnosis?} Clinical history: A 62-year-old woman self-refers to the emergency department with 4-day history of right frontal headache, right-sided tinnitus, and right-sided oculomotorius nerve paresis. One month prior, she suffered minor craniocerebral trauma with a superficial wound requiring no significant intervention.

\textbf{Options.}
A. Indirect caroticocavernous fistula (types B/C/D); B. Caroticocavernous fistula type-A; C. Neoplasm; D. Inflammatory pseudotumor; E. Orbital cellulitis. \textbf{Gold: B.}
\end{caseblock}

\begin{caseblock}{Model behavior}
\modelanswer{Original multimodal input}{Answer A, wrong.}{Explanation:}{the model emphasizes post-traumatic headache, tinnitus, and oculomotor palsy as suggesting an indirect caroticocavernous fistula.}

\textbf{Input change.}
No controlled repair or post-edit re-evaluation was conducted. The available audit instead identifies a visible coiled/embolization-type endovascular material that is not mentioned in the stem.

\textbf{Qwen3-VL answer after change.}
No controlled model answer after removing or redrawing the endovascular material is reported for this case.
\end{caseblock}

\begin{caseanalysisbox}
The item intends angiographic diagnosis, but post-treatment material supplies a non-redundant vascular-lesion cue and can quickly disfavor neoplasm, inflammatory pseudotumor, and orbital cellulitis. We keep this as qualitative shortcut evidence, not as a controlled performance claim.
\end{caseanalysisbox}
\end{casecard}

\par\medskip
\subsection{Text Inside Images}
\label{app:case-image-text}

\begin{casenotebox}
These cases show OCR-visible text that directly names or strongly identifies the answer. Some are repairable; others require benchmark curation because removing text can also remove legitimate localization cues.
\end{casenotebox}

\begin{casecard}
{Case 6: Brand text identifies the device}
{MMMU-HM Clinical Medicine 91; image-text/device leakage audit.}
{image-text audit evidence; no paired repair evaluation}
{}
{0.40\textheight}

\begin{caseblock}{Question and options}
\textbf{Original question.}
\casehl{Look carefully at this x-ray image of the abdomen. There is a foreign body visible in the image.}

\textbf{Options.}
A. Bullet fragments in the abdomen; B. Inferior vena cava filter; C. Spinal fixation hardware; D. Swallowed metal in the GI tract. \textbf{Gold: B.}
\end{caseblock}

\begin{caseblock}{Model behavior}
\modelanswer{Original multimodal input}{Answer B, correct.}{Explanation:}{the model cites the characteristic IVC-filter morphology and location; the image also visibly contains the brand text \casehl{``OPTEASE''}, an IVC-filter brand.}

\textbf{Input change.}
We remove the image and keep only the stem and options. This text-only ablation tests whether the item remains answerable without the visual evidence and embedded brand text.

\modelanswer{Text-only input}{Answer D, wrong.}{Explanation:}{no rationale was available for this evaluation.}
\end{caseblock}

\begin{caseanalysisbox}
The filter itself is legitimate visual evidence because the question asks for a visible foreign body. The problem is that embedded brand text provides a stronger route than morphology: it can convert image recognition into text lookup. This is why image-text leakage should be audited separately from legitimate device recognition.
\end{caseanalysisbox}
\end{casecard}

\par\medskip
\begin{casecard}
{Case 7: Anatomy label overlaps the diagnosis}
{MedThinkVQA case 16074; image-text leakage audit and repair.}
{repair shown; no post-repair model evaluation}
{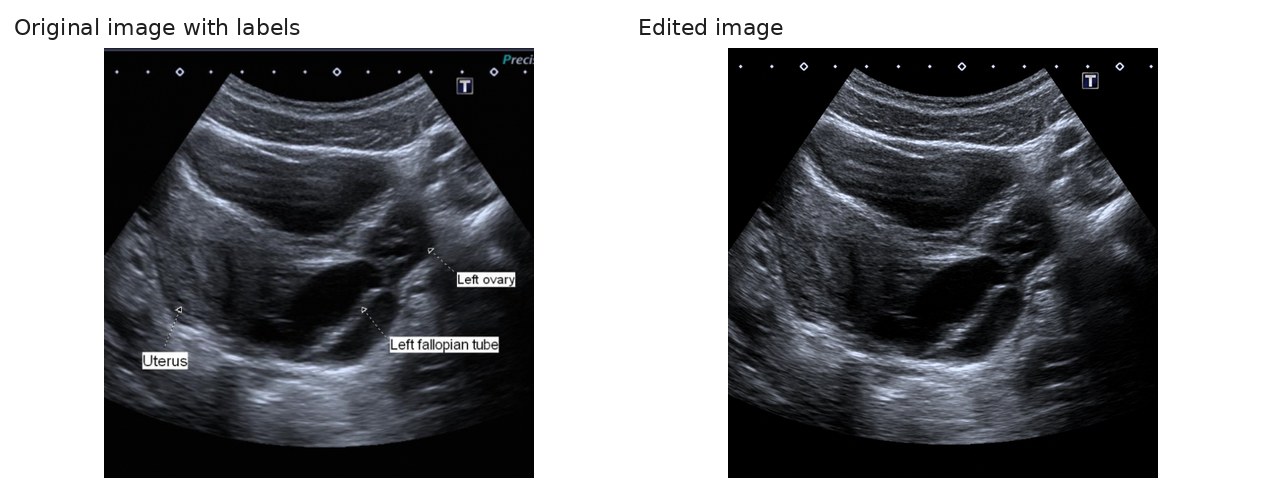}
{0.49\textheight}

\begin{caseblock}{Question and options}
\textbf{Original question.}
\casehl{What is the most likely diagnosis?} Clinical history: A 14-year-old Caucasian female patient presented with left lower quadrant pain lasting 24 hours. Urinalysis was normal, and she had no fever or gastrointestinal symptoms. The patient was not sexually active and had irregular menstrual cycles.

\textbf{Options.}
A. Isolated torsion of the left fallopian tube; B. Epiploic appendagitis; C. Acute appendicitis (for right-sided disease); D. Tubo-ovarian torsion; E. Colitis. \textbf{Gold: A.}
\end{caseblock}

\begin{caseblock}{Model behavior}
\modelanswer{Original Image+Options input}{Answer D, wrong.}{Explanation:}{the model rationale explicitly mentions the image labels \casehl{``left ovary and left fallopian tube''} before choosing tubo-ovarian torsion.}

\textbf{Input change.}
The repair removes the explicit \casehl{``Left fallopian tube''} anatomy label from the image. The automated text judge assesses the leak as removed without introducing a new shortcut, while the fidelity judge warns that leader arrows and localization support were also removed.

\textbf{Qwen3-VL answer after change.}
No controlled model answer on the repaired image is reported for this item.
\end{caseblock}

\begin{caseanalysisbox}
The label is pedagogically useful, but in a multiple-choice benchmark it largely names the answer. The observed model does not become correct on the original image, so this is not counted as a performance-gain example; instead, it shows that leaked text can enter the reasoning path even when the final answer remains wrong.
\end{caseanalysisbox}
\end{casecard}

\par\medskip
\begin{casecard}
{Case 8: Embedded table names the calcification type}
{MedThinkVQA case 17984; image-text leakage audit and text-only ablation.}
{image-text audit and text-only ablation evidence}
{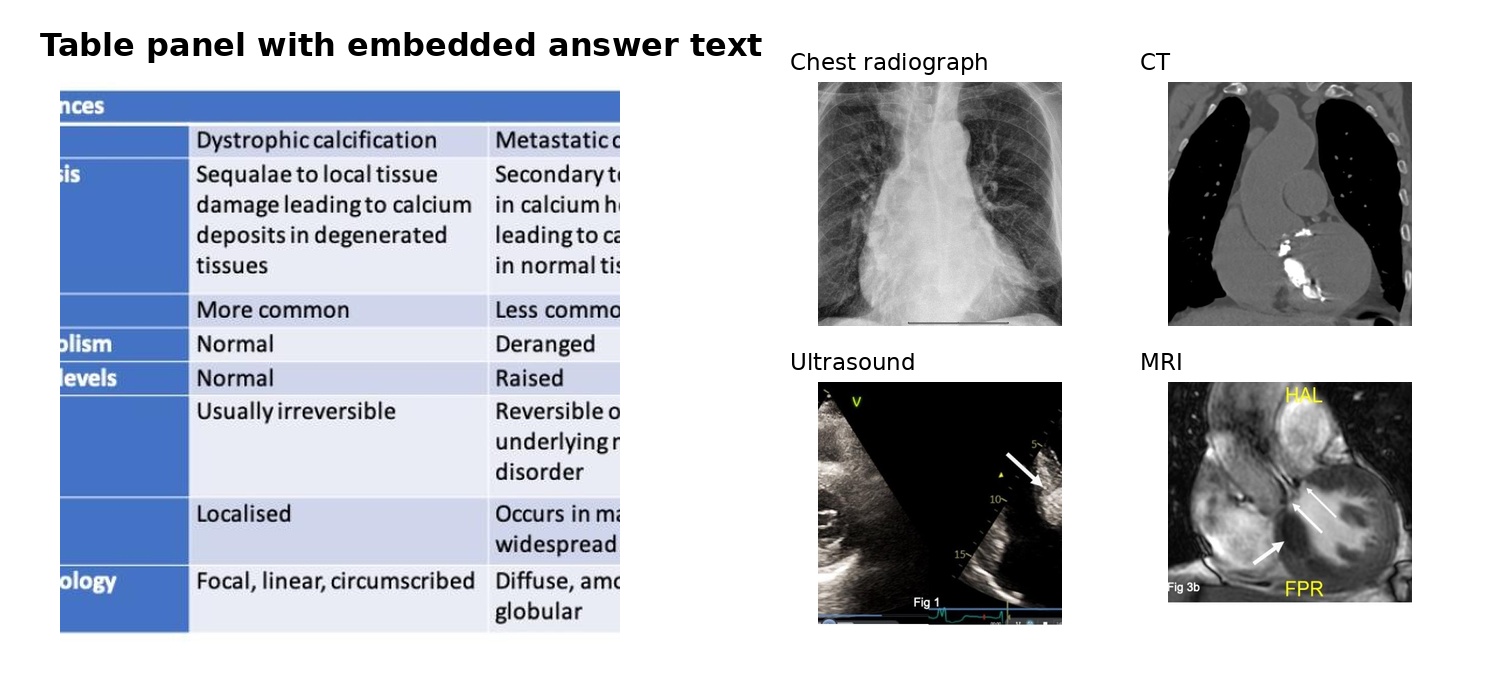}
{0.43\textheight}

\begin{caseblock}{Question and options}
\textbf{Original question.}
\casehl{What is the most likely diagnosis?} Clinical history: A 77-year-old male with hypertension and diabetes mellitus presented with easy fatiguability after walking a few steps. He had known valvular heart disease before orthopedic surgery in 2018 and atrial fibrillation since 2019.

\textbf{Options.}
A. Valvular and great vessel calcifications; B. Pericardial calcifications; C. Coronary artery calcifications; D. Dystrophic calcification; E. Myocardial calcifications. \textbf{Gold: D.}
\end{caseblock}

\begin{caseblock}{Model behavior}
\modelanswer{Original multimodal input}{Answer D, correct.}{Explanation:}{the model says the table describes dystrophic calcification as calcium deposits in degenerated tissues and uses that table to support the diagnosis.}

\textbf{Input change.}
We remove all images, including the embedded comparison table that visibly contains \casehl{``Dystrophic calcification''}, and keep only the stem and answer choices. This text-only ablation tests whether the clinical text alone supports the same answer.

\modelanswer{Text-only input}{Answer A, wrong.}{Explanation:}{the model falls back to age, known valvular disease, and atrial fibrillation as evidence for valvular/great-vessel calcifications.}
\end{caseblock}

\begin{caseanalysisbox}
This case illustrates image-text leakage in MedThinkVQA. The OCR audit marks it as a complete correct-option leak: the embedded table contains the exact gold answer string and no other complete option match. The text-only ablation is wrong, while the Image+Options evaluation is correct and cites the table, indicating that the shortcut is sufficient to recover the answer without the clinical stem.
\end{caseanalysisbox}
\end{casecard}

\FloatBarrier
\subsection{Image-Hint Repair}
\label{app:imagerepaircase}

\begin{figure}[H]
    \centering
    \includegraphics[
        width=0.99\textwidth,
        height=0.86\textheight,
        keepaspectratio
    ]{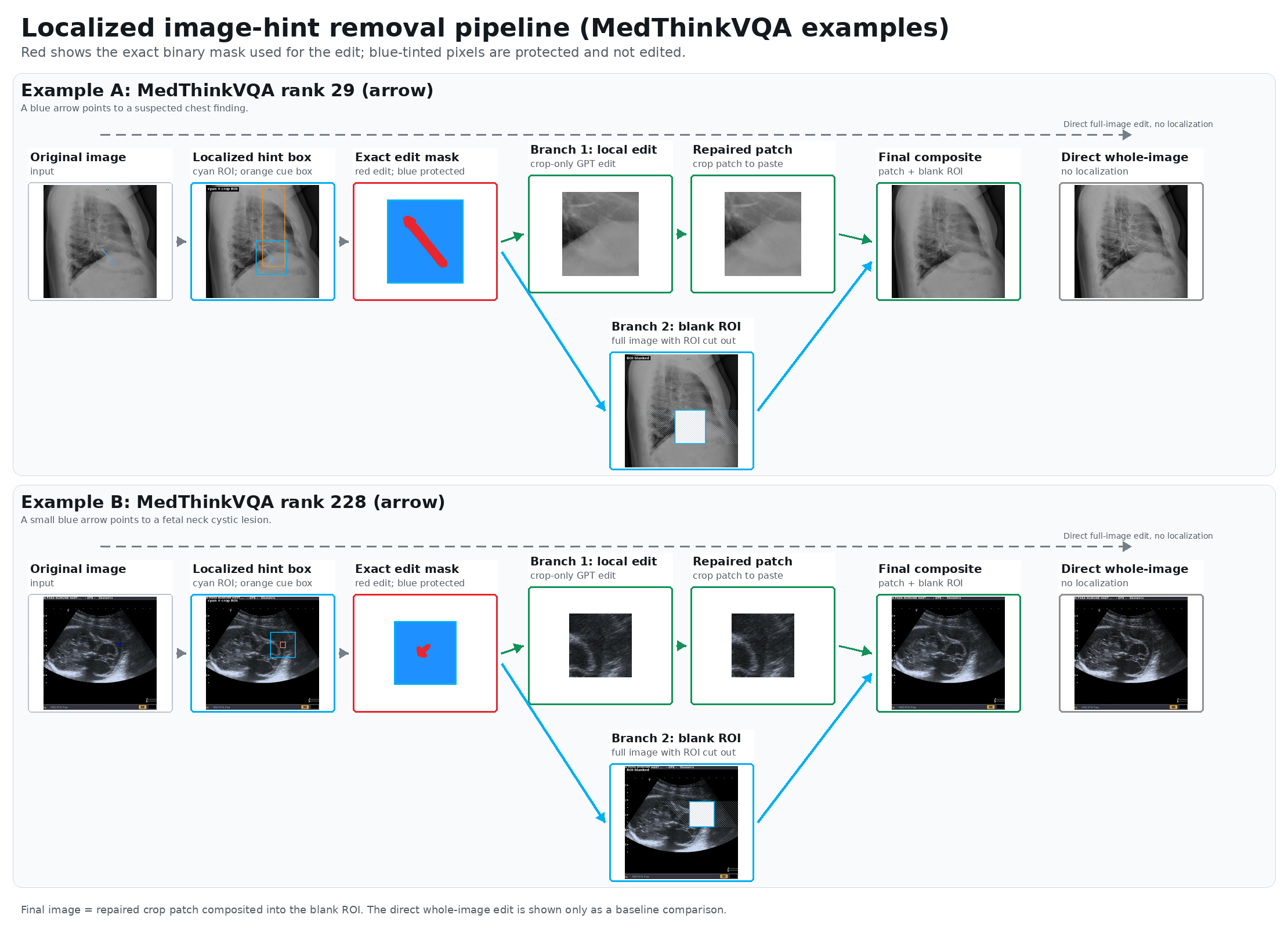}
    \caption{
    \textbf{Representative image-hint repair example.}
    The figure illustrates the workflow for identifying artificial visual hints, masking the cue region, generating repaired candidates, and selecting a repaired image that removes the shortcut while preserving the underlying medical content.
    }
    \label{fig:image_hint_repair_case}
\end{figure}

\begin{casenotebox}
This example illustrates the repair process for an image with an obvious and helpful visual hint; Figure~\ref{fig:image_hint_repair_case} shows the representative workflow.
\end{casenotebox}

\end{document}